\documentclass[10pt,twocolumn,letterpaper]{article}

\usepackage{cvpr}              

\newcommand{\suppl}{\textcolor{magenta}{\emph{Appx.}}\xspace}
\definecolor{headergray}{gray}{0.9}
\definecolor{rowgray}{RGB}{235, 245, 255}
\usepackage{siunitx}    

\definecolor{cvprblue}{rgb}{0.21,0.49,0.74}
\usepackage[pagebackref,breaklinks,colorlinks,allcolors=cvprblue]{hyperref}
\usepackage{amsmath}
\usepackage{amsthm}
\usepackage{amsfonts}
\usepackage{amssymb}
\usepackage{graphicx}
\usepackage{multirow}
\usepackage{xcolor}    
\usepackage{colortbl} 
\renewcommand{\paragraph}[1]{\vspace{1ex}\noindent\textbf{#1}\hspace{1em}}
\usepackage{threeparttable}
\usepackage{booktabs} 
\usepackage{multirow}
\usepackage{pifont}
\newcommand{\cmark}{\textcolor{red}{\ding{51}}}%
\newcommand{\xmark}{\textcolor{green}{\ding{55}}}%
\def\paperID{*****} 
\def\confName{CVPR}
\def\confYear{2026}

\title{AnyPcc: Compressing Any Point Cloud with a Single Universal Model}

\author{Kangli Wang$^{1 *}$ \quad Qianxi Yi$^{1,2 *}$ \quad Yuqi Ye$^{1}$ \quad Shihao Li$^{1}$ \quad Wei Gao$^{1,2 \dagger}$ \\
$^{1}$ SECE, Peking University\quad \textsuperscript{2} Peng Cheng Laboratory \\
{\tt\small kangliwang@stu.pku.edu.cn , gaowei262@pku.edu.cn} \\
\color{magenta} \texttt{\small Project Website:} \href{https://anypcc.github.io/}{\color{magenta} \texttt{\small anypcc.github.io}}
}
\begin{document}

\twocolumn[{%
\renewcommand\twocolumn[1][]{#1}%
\maketitle

\begin{center}
    \captionsetup{type=figure}
    \includegraphics[width=1.0\textwidth]{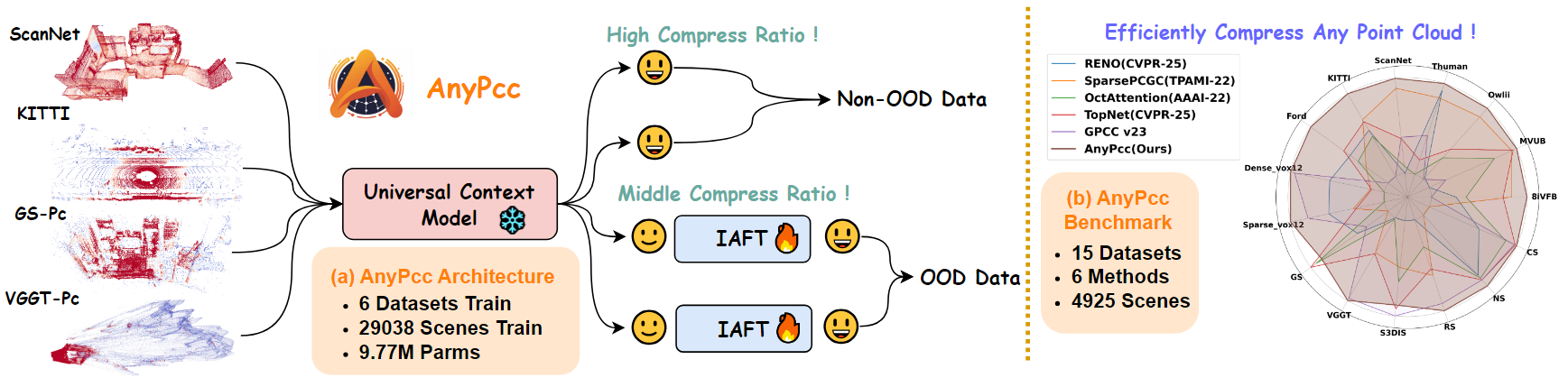}
    \captionof{figure}{\textbf{(a) AnyPcc Architecture.} A single, unified model compresses point clouds from any source, with our Instance-Adaptive Fine-Tuning (IAFT) module boosting performance on out-of-distribution (OOD) data. \textbf{(b) AnyPcc Benchmark.} Our comprehensive benchmark features 15 diverse datasets, including both standard and extreme cases. When compared against five state-of-the-art methods, AnyPcc consistently achieves high compression efficiency across all types of point clouds. }
    \label{fig:overview}
\end{center} 
}]
\begingroup\renewcommand\thefootnote{}\footnote{
\textsuperscript{*} Contributed equally to this work.
\textsuperscript{$\dagger$}Corresponding author.
}\addtocounter{footnote}{-1}\endgroup

\begin{abstract}
Generalization remains a critical challenge in deep learning-based point cloud geometry compression. While existing methods perform well on standard benchmarks, their performance collapses in real-world scenarios due to two fundamental limitations: the lack of context models that are robust across diverse data densities, and the inability to efficiently adapt to out-of-distribution (OOD) data. To overcome both challenges, we introduce AnyPcc, a universal point cloud compression framework. AnyPcc first employs a Universal Context Model that leverages coarse-grained spatial priors with fine-grained channel priors to ensure robust context modeling across the entire density spectrum. Second, our novel Instance-Adaptive Fine-Tuning (IAFT) strategy tackles OOD data by synergizing explicit and implicit compression paradigms. For each instance, it fine-tunes a small subset of network weights and transmits them within the bitstream. The minimal bitrate overhead from these weights is significantly outweighed by the resulting gains in geometry compression. Extensive experiments on a benchmark of 15 diverse datasets confirm that AnyPcc sets a new state-of-the-art in point cloud compression while maintaining low complexity. Our code and datasets have been released to encourage reproducible research.
\end{abstract}

\section{Introduction}
\label{intro}
\begin{figure*}
  \centering
  \includegraphics[width=1\linewidth]{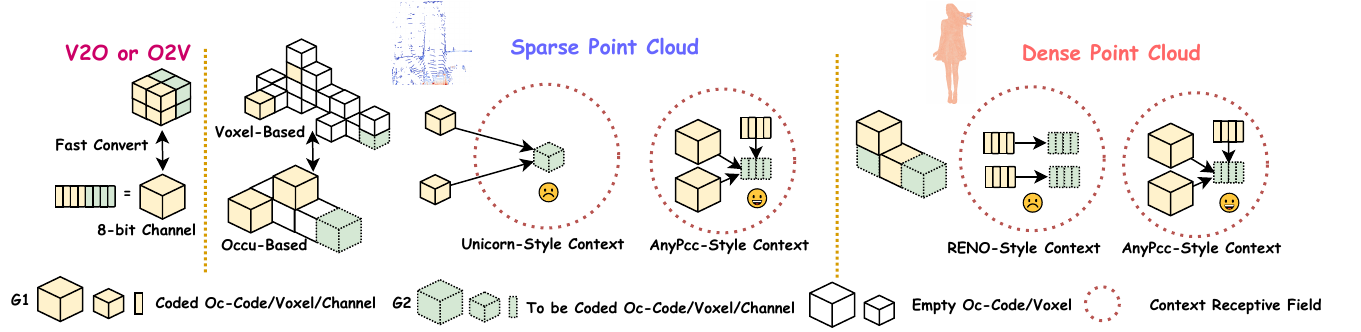}
  \caption{The superiority of our UCM in capturing contextual information across diverse point cloud types (e.g., dense and sparse).}
  \label{fig:context}
\end{figure*}

With the rise of 3D applications like autonomous driving and virtual reality, point clouds have become a standard format for 3D data. This widespread use makes efficient geometry compression essential for reducing storage and transmission costs. While recent learning-based methods outperform traditional standards like G-PCC \cite{gpcc} on certain standard benchmarks, a significant generalization gap often limits their practical use.
We argue this gap originates from two fundamental limitations. First, existing context models \cite{unipcgc,you2025reno,unicorn} are often tailored to \textbf{specific point cloud densities} and thus fail to maintain stable performance across the wide spectrum of densities found in real-world data, such as sparse LiDAR scans and dense reconstruction outputs. Second, these models \textbf{suffer a sharp decline in compression efficiency on out-of-distribution (OOD) data}. Even recent universal models like Unicorn-U~\cite{unicorn}, despite being trained for versatility, fall short due to two key issues: they rely on non-unified architectures, and their performance still collapses under extreme OOD conditions. We address these limitations sequentially, first by tackling context modeling.

Effective context modeling is crucial for efficient point cloud compression. While spatial-prior methods like Unicorn~\cite{unicorn} leverage strong contextual information at the voxel scale, they become unreliable in sparse scenarios. RENO~\cite{you2025reno} re-frames the problem as a channel-wise probability prediction task on occupancy codes. Although their methodologies differ, the spatial context modeled by Unicorn and the channel-wise context by RENO are fundamentally two inter-convertible representations of the same local geometry. Operating at the coarser-grained occupancy code scale ensures sparsity-robustness and achieves a larger effective receptive field. However, by exclusively grouping fine-grained priors, both methods fail to leverage the valuable structural information available from coarser-grained contexts, as illustrated in Figure~\ref{fig:context}. We resolve this with our Universal Context Model (UCM), which synergistically integrates channel grouping for \textbf{fine-grained priors} with spatial grouping for \textbf{coarse-grained structural priors}. This unified approach allows the UCM to capture a richer, more complete set of contextual dependencies, enabling robust compression across the entire density spectrum.

Although our UCM achieves favorable compression results via large-scale pre-training, its performance still degrades on \textbf{\textit{extreme}} OOD samples. Meanwhile, while some works have explored Implicit Neural Representations (INRs) for generalization~\cite{dupont2021coin,inr-pcc,chen2021nerv,huang2025linr}, their requirement to train a unique network from scratch for every instance results in impractically long encoding times. Therefore, to transcend the slow encoding time of INRs and the generalization constraints of a fixed pre-trained model, we propose \textbf{AnyPcc}. Our key idea is to combine a versatile and powerful pre-trained model with fast, instance-specific adaptation. To this end, our Instance-Adaptive Fine-Tuning (IAFT) strategy updates only a minimal subset of the network's parameters (the final linear layers), leaving the vast majority untouched. This lightweight procedure enables AnyPcc to converge to an effective instance-specific model within seconds which facilitated by the robust priors of our UCM. This approach transcends the slow encoding of traditional INRs and the generalization limits of a fixed pre-trained model, achieving an outstanding balance between compression performance, generalization, and practical efficiency. To validate this, we curated a comprehensive training corpus and established a demanding evaluation benchmark that extends far beyond standard test sets, providing a rigorous measure of its real-world viability, as shown in Figure \ref{fig:overview}.
Our contributions are summarized as follows:

\begin{itemize}

\item We introduce \textbf{AnyPcc}, a universal compression framework designed to resolve the critical trade-off between efficiency and generalization. It is the first method to achieve high compression and robust performance across diverse point cloud types using \textbf{a single, unified model}.

\item Our \textbf{Universal Context Model (UCM)} resolves a key limitation of prior works by being the first to synergistically integrate fine-grained channel priors with coarse-grained spatial priors, achieving robust context modeling across the entire density spectrum.

\item We pioneer an \textbf{Instance-Adaptive Fine-Tuning (IAFT)} strategy that resolves the trade-off between explicit and implicit compression. It rapidly fine-tunes a small subset of a pre-trained model's weights per-instance, yielding a highly efficient compression model in seconds.

\item Extensive experiments on a benchmark of \textbf{15 diverse datasets} demonstrate that AnyPcc not only outperforms the latest G-PCC v23 standard but also sets a new state-of-the-art among learning-based methods, establishing a new paradigm for universal point cloud compression.

\end{itemize}
\section{Related Work}
\label{sec:related_work}
\subsection{Point Cloud Geometry Compression}
\paragraph{Category-Specific Methods.}
In the domain of dense object point cloud compression \cite{wu2024geometric,you2024pointsoup,pang2022grasp,xia2023learning,xu2024fast,msvoxel,gao2025deep,gao2025dpcset}, several methods leverage sparse tensor representations. For instance, PCGCv2 \cite{pcgcv2}, SparsePCGC \cite{sparsepcgc}, UniPCGC \cite{unipcgc} and  methods \cite{liu2025voxel,zhang2025adadpcc,yu2025hierarchical,akhtar2024inter,pan2024patchdpcc,zheng2024viewpcgc,fan2022d}  employ sparse convolutions to exploit local correlations for effective compression. Concurrently, approaches like VoxelDNN \cite{voxeldnn} and SparseVoxelDNN \cite{sparsevoxeldnn} achieve higher compression efficiency through autoregressive codecs for feature extraction. For LiDAR point clouds compression \cite{yu2025re,sun2025lpcm,cui2025gaem,que2021voxelcontext,cui2023octformer,abbasi2022lidar,luo2024scp,wiesmann2021deep,jin2024ecm}, seminal works such as MuSCLE \cite{biswas2020muscle}, Octsqueeze \cite{huang2020octsqueeze}, OctAttention \cite{fu2022octattention}, EHEM \cite{song2023efficient} and TopNet \cite{topnet} utilize octree-based \cite{schnabel2006octree} representations, employing attention mechanisms to predict the occupancy codes of parent nodes efficiently. Other methods, including Unicorn \cite{unicorn} and RENO~\cite{you2025reno}, also adapt the sparse tensor paradigm to LiDAR data. More recently, with the rise of 3D Gaussian Splatting \cite{kerbl20233d,lu2024scaffold,lin2025vgd} (3DGS), GausPcgc \cite{wang2025novel} is introduced, which enables efficient compression of 3DGS coordinates. Nevertheless, a unifying drawback common to all of these approaches is their architectural over-specialization for particular point cloud attributes, which in turn severely compromises their generalizability.

\paragraph{The Generalization Problem.}
Despite recent efforts toward versatility from frameworks like Unicorn~\cite{unicorn} and standardization bodies (MPEG~\cite{pcc_ctc,pcc_tmapv1}, AVS, JPEG~\cite{guarda2025jpeg} AI-PCC), a critical generalization problem persists. Unicorn-U \cite{unicorn} serves as a prime example, exhibiting two critical flaws: it relies on a non-unified, hybrid architecture (e.g., attention and convolution)  that compromises practical utility, and its performance collapses on out-of-distribution (OOD) data.
This dependency on curated training data is untenable for the vast and varied landscape of real-world point clouds. Dedicated training data is often unavailable for many critical types, such as medical scans, 3D Gaussian Splats, or point clouds generated by networks like Dust3R~\cite{wang2024dust3r} and VGGT~\cite{wang2025vggt}. Consequently, the compression efficiency of existing methods degrades drastically on such OOD data. This performance collapse reveals a critical, yet largely unaddressed, challenge in the current research landscape: \textit{\textbf{achieving true generalization capability in point cloud compression networks.}}

\subsection{Implicit Compression}
\begin{figure*}
    \centering
    \includegraphics[width=1\linewidth]{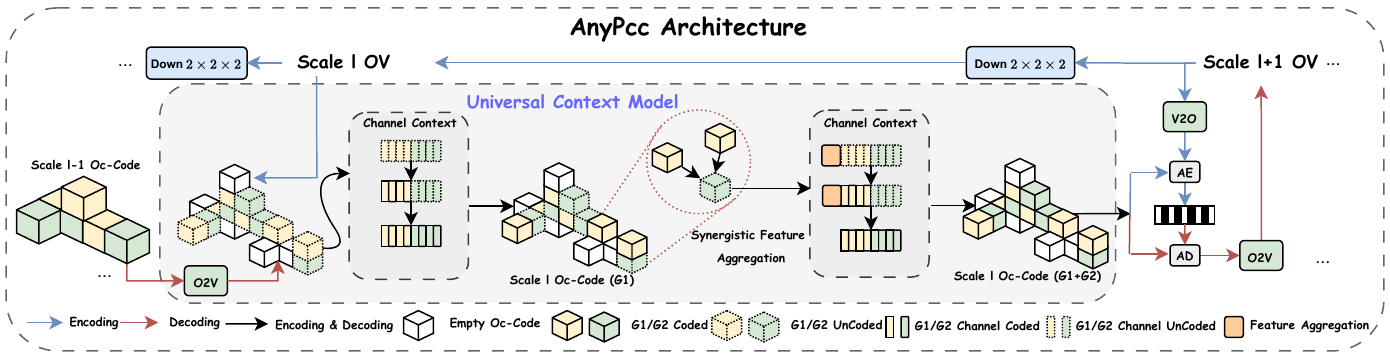}
    \caption{llustration of the proposed AnyPcc framework. AnyPcc formulates the compression task as a sequential prediction of multi-scale occupancy codes. It develops a \textbf{Universal Context Model (UCM)} that incorporates strong inductive biases from spatio-channel partitioning, enabling robust generalization across diverse data sources.}
    \label{fig:anypcc}
\end{figure*}

Implicit neural representations (INRs) offer strong generalization by overfitting a coordinate-based network (e.g., an MLP) to a single data instance, using the optimized weights as the compressed representation~\cite{dupont2021coin,zhang2024gaussianimage,dupont2022coin++,chen2021nerv,inr-pcc,xue2024neri,van2021overfitting,mao2023last,zhang2025efficient}. However, this per-instance optimization is prohibitively slow, as it requires full network training, and the process is often lossy.

Some works have sought to mitigate these issues. For instance, while LINR-PCGC~\cite{huang2025linr} achieves lossless compression by overfitting a base model, it requires full network tuning and is limited to multi-frame scenarios due to the impractical storage overhead of per-frame weights. In image and video compression, recent works~\cite{lv2023dynamic,tsubota2023universal,oh2024parameter} have employed parameter-efficient fine-tuning (PEFT) techniques~\cite{wang2020k,hu2022lora} to adaptively fine-tune pre-trained models. This hybrid paradigm, which synergizes the robust priors of a pre-trained model with the flexibility of implicit, per-instance adaptation, remains unexplored in point cloud compression. The highly irregular structure of point clouds further necessitates a dedicated investigation into designing an effective explicit-implicit model tailored for this domain.

\section{Method}
\label{sec:method}
\subsection{Overview}
Our proposed framework, AnyPcc, achieves universal point cloud compression through three core components. First, we introduce a \textbf{Universal Context Model (UCM)}, pre-trained on diverse data to adaptively handle point clouds of varying densities. Second, we employ an \textbf{Instance-Adaptive Fine-tuning} strategy to efficiently compress out-of-distribution (OOD) samples. Finally, we demonstrate how our framework can be seamlessly extended from lossless to lossy compression via a probability thresholding mechanism, \textbf{creating a single, unified solution}. These components are detailed in the following subsections.

\subsection{Universal Context Model}

\paragraph{Design Insight.}
The task of point cloud geometry compression can be viewed as predicting occupancy probabilities, where a more accurate model yields a shorter bitstream by arithmetic coding. 
While prior work has leveraged priors effectively, it has focused exclusively on fine-grained information. 
While spatial context models~\cite{sparsepcgc,unipcgc} focus on the voxel scale and channel-wise context models~\cite{you2025reno} on the occupancy code scale, they share a fundamental limitation. Both overlook the coarse-grained structural context available from the spatial relationships between the occupancy codes themselves.
Our UCM resolves this by introducing a synergistic partitioning scheme that is the first to simultaneously model fine-grained dependencies via channel grouping and coarse-grained context via spatial grouping. This unified approach creates a model robust across the entire density spectrum, as shown in Figure~\ref{fig:context} and~\ref{fig:anypcc}.

\begin{figure*}
    \centering
    \includegraphics[width=1\linewidth]{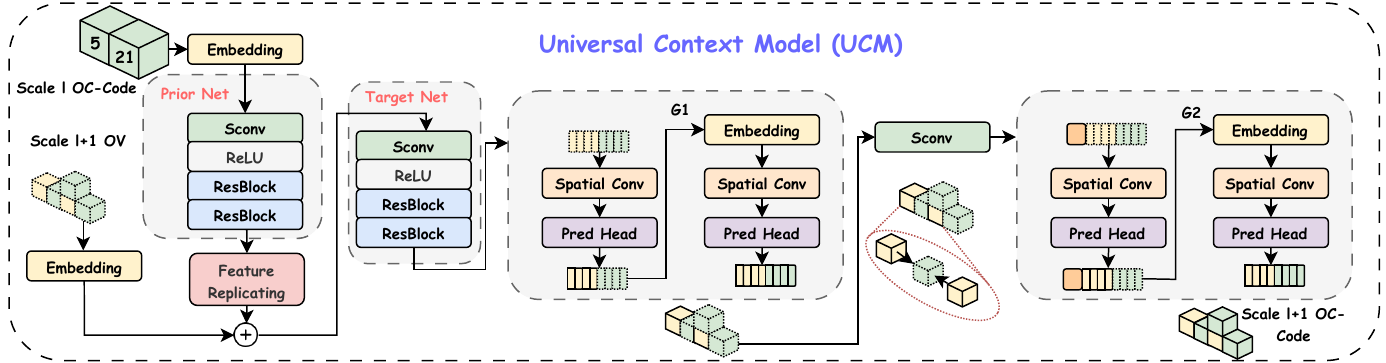}
    \caption{An illustration of our recursive, parameter-shared Universal Context Model (UCM) operating at scale $l+1$. At each scale, the UCM processes the input voxels to predict the probability distribution of the corresponding occupancy codes. The same model architecture and parameters are recursively applied to all other scales, ensuring a consistent prediction process throughout the hierarchy.}
    \label{fig:ucm}
\end{figure*}

Our spatio-channel design is grounded in two theoretical principles, which we prove in \suppl~\ref{sec:suppl_proofs}. We first establish that modeling on 8-bit occupancy code channels is information-theoretically equivalent to modeling on the corresponding $2\times2\times2$ voxel block (Theorem 1). We then prove this provides a significant receptive field advantage crucial for sparse data (Theorem 2). 
Leveraging these principles, the UCM employs a hierarchical framework over a multi-scale occupancy code representation. The prediction proceeds scale-by-scale, from coarse to fine, where the context at each scale is drawn from three sources: the occupancy codes of the parent scale, fine-grained channel-wise priors, and coarse-grained spatial priors. Critically, unlike in image models~\cite{he2021checkerboard,he2022elic,jiang2023mlic}, our context grouping operates directly on geometric occupancy codes instead of a latent space, creating a true coarse-to-fine geometric partitioning. This design ensures robust prediction, as detailed in Figure~\ref{fig:ucm}. (More in \suppl~\ref{sec:lossless_compression})

\paragraph{Notation.}
V2O/O2V denote the Voxel-to-Occupancy-Code conversion and vice-versa (Fig.~\ref{fig:context}). Oc-Code is an occupancy code and OV is an occupied voxel. The Arithmetic Encoder (AE) compresses Oc-Codes to a bitstream using predicted probabilities and the Arithmetic Decoder (AD) reconstructs them from the bitstream.

\paragraph{Hierarchical Context Propagation.}
Our model adopts a coarse-to-fine hierarchical structure. At each scale $l$, the geometric information is captured by a set of occupied voxel coordinates $\mathcal{C}^{(l)}$ and their associated 8-bit occupancy codes $\{o_i^{(l)}\}$.
The context propagation mechanism, involves two main stages. First, a coarse-context encoding network, $\Psi_{\text{prior}}$, processes the occupancy codes and coordinates from scale $l$ to generate a powerful latent representation $\mathbf{Z}^{(l)}$. This network implicitly handles the conversion from discrete codes to continuous features and aggregates local contextual information:
\begin{equation}
    \mathbf{Z}^{(l)} = \Psi_{\text{prior}}( \{o_i^{(l)}\}, \mathcal{C}^{(l)} ).
\end{equation}
Second, this latent representation is propagated to the finer resolution scale $l+1$. The features are upsampled and then refined by a target-processing network $\Psi_{\text{target}}$ to produce the final predictive context $\{\mathbf{c}_j^{(l+1)}\}$ at the finer scale:
\begin{equation}
    \{\mathbf{c}_i^{(l+1)}\} = \Psi_{\text{target}}(\text{Upsample}(\mathbf{Z}^{(l)})),
\end{equation}
where $\Psi_\text{target}$ and $\Psi_\text{prior}$ are shown in Figure \ref{fig:ucm}.
The resulting context field $\{\mathbf{c}_i^{(l+1)}\}$ provides the foundation for our detailed spatio-channel prediction task.

\paragraph{Spatio-Channel Context Factorization.}
After aggregating information from a coarser scale $l$, the compression task at the current scale $l+1$ is to predict its occupancy codes. To efficiently model this, we decompose the joint probability distribution of the occupancy codes by factorizing their context along both spatial and channel-wise dimensions. This dual factorization dramatically reduces predictive complexity and improves coding efficiency.

\paragraph{Spatial Context Partitioning.}
To model the coarse-grained structural priors, we first spatially partition the target occupancy codes at scale $l+1$ into two disjoint sets using a 3D checkerboard pattern, as shown in Figure \ref{fig:anypcc}. An occupancy code at location $(x, y, z)$ is assigned to group $\mathcal{G}_1$ (yellow cells) if the coordinate sum $(x+y+z)$ is even, and to group $\mathcal{G}_2$ (green cells) if it is odd:
\begin{align}
    \mathcal{G}_1 &= \{i \in \mathcal{V}^{(l+1)} \mid (x_i+y_i+z_i) \pmod 2 = 0\}, \\
    \mathcal{G}_2 &= \{i \in \mathcal{V}^{(l+1)} \mid (x_i+y_i+z_i) \pmod 2 = 1\}.
\end{align}
This partitioning strategy transforms the prediction task into a two-step auto-regressive process. The joint probability of all $\{o_i\}$ at the current scale $l+1$ is factorized as:
\begin{equation}
    P(\{o_i\}) = \prod_{i \in \mathcal{G}_1} P(o_i | \mathbf{c}_i) \cdot \prod_{j \in \mathcal{G}_2} P(o_j | \mathbf{c}_j, \{o_k\}_{k \in \mathcal{N}(j) \cap \mathcal{G}_1}),
\end{equation}
where $\mathbf{c}_i$ and $\mathbf{c}_j$ are the context vectors for locations $i \in \mathcal{G}_1$ and $j \in \mathcal{G}_2$ respectively, drawn from the context field $\{\mathbf{c}_k^{(l+1)}\}$ propagated from the coarser scale $l$. $\mathcal{N}(j)$ is the $k^3$ neighborhood of location $j$. This spatial factorization establishes a powerful dependency model on the coarse-grained context, allowing the network to leverage information from immediately adjacent, already-decoded occupancy codes for more accurate probability estimation.

\paragraph{Channel Context Partitioning.}
Complementary to the coarse-grained spatial factorization, we introduce a channel-wise partitioning to model the fine-grained context within each 8-bit occupancy code $o \in \{0, \dots, 255\}$. The code is decomposed into two 4-bit sub-symbols: a least significant part $o_0$ and a most significant part $o_1$.
\begin{equation}
    o_0 = o \pmod{16}, \quad o_1 = \lfloor o / 16 \rfloor.
\end{equation}
This channel-wise partitioning enables a two-stage prediction cascade. For any location $i$, the probability of its occupancy code $o_i$ is factorized as:
\begin{equation}
    P(o_i | \cdot) = P(o_{i,0} | \cdot) \cdot P(o_{i,1} | o_{i,0}, \cdot),
\end{equation}
where $(\cdot)$ represents the full conditioning context available at location $i$. For instance, for a location $i \in \mathcal{G}_1$, this context is simply $\mathbf{c}_i$. The sub-symbol probabilities are modeled sequentially via a prediction module composed of a Spatial Convolution $\mathcal{S}$ and a Prediction Head $\mathcal{H}$. For the second sub-symbol, the context is updated to $\mathbf{c}'_i = \mathbf{c}_i + \text{Emb}(o_{i,0})$. The probabilities are then given by:
\begin{align}
    P(o_{i,0} | \mathbf{c}_i) &\propto \exp\left(\mathcal{H}_{\text{g1,0}}(\mathcal{S}_{\text{g1,0}}(\mathbf{c}_i))\right), \\
    P(o_{i,1} | o_{i,0}, \mathbf{c}_i) &\propto \exp\left(\mathcal{H}_{\text{g1,1}}(\mathcal{S}_{\text{g1,1}}(\mathbf{c}'_i))\right).
\end{align}
This cascading mechanism improves coding efficiency with refining the context for the second sub-symbol with the first. The process is analogous for $\mathcal{G}_2$, which additionally incorporates spatial context from decoded neighbors in $\mathcal{G}_1$.

\paragraph{Synergistic Context Feature Aggregation.}
The synergy between the spatial partitions is realized when predicting the second group $\mathcal{G}_2$. The initial context for a location $j \in \mathcal{G}_2$ is explicitly enhanced by aggregating feature from its decoded neighbors in $\mathcal{G}_1$, as highlighted in Figure~\ref{fig:anypcc}. This process involves three steps:

First, we create a sparse, augmented context field $\hat{\mathbf{C}}$ populated only at locations corresponding to the already-decoded group $\mathcal{G}_1$. For each location $k \in \mathcal{G}_1$, we augment its initial context $\mathbf{c}_k$ with an embedding of its decoded occupancy code $o_k$:
\begin{equation}
    \hat{\mathbf{c}}_k = \mathbf{c}_k + \text{Emb}_{\text{prior}}(o_k), \quad \forall k \in \mathcal{G}_1.
\end{equation}

Second, this local information is gathered for each location $j \in \mathcal{G}_2$ using a sparse convolution ($\text{Sconv}$). The aggregated context $\mathbf{c}_{\text{agg}, j}$ is computed by applying a weighted sum over the augmented contexts of its neighbors in $\mathcal{G}_1$:
\begin{equation}
    \mathbf{c}_{\text{agg}, j} = \left[ \text{Sconv}(\hat{\mathbf{C}}) \right]_j = \sum_{k \in \mathcal{N}(j) \cap \mathcal{G}_1} \mathbf{W}_{\text{sconv}}(k-j) \cdot \hat{\mathbf{c}}_k.
    \label{eq:sconv_detail}
\end{equation}

Finally, the enhanced context $\mathbf{c}''_j$ for location $j$ is formed by fusing its original context $\mathbf{c}_j$ with the aggregated neighbor context $\mathbf{c}_{\text{agg}, j}$ via a fusion network $\Phi_{\text{fuse}}$:
\begin{equation}
    \mathbf{c}''_j = \Phi_{\text{fuse}}([\mathbf{c}_j, \mathbf{c}_{\text{agg}, j}]),
\end{equation}
where $[\cdot, \cdot]$ denotes concatenation. This process allows the model to directly leverage the decoded geometric structure from $\mathcal{G}_1$ to resolve ambiguities and improve predictions for the subsequent group $\mathcal{G}_2$. After all partitions at scale $l+1$ have been processed, the entire procedure is repeated for the next finer scale $l+2$. 

\subsection{Instance-Adaptive Fine-Tuning}

While the pre-trained UCM provides a strong general prior, its performance will degrade when encountering OOD instances. To bridge this generalization gap, we introduce Instance-Adaptive Fine-Tuning (IAFT), a novel strategy that synergizes the strengths of a general pre-trained model with the instance-adaptive finetuning.

\paragraph{Parameter-Efficient Adaptation.}
The core idea of IAFT is to specialize the UCM to a new instance through parameter-efficient fine-Tuning. To achieve this, we partition the UCM's parameters $\Theta$ into a large, frozen backbone $\Theta_{\text{frozen}}$ and a small, tunable set $\Theta_{\text{tune}}$. The frozen set $\Theta_{\text{frozen}}$ comprises the vast majority of the network, including all feature extraction modules and the spatial convolution operators $\mathcal{S}$. These components provide a robust general prior. Conversely, the tunable set $\Theta_{\text{tune}}$ consists exclusively of the parameters of the lightweight prediction heads $\mathcal{H}$, which are responsible for mapping the final features to occupancy probabilities.

\paragraph{On-the-Fly Optimization.}
For each new point cloud instance, we perform a rapid, on-the-fly optimization to find a specialized set of weights $\Theta_{\text{tune}}^*$ that minimizes an instance-specific loss function. This process is formulated as:
\begin{equation}
    \Theta_{\text{tune}}^* = \arg\min_{\Theta_{\text{tune}}} \mathcal{L}_{\text{tune}},
\end{equation}
where the loss $\mathcal{L}_{\text{tune}}$ is the negative log-likelihood of the instance's occupancy codes $\{o_i\}$, which directly models the bitrate, regularized by an L1 term to promote sparsity:
\begin{align}
    \mathcal{L}_{\text{tune}} &= -\sum_{i} \log P(o_i | \text{context}_i; \Theta) + \lambda_{\text{L1}} \|\Theta_{\text{tune}}\|_1.
    \label{eq:tune_loss}
\end{align}
To make this adaptation highly efficient, we employ a caching mechanism. We first perform a single forward pass through the frozen backbone $\Theta_{\text{frozen}}$ to compute and cache the inputs to the prediction heads. The subsequent optimization iterations then operate solely on these cached features to rapidly update the small $\Theta_{\text{tune}}$ set. Thanks to the powerful priors from the backbone, this process converges in just a few hundred gradient steps (mere seconds) and yielding a highly accurate, instance-specific predict model.

\paragraph{Encoding and Decoding.}
The final compressed bitstream is composed of two components:
\begin{itemize}
    \item \textbf{Model Component ($\mathcal{B}_{\text{weights}}$):} We employ uniform scalar quantization with a step size $\Delta = 2^{-0.5 \times 16}$ to the weights $\Theta_{\text{tune}}^*$ , followed by \texttt{DeepCABAC} coding~\cite{wiedemann2020deepcabac}.
    \item \textbf{Geometry Component ($\mathcal{B}_{\text{geom}}$):} The occupancy codes $\{o_i\}$ are losslessly compressed by an arithmetic coder guided by the probability distributions from the now-specialized UCM.
\end{itemize}
The total bitrate is the sum of the lengths of these two components. The decoding process is symmetric: the decoder first reads $\mathcal{B}_{\text{weights}}$ to reconstruct the specialized prediction heads, updates its local copy of the UCM, and then uses this instance-specific model to entropy-decode the geometry from $\mathcal{B}_{\text{geom}}$. By transmitting a small model update, the significant bitrate reduction in encoding the geometry far outweighs the overhead of sending the weights, leading to substantial gains in coding efficiency and generalization.

\subsection{Unified Lossless and Lossy Compression}

Although our UCM framework is primarily designed for lossless compression, it can be seamlessly extended to lossy scenarios. For sparse LiDAR point clouds, a straightforward lossy approach is to simply omit the entropy coding of the $n$ finest scales. However, this method would cause severe geometric degradation for dense point clouds. Therefore, We employ a more sophisticated strategy for dense data: the encoder transmits only the ground-truth point count ($k$) for a given scale. The decoder then reconstructs the geometry by identifying the $k$ most probable occupied locations and their corresponding occupancy codes based on the model's predictions. 
Please refer to \suppl~\ref{sec:implementation_lossy} for detailed model specifications and \suppl~\ref{sec:prac_deploy} for an analysis of practical deployment trade-offs.

\section{Experimental Results}
\label{sec:results}
\subsection{Experiment Setup}
\paragraph{Dataset.}
To cultivate a powerful and generalizable Universal Context Model (UCM), we curate a comprehensive training corpus by merging multiple datasets. This corpus includes a wide array of point clouds recommended by the MPEG AI-PCC and AVS AI-PCC working groups, such as KITTI \cite{kitti}, Ford \cite{agarwal2020ford}, 8iVFB \cite{dataset8i}, MVUB \cite{mvub}, and ScanNet \cite{dai2017scannet}. To further diversify the training data and enhance model robustness, we also incorporate the GausPcc-1K \cite{wang2025novel,ling2024dl3dv} and Thuman \cite{thuman} datasets.
To truly challenge our proposed framework and assess its real-world viability, our evaluation protocol goes far beyond standard benchmarks. In addition to all mainstream datasets, we deliberately introduce point clouds from modern reconstruction techniques like VGGT methods and 3D Gaussian Splatting. Furthermore, we synthesize three challenging datasets to simulate common data imperfections: NS (added noise), RS (point dropout), and CS (non-rigid shape deformations). This comprehensive and demanding benchmark provides a more realistic measure of a model's stability and generalization capabilities (see \suppl~\ref{sec:suppl_dataset}).
\begin{table*}[]
\centering
\begin{threeparttable}
\caption{Performance comparison on the AnyPcc Benchmark. The table presents the compression performance of AnyPcc against six methods across 15 diverse datasets, with the best and second-best results highlighted in \colorbox[HTML]{FFC7CE}{\textcolor{black}{red}} and \colorbox[HTML]{FFEB9C}{\textcolor{black}{yellow}} cells.}
\label{table:all_method}
\setlength{\tabcolsep}{5pt} 
\begin{tabular}{c|cc|cccccc|cc} 
\toprule
\rowcolor{headergray}
\textbf{Dataset} & \textbf{Cond\tnote{†}} & \textbf{OOD} & \textbf{RENO} & \textbf{SparsePCGC} & \textbf{Unicorn\tnote{*}} & \textbf{OctAttention} & \textbf{TopNet} & \textbf{GPCC} & \textbf{Ours} & \textbf{Ours-U} \\ 
\midrule
\textbf{8iVFB}        & \multirow{7}{*}{E} & \xmark & 0.70         & 0.57     & 0.57 & 0.68 & 0.59 & 0.76             & \cellcolor[HTML]{FFC7CE}{0.54}  & \cellcolor[HTML]{FFEB9C}{0.57}       \\
\textbf{MVUB}         &                       & \xmark & 1.00         & \cellcolor[HTML]{FFEB9C}{0.69}  & \cellcolor[HTML]{FFEB9C}{0.69} & 0.76 &  \cellcolor[HTML]{FFEB9C}{0.69}            & 0.94             & \cellcolor[HTML]{FFC7CE}{0.67}    & 0.75     \\
\textbf{Owlii}        &                       & \xmark & 0.59         & 0.48   & 0.48  & 0.62 &    0.56       & 0.59              & \cellcolor[HTML]{FFC7CE}{0.47}  & \cellcolor[HTML]{FFC7CE}{0.47}       \\
\textbf{Thuman}       &                       & \xmark & 1.64         & 1.70  & 1.70  & 2.31 & 2.20           & 2.00             & \cellcolor[HTML]{FFC7CE}{1.58}  & \cellcolor[HTML]{FFEB9C}{1.64}       \\
\textbf{ScanNet}      &                       & \xmark & 2.15         & \cellcolor[HTML]{FFEB9C}{1.86}  & \cellcolor[HTML]{FFEB9C}{1.86} & 2.13 & 2.03              & 2.03             & \cellcolor[HTML]{FFC7CE}{1.83} & 1.88        \\
\textbf{KITTI}        &                       & \xmark &  7.06        & 6.80 & 6.50 & 7.21 & 6.85             & 8.19              & \cellcolor[HTML]{FFC7CE}{6.18}   & \cellcolor[HTML]{FFEB9C}{6.45}      \\
\textbf{Ford}         &                       & \xmark & 9.38        & 9.77 & \cellcolor[HTML]{FFEB9C}{8.44} & 9.10 & 8.54               & 10.32            & \cellcolor[HTML]{FFC7CE}{8.40}  & 8.57       \\ 
\midrule
\textbf{Dense}           & \multirow{5}{*}{M}& \xmark & 5.81         & 6.37 & 5.48 & 6.55 & 6.38 & \cellcolor[HTML]{FFEB9C}{5.32}      & \cellcolor[HTML]{FFC7CE}{5.27}  & 5.55      \\
\textbf{Sparse}         &                       & \xmark & 9.64         & 9.98  & 9.42 & 10.40 & 10.02              & 9.35             & \cellcolor[HTML]{FFC7CE}{9.11}  & \cellcolor[HTML]{FFEB9C}{9.26}       \\
\textbf{GS}         &                       & \xmark & 13.89         & 15.82  & / & \cellcolor[HTML]{FFEB9C}{11.31} & \cellcolor[HTML]{FFC7CE}{10.95}              & 14.46             & 11.65 & 11.74        \\
\textbf{VGGT}         &                       & \cmark & 8.24         & 7.84   & / & 8.22 & 7.83            & 7.33             & \cellcolor[HTML]{FFEB9C}{7.30} & \cellcolor[HTML]{FFC7CE}{7.06}       \\
\textbf{S3DIS}        &                       & \cmark & 13.06        & 11.88  & / & 11.52 & 10.84             & \cellcolor[HTML]{FFC7CE}{10.66}            & 10.93  & \cellcolor[HTML]{FFEB9C}{10.79}      \\ 
\midrule
\textbf{RS}       & \multirow{3}{*}{H}                  & \cmark & 4.02        & 3.88  & / & 4.05 & 3.92 & 3.72 & \cellcolor[HTML]{FFEB9C}{3.68} & \cellcolor[HTML]{FFC7CE}{3.50}     \\  
\textbf{NS}       &                   & \cmark & 4.96        &  6.54 & / & 4.89 & 4.84 & 4.85 & \cellcolor[HTML]{FFEB9C}{4.69} & \cellcolor[HTML]{FFC7CE}{4.67}     \\  
\textbf{CS}       &                 & \cmark & 3.94        & 4.94  & / & 3.40 & 3.21 & 3.23 & \cellcolor[HTML]{FFEB9C}{3.18} & \cellcolor[HTML]{FFC7CE}{3.08}     \\  \midrule
\multicolumn{3}{c|}{\textbf{CR Gain over GPCC} $\downarrow$} & 2.96\% & 2.07\% & / & 1.32\% & -4.04\% & 0.00\% & \cellcolor[HTML]{FFC7CE}{-11.93\%} & \cellcolor[HTML]{FFEB9C}{-10.75\%} \\
\multicolumn{3}{c|}{\textbf{Enc/Dec Time (s)} $\downarrow$} & 0.22/0.23 & 2.6/2.2 & / & 7.7/1324 & 8.7/1740 & 3.8/2.7 & \multicolumn{2}{c}{2.84/0.46}  \\
\multicolumn{3}{c|}{\textbf{Total Parameters (M)} $\downarrow$} & 9.03 & 26.43 & / & 29.61 & 23.59 & / & 68.39 & 9.77 \\
\bottomrule
\end{tabular}
\begin{tablenotes}
    \item[†] Cond represents the test difficulty of the testsets, and we divide the test set into easy (E), medium (M), and hard (H).
      \item[*] The results for Unicorn are cited directly from the original publication as its implementation is not open-source.
    \end{tablenotes}
\end{threeparttable}
\end{table*}

\paragraph{Implementation.}
Our framework is built with PyTorch and TorchSparse \cite{tangandyang2023torchsparse}, and all experiments are conducted on a single NVIDIA RTX 3090 GPU. We propose and evaluate two distinct versions of our model:
\begin{itemize}
    \item \textbf{Ours:} This version aligns with standard practice by training a \textbf{dedicated model} for each dataset category.
    
    \item \textbf{Ours-U:} This is a single, \textbf{unified model} trained on a large-scale mixed dataset. The same set of weights is applied to all test sets, greatly improving its practical utility.
\end{itemize}
For both versions, we apply our UCM on E samples, and augment it with IAFT (200 iters) for M and H samples.

\paragraph{Baseline and Metrics.}
We benchmark our method against several state-of-the-art (SOTA) open-source solutions, including RENO \cite{you2025reno}, SparsePCGC \cite{sparsepcgc}, OctAttention \cite{fu2022octattention}, and TopNet \cite{topnet}, as well as the latest traditional codec, GPCC v23. Since the code for Unicorn \cite{unicorn} is unavailable, we report its performance from the original paper under aligned experimental settings.
All models are evaluated under identical training and testing conditions for a fair comparison. For lossless compression, we report the efficiency in bits per point (bpp) and the Compression Ratio Gain (CR-Gain). The CR-Gain is calculated relative to an anchor codec as $(\text{bpp}_{\text{method}} - \text{bpp}_{\text{anchor}}) / \text{bpp}_{\text{anchor}} \times 100\%$, where a more negative value indicates greater bitrate savings. For lossy compression, we assess the rate-distortion performance using bpp for the rate and Peak Signal-to-Noise Ratio (PSNR) for the distortion.
\subsection{Lossless Compression}

\paragraph{Results.}
As benchmarked in Table~\ref{table:all_method} across 15 datasets (including 5 for Out-of-Distribution generalization), our methods demonstrate clear superiority. Our models, \textbf{Ours} and \textbf{Ours-U}, achieve SOTA on 13 datasets and deliver substantial CR-Gain of \textbf{11.93\%} and \textbf{10.75\%} over the GPCC v23 anchor. In stark contrast, most baselines fail to match this anchor: RENO, SparsePCGC, and OctAttention show performance degradations of 2.96\%, 2.07\%, and 1.32\% relative to the anchor, while TopNet is the only competing method to provide a positive gain, at 4.04\%. A key trade-off emerges: our specialized model excels on in-distribution data, but our universal model (\textbf{Ours-U}) shows superior generalization across all OOD datasets, affirming its practical value. For OOD evaluation, models trained on KITTI are used.
Moreover, we provide both the results for RENO-U and SparsePCGC-U with unified data training and the more evaluations on dense point clouds in \suppl~\ref{sec:unidata_train}.
\begin{figure*}[t]
    \centering
    \begin{subfigure}{\textwidth}
        \centering
        \includegraphics[width=0.24\textwidth]{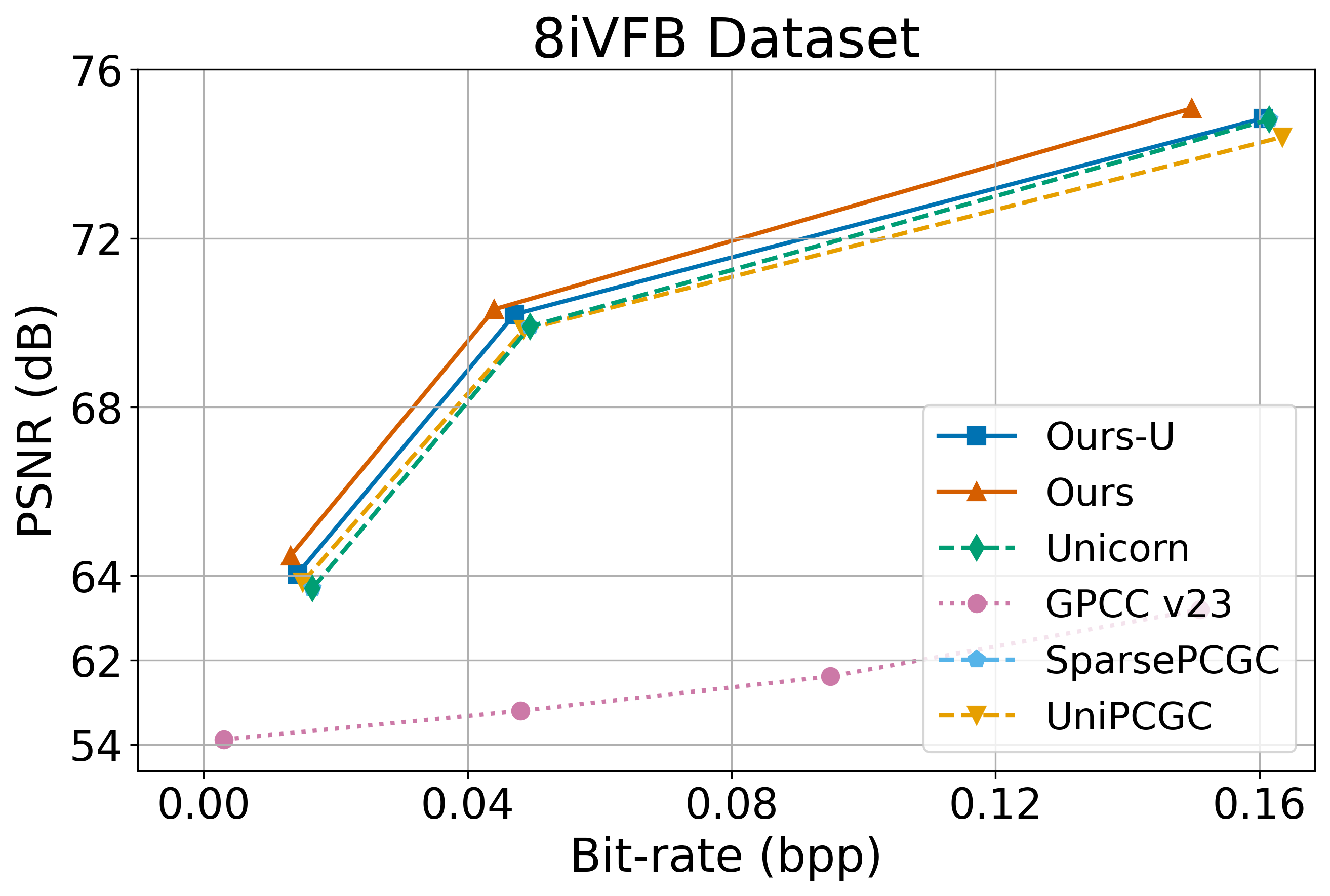}\hfill
        \includegraphics[width=0.24\textwidth]{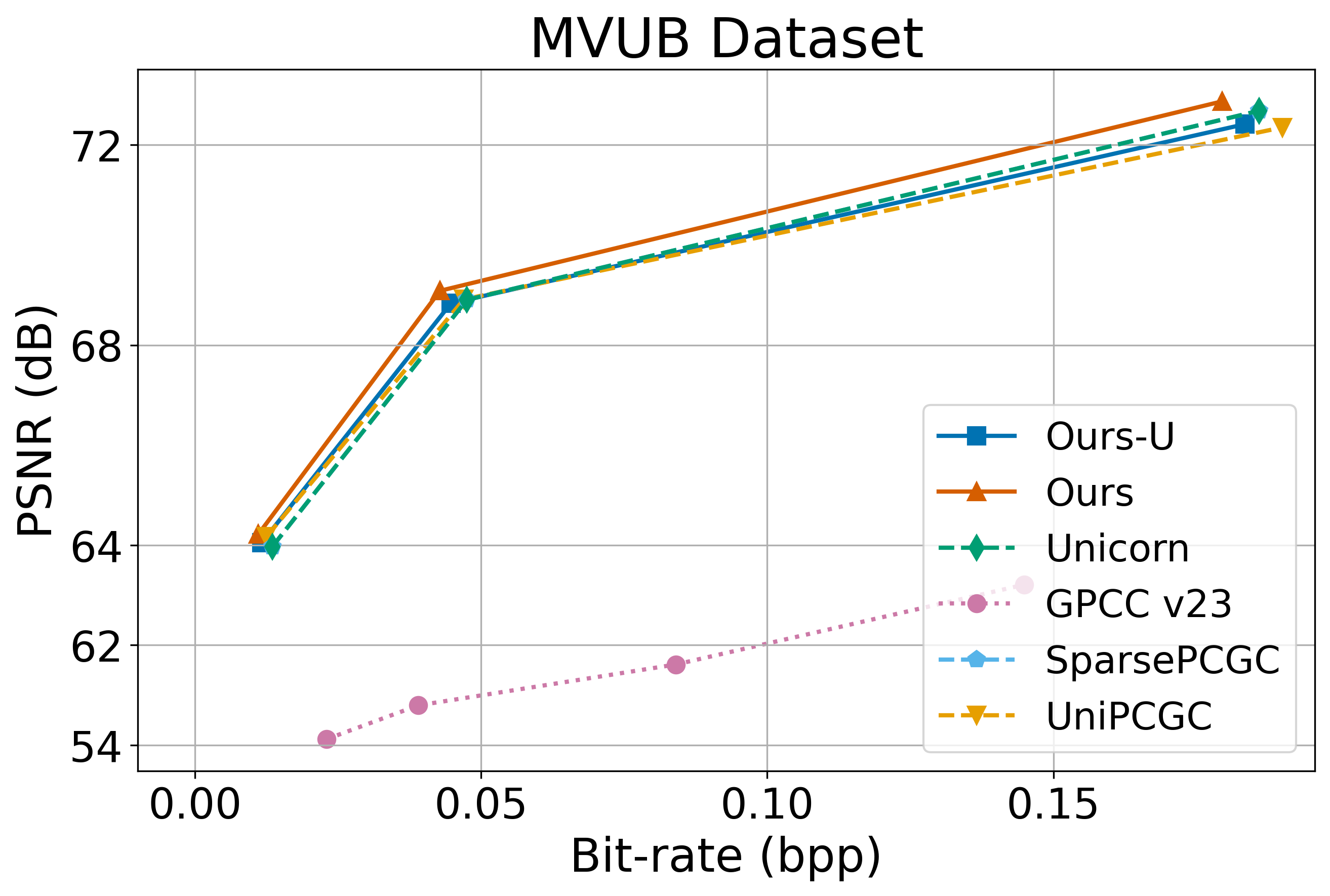}\hfill
        \includegraphics[width=0.24\textwidth]{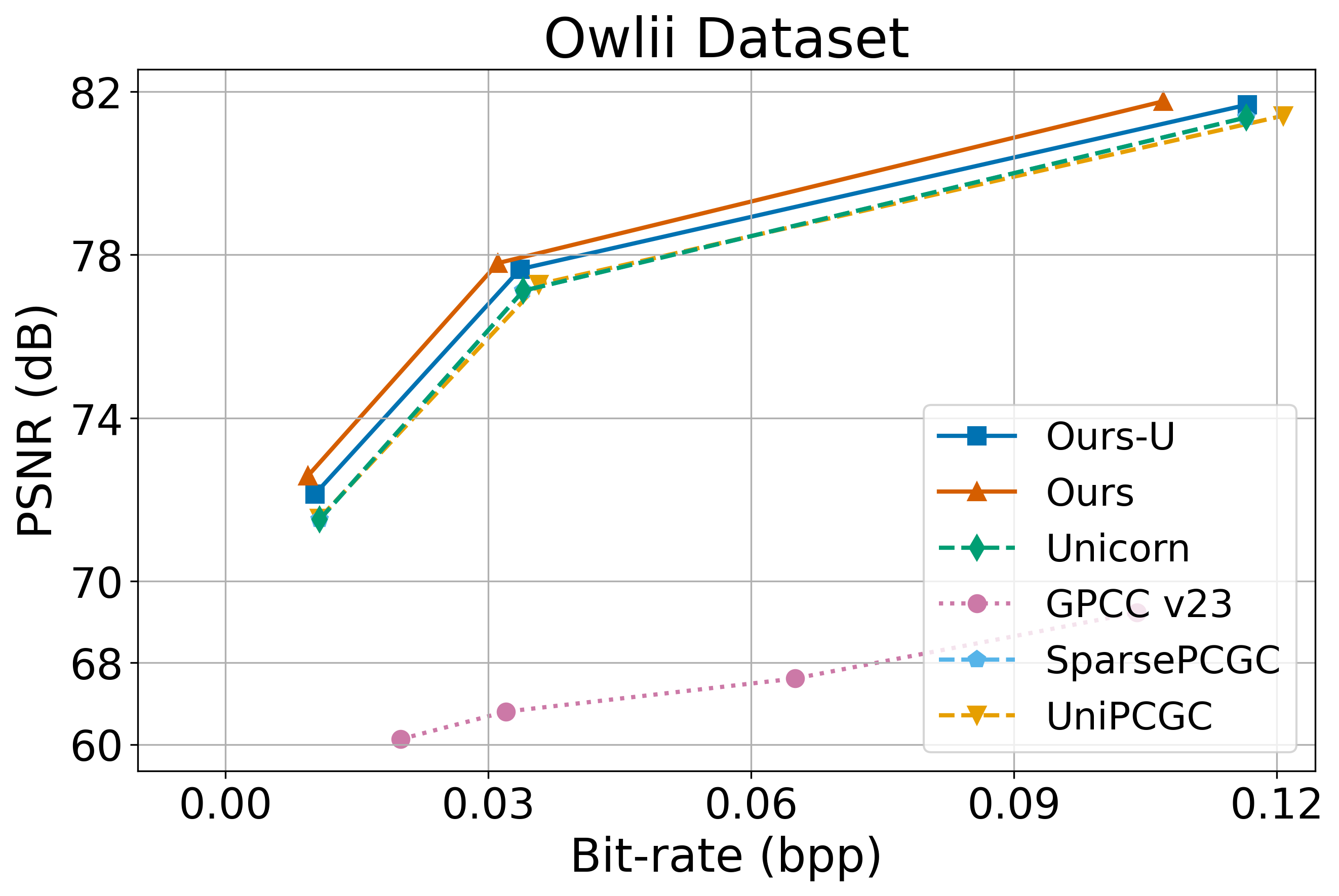}\hfill
        \includegraphics[width=0.24\textwidth]{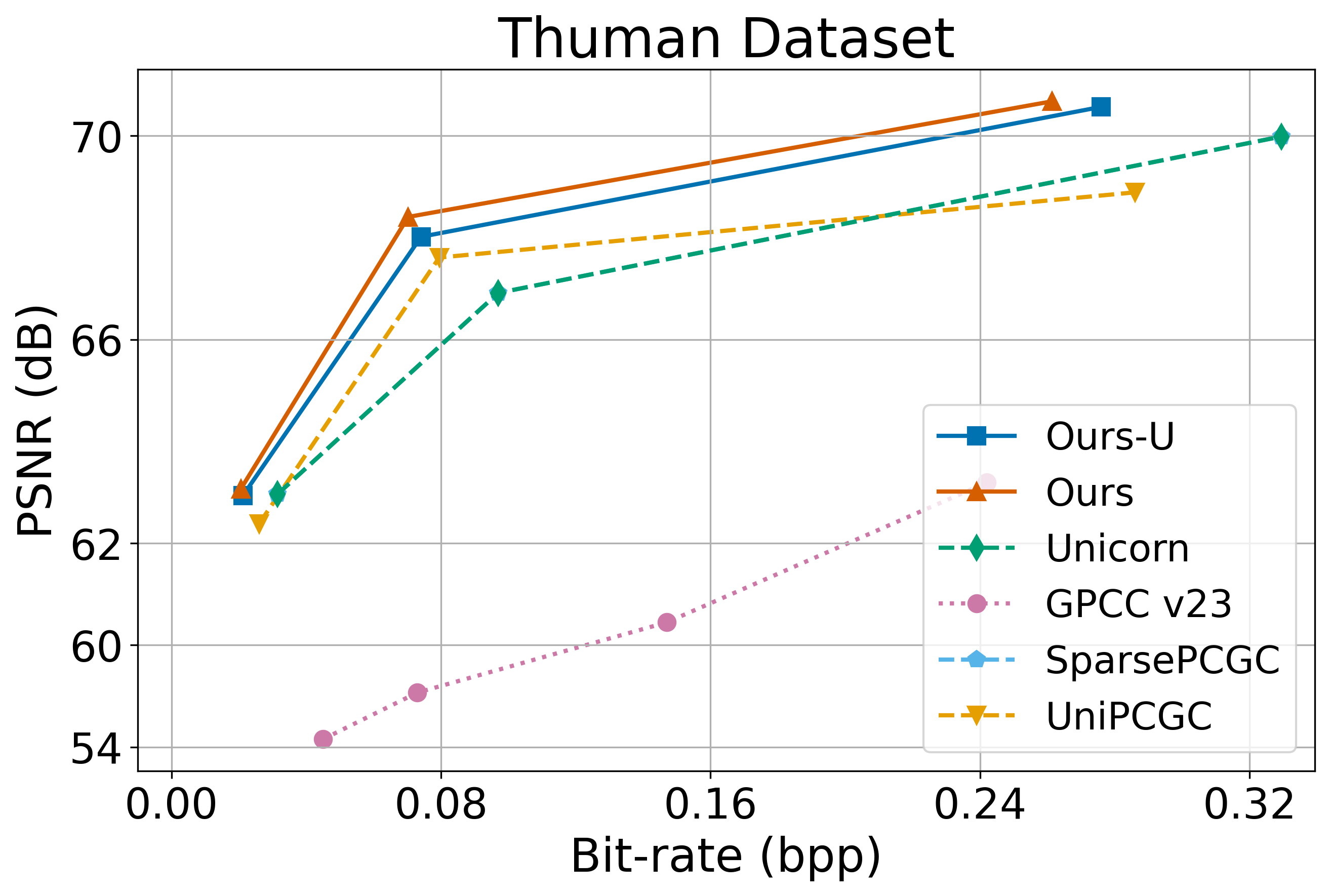}
        \includegraphics[width=0.24\textwidth]{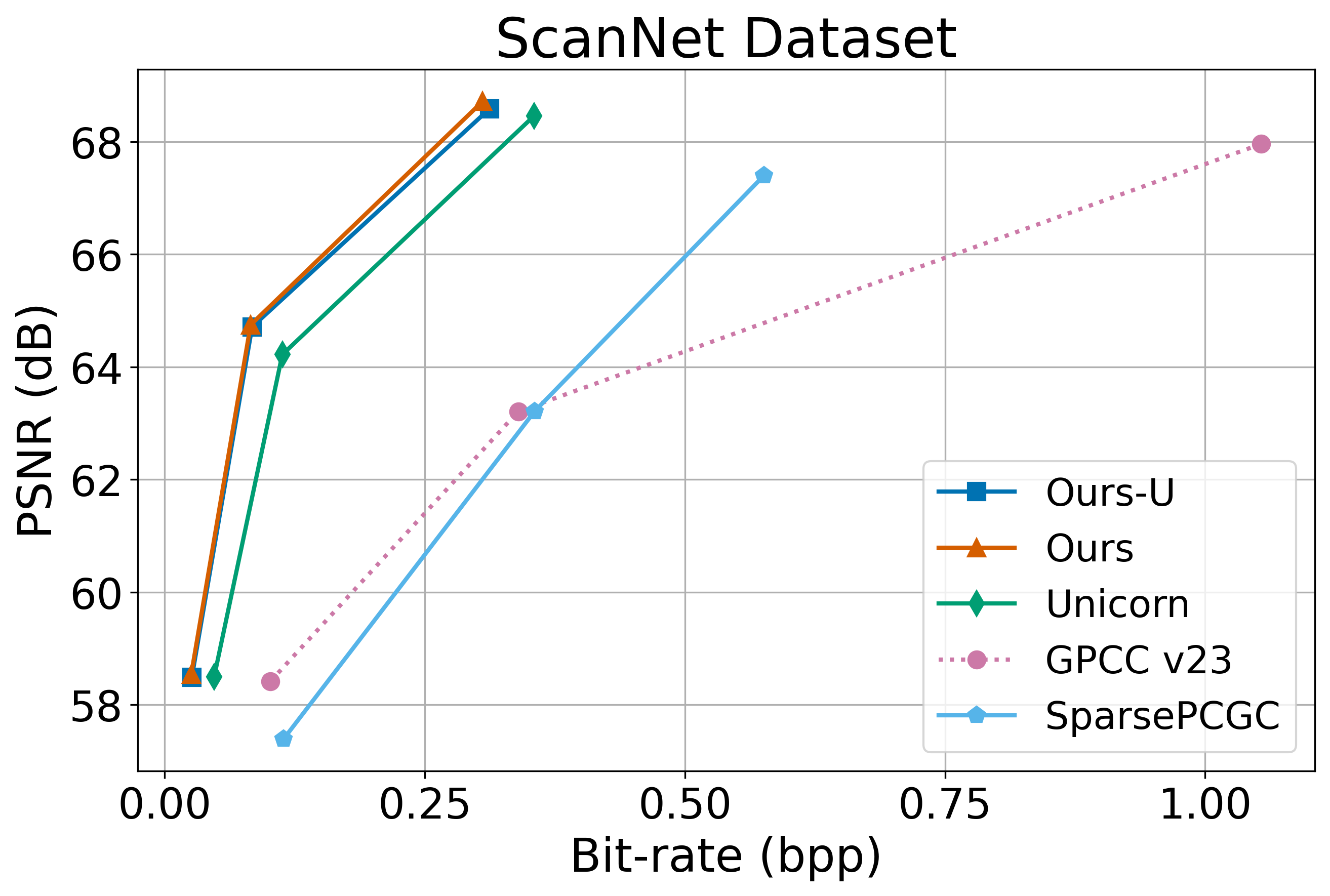}\hfill
        \includegraphics[width=0.24\textwidth]{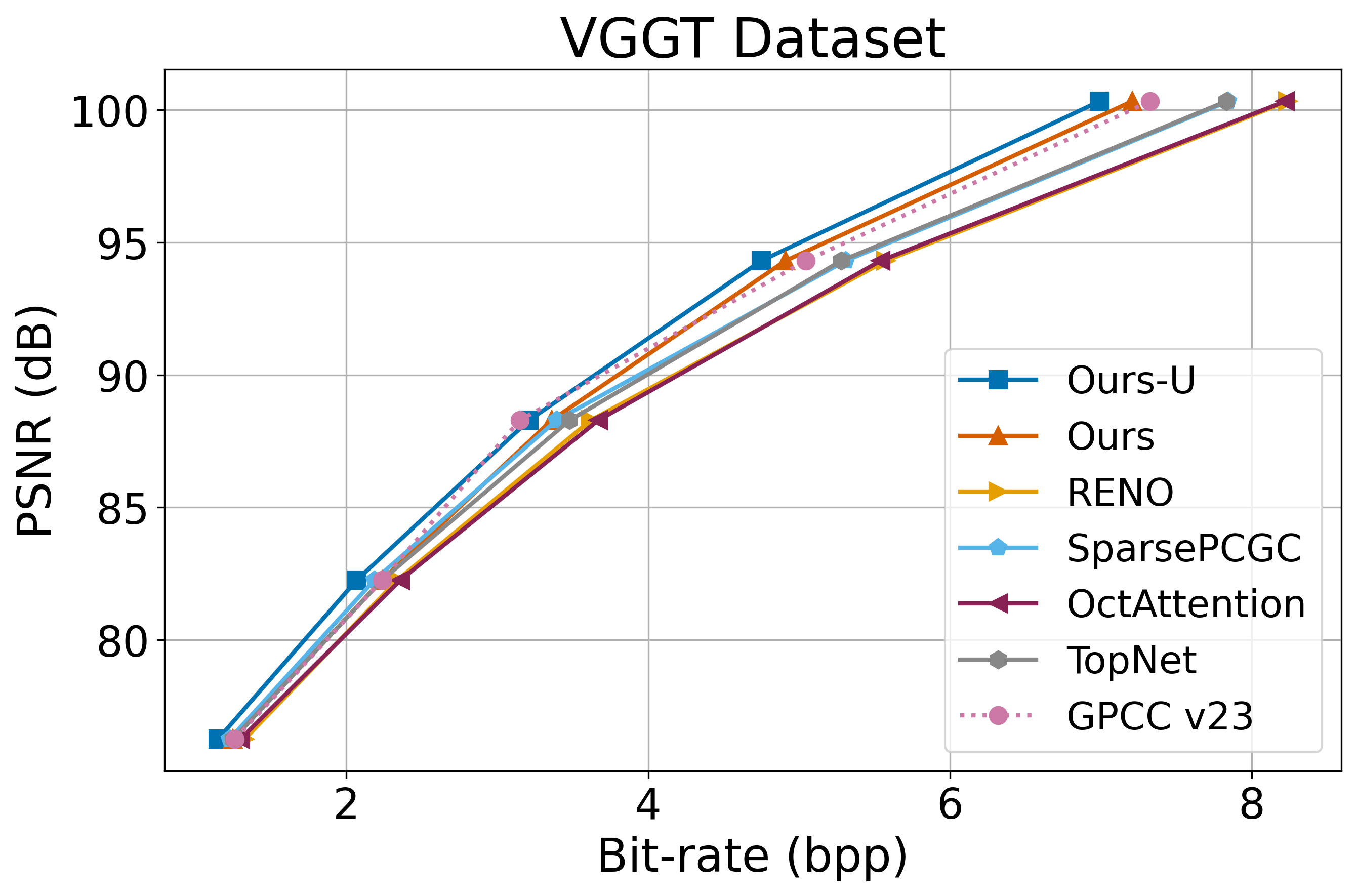}\hfill
        \includegraphics[width=0.24\textwidth]{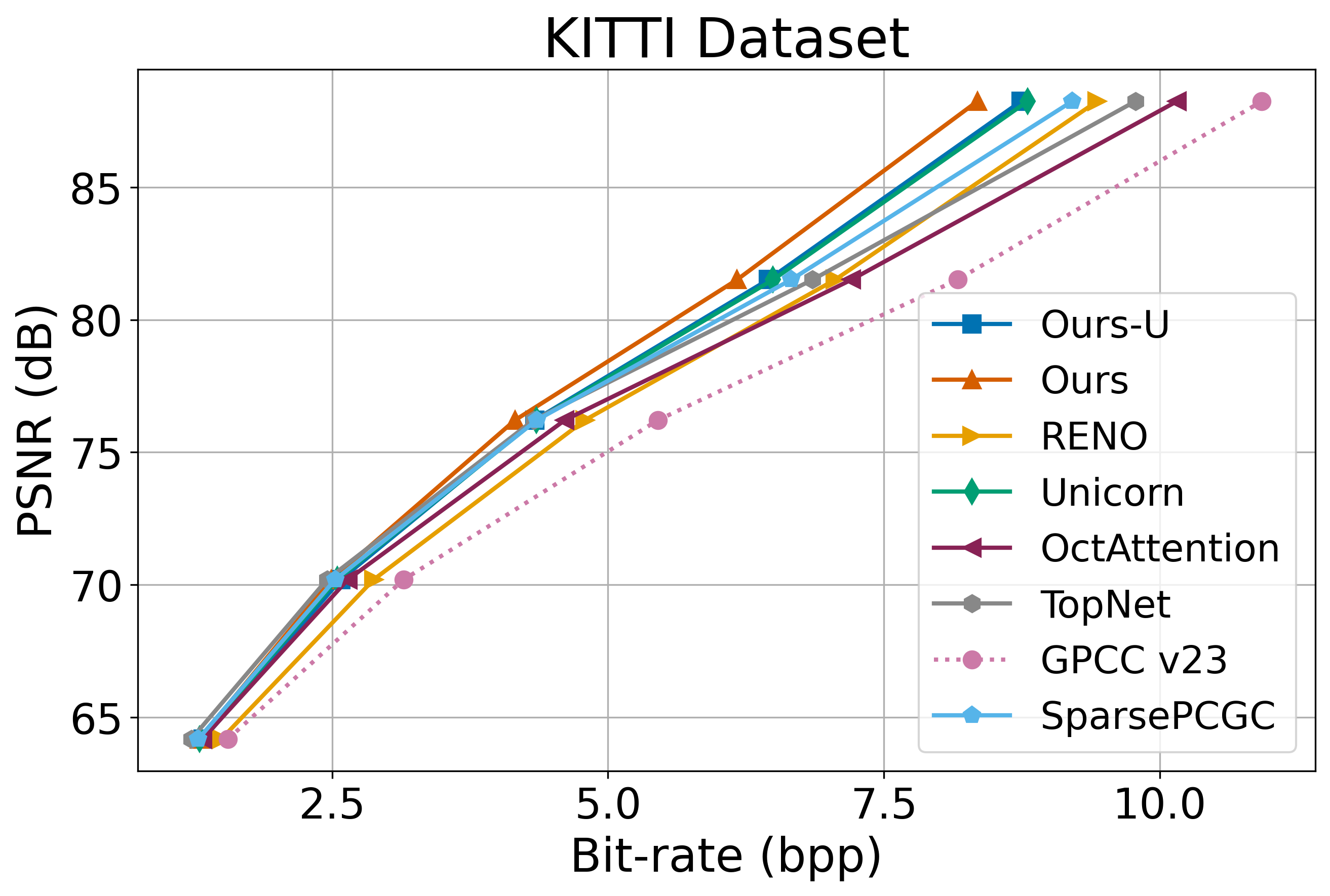}\hfill
        \includegraphics[width=0.24\textwidth]{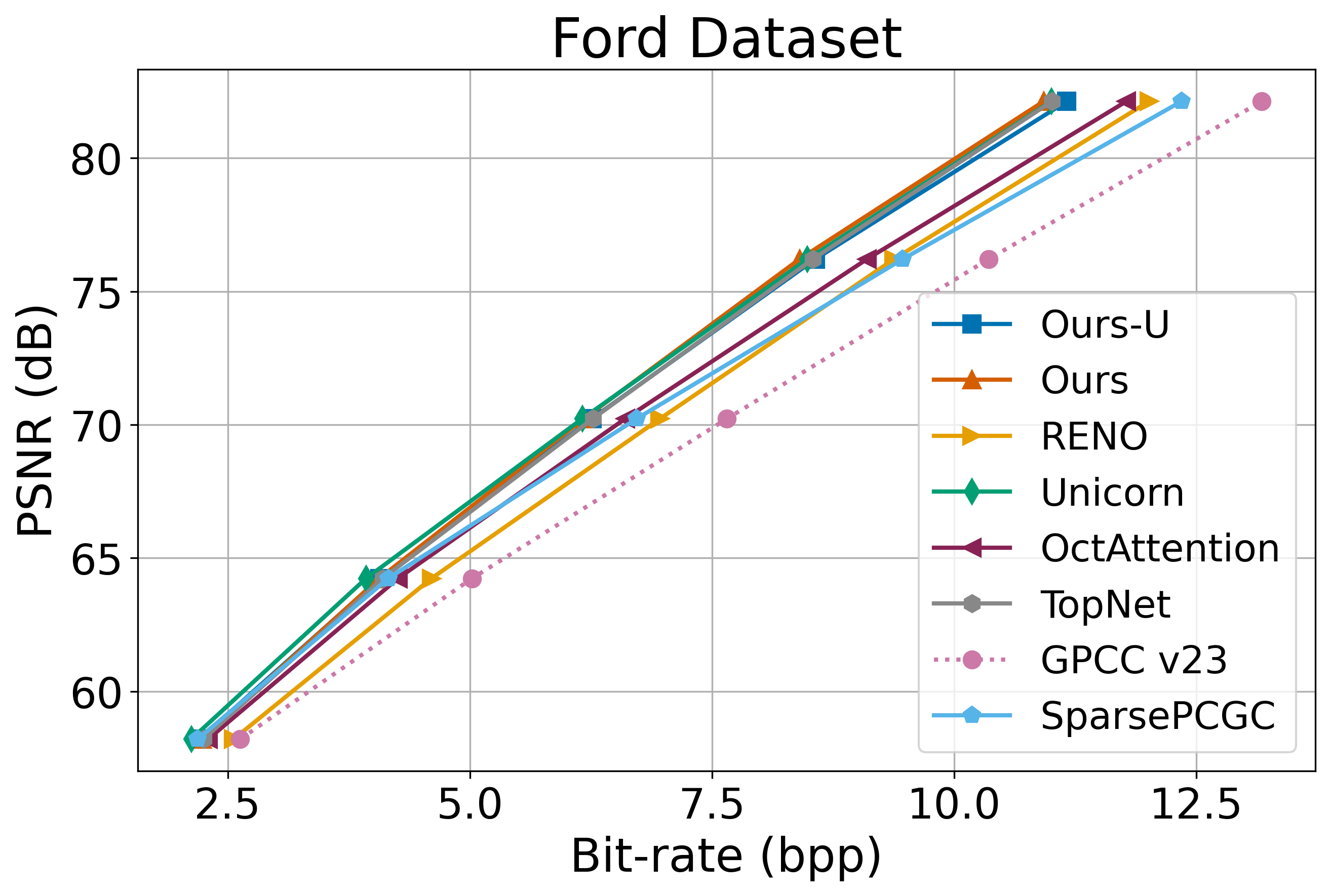}
    \end{subfigure}
    \caption{Performance comparison using rate-distortion curves. Comparisons on the human-body and ScanNet datasets exclude RENO, OctAttention, and TopNet, as they are specifically designed for lossy LiDAR compression and thus incompatible.}
    \label{fig:rdcurves}
\end{figure*}

\paragraph{Model Parameters.} Our primary method (Ours), like all baselines, requires training seven distinct models for the benchmark, as shown in Table \ref{table:all_method}. In contrast, our universal model (Ours-U) utilizes a single set of weights for all datasets. This not only highlights its generalization but also drastically reduces storage and deployment overhead, marking a significant step towards a practical solution.

\paragraph{Codec Time.} Our decoding time is comparable to RENO, the fastest baseline, as shown in Table \ref{table:all_method}. While our default encoding takes 2.84 seconds, it is highly flexible. By adjusting fine-tuning iterations, the encoding time can be controlled within a 0.44s to 2.84s range, offering a trade-off between speed and compression efficiency. A detailed analysis is in the \suppl~\ref{sec:fine-tune time}.

\subsection{Lossy Compression}
The lossy compression results are shown in Figure \ref{fig:rdcurves}. LiDAR-based methods are not shown for the dense human-body and ScanNet datasets due to their suboptimal performance. The results confirm that AnyPcc performs robustly across all datasets. Crucially, the strong performance of our unified model, Ours-U, demonstrates that AnyPcc can serve as a single-model solution for both efficient lossless and lossy point cloud compression. For an analysis of model parameters and codec times in the lossy compression, please refer to the \suppl~\ref{sec:lossy compression res}.

\subsection{Ablation Studies}

\begin{table}[t]
\centering
\caption{Effectiveness analysis of key modules in UCM.}
\label{table:abla_module}
\begin{tabular}{c|ccc|cc}
\toprule
\rowcolor{headergray}
\textbf{} & \textbf{SC}         & \textbf{SG}& \textbf{CG} & \multicolumn{1}{c}{\textbf{CR-Gain}} & \multicolumn{1}{c}{\textbf{Params (M)}} \\ \midrule
\textbf{Baseline}  &       \xmark               &     \xmark          &      \xmark         & 0.00\%                      & 5.15                           \\
\textbf{Abla1}     & \xmark & \cmark             &         \xmark      & -6.56\%                     & 5.68                           \\
\textbf{Abla2}     &       \xmark               &       \xmark        & \cmark             & 0.13\%                      & 5.15                           \\
\textbf{Abla3}     & \cmark                    & \cmark             &  \xmark             & -7.74\%                     & 9.78                           \\
\textbf{Abla4}     & \cmark                    &     \xmark          & \cmark             & -5.33\%                     & 7.19                           \\
\textbf{Abla5}     &       \xmark               & \cmark             & \cmark             & -6.50\%                     & 5.67                           \\
\textbf{Ours}      & \cmark                    & \cmark             & \cmark             & \textbf{-9.88}\%                     & 9.77                           \\ \toprule
\end{tabular}
\end{table}
\paragraph{Key Modules.} 
Table \ref{table:abla_module} details our ablation study on Spatial Convolution (SC), Spatial Grouping (SG), and Channel Grouping (CG). Combining all three yields the optimal performance. Notably, while SC and SG individually enhance results, using CG alone (as in RENO) causes degradation. This indicates that CG's fine-grained priors strictly depend on the coarse-grained spatial priors from SC and SG.

\paragraph{Channel Count.} 
Table \ref{table:abla_merged} shows the effect of varying the channel count (32 / 64 / 128) on performance and complexity. While higher channel counts lead to better performance, the model with 128 channels proves to be too computationally expensive. We therefore choose C=64 for our model, as it offers the best trade-off between performance and speed.

\paragraph{IAFT Module.} 
We also conduct an ablation study on the Instance-Adaptive Fine-Tuning (IAFT) module, with the results presented in Table \ref{table:abla_merged}. The table shows that fine-tuning for 800 iterations yields a performance gain of approximately 5\%, at the cost of an additional 10 seconds in encoding time. 
This allows for dynamic control over the complexity-performance trade-off. 
As shown in Table \ref{table:abla_merged}, IAFT reduces the entropy-coded bpp by 1.883 on the GS dataset, at the cost of 0.319 bpp for the model. This confirms that for samples with irregular distributions, the bitrate savings from entropy coding substantially outweigh the network overhead.

\begin{table}[t]
    \centering
    \caption{Ablation study of channel count and IAFT module, including Bpp composition on the GS dataset.}
    \label{table:abla_merged}
    \setlength{\tabcolsep}{6pt} 
    \small
    \begin{tabular}{ccc}
        \toprule
        \rowcolor{black!10} 
        \textbf{Metrics}           & \textbf{UCM} & \textbf{UCM+IAFT} \\ \midrule
        
        \multicolumn{3}{c}{\textit{Overall Performance (32/64/128 channels)}} \\ \midrule
        \textbf{Bpp}               & 5.42/5.32/5.25         & 5.14/5.04/4.99              \\
        \textbf{Enc Time (s)}      & 0.31/0.44/1.13         & 11.82/12.11/13.08           \\
        \textbf{Dec Time (s)}      & 0.32/0.46/1.14         & 0.32/0.46/1.14              \\
        \textbf{Model Size (M)}    & \multicolumn{2}{c}{2.45/9.77/39.01}             \\ \midrule
        
        \multicolumn{3}{c}{\textit{Bpp Composition on GS dataset}} \\ \midrule
        \textbf{Entropy Coding Bpp} & 13.307                 & 11.424                     \\
        \textbf{Net Size Bpp}      & 0                      & 0.319                      \\ \bottomrule
    \end{tabular}
\end{table}

\paragraph{More Studies.} 
Further evaluations (e.g., on quantization precision, fine-tuning iterations, training data ratio, context model) and visual comparisons are detailed in \suppl~\ref{sec:contex_model_abla}.
\section{Conclusion}
\label{sec:conclusion}
This paper introduced \textbf{AnyPcc}, a universal framework for geometry point cloud compression designed to efficiently compress point clouds of any type. Powered by a robust generic context model and an instance-adaptive fine-tuning strategy, our approach uniquely combines the advantages of both explicit and implicit compression methods. This allows for effective compression of both in-distribution (ID) and out-of-distribution (OOD) samples. Experimental results on a comprehensive benchmark of 15 datasets show that AnyPcc achieves SOTA compression performance.

\clearpage
\maketitlesupplementary

\section{Theoretical Analysis}
\label{sec:suppl_proofs}


\subsection{Proof of Equivalence between Channel-wise and Spatial Auto-regressive Models}

\textbf{Theorem 1.} \textit{Auto-regressive prediction on the 8 channels (bits) of an occupancy code is information-theoretically equivalent to an auto-regressive prediction on the corresponding 8 sub-voxels in space.}

\begin{proof}
Let $l$ denote a coarse scale and $l+1$ a finer scale in the voxel hierarchy. A single parent voxel at scale $l$, located at coordinate $\mathbf{v}^{(l)} \in \mathbb{Z}^3$, corresponds to a $2 \times 2 \times 2$ block of 8 child voxels at scale $l+1$. The coordinate of a child voxel is given by $\mathbf{v}_{\mathbf{\delta}}^{(l+1)} = 2\mathbf{v}^{(l)} + \mathbf{\delta}$, where the offset $\mathbf{\delta} \in \{0, 1\}^3$.

Let $O(\mathbf{v}^{(l+1)}) \in \{0, 1\}$ be the binary occupancy state of a voxel at scale $l+1$. The 8-bit occupancy code for the parent voxel $\mathbf{v}^{(l)}$, denoted as $o(\mathbf{v}^{(l)})$, encapsulates the occupancy states of its 8 children. We can define a fixed mapping (e.g., lexicographical or Morton order) $\pi: \{0, 1, \dots, 7\} \to \{0, 1\}^3$ that links a bit index $k$ to a spatial offset $\mathbf{\delta}$. The $k$-th bit of the occupancy code is thus defined as:
\begin{equation}
    o_k(\mathbf{v}^{(l)}) = O(\mathbf{v}_{\pi(k)}^{(l+1)}).
    \label{eq:bit_to_voxel}
\end{equation}

A \textbf{spatial auto-regressive model at fine scale $l+1$} factorizes the joint probability of the 8 child voxels' occupancy as a product of conditional probabilities, following the order defined by $\pi$:
\begin{equation}
    P(\{O_k^{(l+1)}\}_{k=0}^7 | \mathcal{C}) = 
    \prod_{k=0}^{7} P(O_k^{(l+1)} | \{O_j^{(l+1)}\}_{j<k}, \mathcal{C}),
    \label{eq:spatial_model_supp}
\end{equation}
where $\mathcal{C}$ represents the external context derived from scale $l$ and neighboring voxels, and we use the shorthand $O_k^{(l+1)} \equiv O(\mathbf{v}_{\pi(k)}^{(l+1)})$.

A \textbf{channel-wise auto-regressive model at coarse scale $l$} similarly factorizes the probability of the 8-bit occupancy code:
\begin{equation}
    P(o(\mathbf{v}^{(l)}) | \mathcal{C}) = \prod_{k=0}^{7} P(o_k(\mathbf{v}^{(l)}) | \{o_j(\mathbf{v}^{(l)})\}_{j<k}, \mathcal{C}).
    \label{eq:channel_model_supp}
\end{equation}

By substituting Equation~\eqref{eq:bit_to_voxel} into Equation~\eqref{eq:channel_model_supp}, we can directly observe that the channel-wise model becomes identical to the spatial model in Equation~\eqref{eq:spatial_model_supp}. The set of previously decoded bits $\{o_j(\mathbf{v}^{(l)})\}_{j<k}$ provides the exact same information as the set of previously decoded voxels $\{O(\mathbf{v}_{\pi(j)}^{(l+1)})\}_{j<k}$. Therefore, the two modeling approaches are mathematically equivalent. All predictive operations for the local $2 \times 2 \times 2$ block can be performed in the channel dimension of the occupancy code.
\end{proof}

\subsection{Proof of Effective Receptive Field Expansion under Sparse Convolutions}

\textbf{Theorem 2.} \textit{A sparse convolution with a kernel of size $k$ over the occupancy code space (scale $l$) achieves an effective receptive field equivalent to that of a sparse convolution with a kernel of size $2k$ applied directly on the finer voxel space (scale $l+1$).}

\begin{figure*}[t]
    \centering
    \includegraphics[width=1\linewidth]{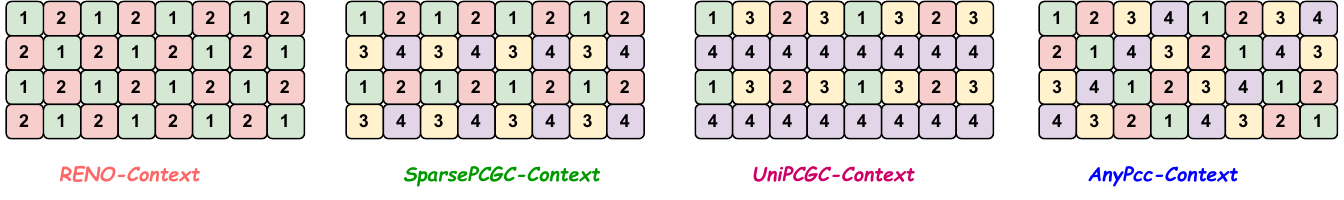}
    \caption{Illustration of our context model in 2D, highlighting its distinction from prior works. The principles shown are directly generalizable to 3D. }
    \label{fig:context_example}
\end{figure*}
\begin{figure}[t]
    \centering
    \includegraphics[width=1\linewidth]{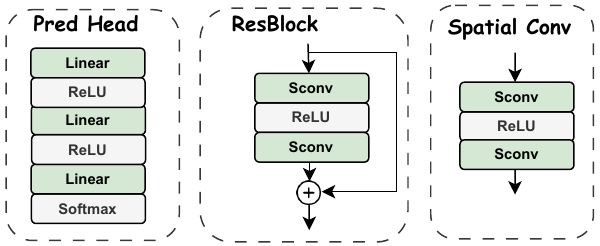}
    \caption{Illustration of the detailed architecture of UCM. }
    \label{fig:detailed}
\end{figure}

\begin{proof}
This analysis is particularly critical in the context of sparse point clouds, where submanifold sparse convolutions are employed. Unlike dense convolutions where the receptive field expands predictably (e.g., two stacked $3 \times 3$ kernels have a $5 \times 5$ field), submanifold convolutions only propagate information between active sites (occupied voxels). In highly sparse scenarios, this can severely restrict information flow, making the effective receptive field much smaller than the theoretical kernel size would suggest. Our proof demonstrates how operating on the occupancy code space mitigates this issue.

For clarity, we analyze a 1D convolution, which generalizes to 3D. Consider a sparse convolution with a kernel of size $k$ applied to features associated with occupancy codes at scale $l$. An output feature at an active site is computed from a neighborhood of up to $k$ active input sites. Let the spatial coordinates of $k$ such active parent voxels at scale $l$ along an axis be $x, x+1, \dots, x+k-1$.

The distance between the centers of the two extreme voxels in this neighborhood is $(x+k-1) - x = k-1$. Each of these parent voxels at scale $l$ maps to a pair of child voxels at scale $l+1$. For a parent at coordinate $x$, its children are at coordinates $2x$ and $2x+1$.

The receptive field in the finer voxel space (scale $l+1$) is the set of all child voxels corresponding to the parent voxels in the kernel's neighborhood. The first parent voxel at $x$ corresponds to child voxels at $\{2x, 2x+1\}$. The last parent voxel at $x+k-1$ corresponds to child voxels at $\{2(x+k-1), 2(x+k-1)+1\}$.

The total span of the receptive field is the maximum possible distance between any two child voxels within this span. This is realized between a child from the first parent voxel and a child from the last. Let their coordinates be $s_{\text{first}}$ and $s_{\text{last}}$.
\begin{align}
    s_{\text{first}} &= 2x + \delta_{\text{first}}, \\
    s_{\text{last}} &= 2(x+k-1) + \delta_{\text{last}},
\end{align}
where $\delta \in \{0, 1\}$. The distance is $|s_{\text{last}} - s_{\text{first}}| = |2(x+k-1) + \delta_{\text{last}} - (2x + \delta_{\text{first}})| = |2k-2 + \delta_{\text{last}} - \delta_{\text{first}}|$.

To maximize this distance, we set $\delta_{\text{last}}=1$ and $\delta_{\text{first}}=0$. The maximum distance is:
\begin{equation}
    D_{\text{max}} = 2k-2 + 1 - 0 = 2k-1.
\end{equation}

Thus, a sparse convolution with kernel size $k$ at the coarse scale $l$ can connect features that are up to $2k-1$ voxels apart at the fine scale $l+1$. A direct convolution at the fine scale would require a kernel of size $2k$ to achieve the same span ($2k-1$), which confirms the theorem. This demonstrates a crucial advantage: by operating on the relatively denser occupancy code representation, our model can establish long-range dependencies that would be broken by extreme sparsity in a direct voxel-level sparse convolution, thereby ensuring a more robust contextual range.
\end{proof}

\section{Implementation Details}

\begin{table*}[h!]
\centering
\caption{IImplementation details for our model configuration, training, and compression strategies.}
\label{tab:implementation_details}
\begin{tabular}{@{}ll@{}}
\toprule
\rowcolor{headergray}
\multicolumn{2}{c}{\textbf{Model Configuration (Lossless Path)}} \\ \midrule
Feature Channels C            & 64 \\
Convolution Kernel Size      & 5 \\
Core Blocks                  & \texttt{Target-Net/Prior-Net} blocks for prior and target feature processing \\
Prediction Head MLP Arch.    & C $\rightarrow$ 32 $\rightarrow$ 32 $\rightarrow$ 16 \\
Grouping Strategy            & 2-Stage Spatial Grouping \& 2-Stage Channel Grouping \\ \midrule
\rowcolor{headergray}
\multicolumn{2}{c}{\textbf{Model Configuration (Lossy Path)}} \\ \midrule
Feature Channels C            & 64 \\
Convolution Kernel Size      & 5 \\
Core Blocks                  & \texttt{Target-Net/Prior-Net} blocks for prior and target feature processing \\
Prediction Head MLP Arch. & C $\rightarrow$ 32 $\rightarrow$ 32 $\rightarrow$ 256 \\
Grouping Strategy            & Single-Grouping \\
Strategy                  & Top-k Occupancy Prediction \\
Configuration Parameter   & \texttt{--lossy\_top\_layers} $N = (1,2,3)$ \\
Mechanism                 & Switches to the lossy model for the $N$ finest-resolution layers. \\ \midrule
\rowcolor{headergray}
\multicolumn{2}{c}{\textbf{Ours Model Pre-training Hyperparameters}} \\ \midrule
Optimizer                    & Adam \\
Initial Learning Rate        & 5e-4 \\
Learning Rate Schedule       & Decays by a factor of 0.1 at steps 100k and 150k \\
Total Training Steps         & 170,000 \\
Batch Size                   & 1 \\
Dataset Sampling             & Weighted Random Sampling \\
Sampling Weights             & \begin{tabular}[t]{@{}l@{}}
                                 \texttt{KITTI}: 2.5, \texttt{8iVFB}: 2.5, \texttt{GS}: 2.0,
                                 \texttt{ScanNet}: 1.0, \texttt{Ford}: 1.0, \texttt{Thuman}: 1.0
                               \end{tabular} \\ \midrule
                               \rowcolor{headergray}
\multicolumn{2}{c}{\textbf{Instance-Adaptive Fine-Tuning (IAFT) Hyperparameters}} \\ \midrule
Tunable Modules              & All \texttt{nn.Linear} layers within the four prediction heads \\
Optimizer                    & Adam \\
Learning Rate                & 4e-3 \\
LR Scheduler                 & None \\
Fine-tuning Iterations       & 200 \\
Loss Function                & BPP (Cross-Entropy) + L1 Regularization \\
L1 Regularization $\lambda$  & 4e-4\\ \bottomrule
\end{tabular}
\end{table*}

\subsection{Lossless Compression}
\label{sec:lossless_compression}
Building upon the introduction of our Universal Context Model (UCM) and Instance-Adaptive Fine-Tuning (IAFT) strategy in the main paper, this section expands on our method. We provide the example of context model in Figure \ref{fig:context_example} and the detailed component architectures of our UCM in Figure \ref{fig:detailed} for reference. The main purpose of this section is to offer a detailed comparison of the UCM's context model against prior works. To facilitate this analysis, we present all comparisons in a consistent 2D setting at the same spatial scale, where a $2 \times 2$ block represents the sub-voxels of an occupancy code. The fundamental advantage of our approach is its operation at occupancy code scale, which grants both greater efficiency and a larger receptive field, as illustrated in Figure 2 of the main text. The following analysis will demonstrate this key distinction.
\begin{itemize}
    \item \textbf{RENO} \cite{you2025reno} partitions the channels of each occupancy code into two groups for autoregressive processing. Consequently, it overlooks the coarse-grained spatial priors at the occupancy code scale while also lacking any interaction of neighboring information between the two groups. These limitations severely restrict its compression efficiency.

    \item \textbf{SparsePCGC} \cite{sparsepcgc} implements spatial grouping at the voxel level. It shares a key limitation with RENO's channel grouping: both methods exclusively capture fine-grained priors and fail to incorporate coarse-grained spatial information. A further drawback is that its operation at the voxel scale results in a smaller receptive field compared to RENO.

    \item \textbf{UniPCGC} \cite{unipcgc} extends SparsePCGC by introducing a non-uniform grouping strategy. It reorganizes and further subdivides SparsePCGC's first group (in the figure, splitting the first group into two new, smaller groups). This design introduces a strong inductive bias: it assumes that the first group provides sufficient contextual priors, and that a more fine-grained modeling of it will yield performance gains. This strategy is effective for uniformly distributed point clouds but fails to generalize to the majority of real-world point clouds, which have highly irregular distributions.

    \item \textbf{AnyPcc} operates at the occupancy code scale to leverage a more comprehensive context. It is uniquely designed to integrate both fine-grained channel priors (\textit{intra-occupancy-code}) and coarse-grained spatial priors (\textit{inter-occupancy-code}), as illustrated in the Figure \ref{fig:context_example}. The model first encodes initial groups \{1, 2\} which build a coarse-grained representation of the local geometry. This representation then serves as a rich spatial prior for encoding the subsequent groups \{3, 4\}. Concurrently, dependencies within the groups (e.g., group 2 referencing 1, and 4 referencing 3) capture the fine-grained channel priors. By holistically integrating both types of dependencies, AnyPcc achieves consistently robust performance across point clouds of all types and densities.

\end{itemize}

\subsection{Lossy Compression}
\label{sec:implementation_lossy}

\begin{table*}[]
    \centering
\begin{threeparttable}
\caption{Detailed description of the train dataset.}
\label{table:trainset_des}
\begin{tabular}{cccccccc}
\toprule
\rowcolor{headergray}
\textbf{Dataset}       & \textbf{Bit-Depth} & \textbf{Frames / Seq} & \textbf{Total Frames}  & \multicolumn{1}{c}{\textbf{Cond.}}  &  \textbf{OOD} & \textbf{Avg. Points} & \textbf{Seq\tnote{*}} \\ \hline
\textbf{8iVFB}            & 10-bit             & 300                     & 600             &            Easy     & \xmark  & 954807 & Longdress Soldier                 \\
\textbf{MVUB}          & 10-bit             & /                       & 941             &          Easy   & \xmark   & 1339529 & Andrew David Sarah                     \\
\textbf{Thuman2.0}        & 10-bit             & /                       & 500              &       Easy       & \xmark    & 497907 & 0000-0499                    \\
\textbf{ScanNet}       & 2cm                & /                     & 1513             &       Easy      & \xmark     & 147977 & Scan                    \\
\textbf{KITTI}         & 1mm                & /                     & 23201            &         Easy     & \xmark         & 122780 & 00-10               \\
\textbf{Ford}          & 1mm                & 1500                     & 1500             &            Easy    & \xmark    & 82627 & 01                  \\
\textbf{GausPcc-1K}            & 1mm                & /                       & 1000              &               Medium     & \xmark     & 431165 & -             \\
\bottomrule
\end{tabular}
\begin{tablenotes}
    \item[*] Seq refers to the specific sequence used in the train, and - means that all sequences of the data set are used.
    \end{tablenotes}
    \end{threeparttable}
\end{table*}

In this section, we provide further details on our lossy compression strategy. Our multi-scale representation facilitates a straightforward approach to lossy compression. In a lossless layer, the bitrate $\mathcal{R}_{\text{lossless}}$ is determined by the entropy of the ground-truth occupancy codes $\mathbf{o}_{\text{gt}}$ given the predicted probabilities:
\begin{equation}
    \mathcal{R}_{\text{lossless}} = \sum_{i} -\log_2 P(o_i = o_{i, \text{gt}} \mid \text{Context}_i).
\end{equation}
A naive lossy approach could simply omit this entropy coding for higher-resolution layers. This approach is well-suited for sparse point clouds.
However, this approach leads to a significant loss of points from the point cloud, resulting in a substantial drop in PSNR.

To address this, we adopt a more sophisticated strategy. Instead of transmitting entropy-coded symbols for the designated lossy layers, our encoder transmits only the ground-truth point count \cite{sparsepcgc}, $k = |\mathcal{P}_{\text{gt}}|$. The decoder's task is then to predict the $k$ most probable occupied locations from the set of all candidate locations $\mathcal{C}_{\text{cand}}$. This process achieves effective lossy compression while preserving the correct point count.
A critical design choice for this is the context model. Our full, lossless models employs a multi-stage prediction scheme, which can lead to error accumulation in a lossy decoding scenario. Therefore, we deliberately switch to the more robust, single-stage model for the lossy layers.
The lossy decoding process proceeds as follows. For each candidate location $c_i \in \mathcal{C}_{\text{cand}}$, the baseline model predicts a full probability distribution $P_i$ over the 256 possible occupancy codes.

\begin{enumerate}
    \item \textbf{Occupancy Probability Estimation:} First, we compute the probability that each location is occupied by summing the probabilities of all non-zero outcomes:
    \begin{equation}
        p^{\text{occ}}_i = \sum_{v=1}^{255} P_i(o=v) = 1 - P_i(o=0).
    \end{equation}
    \item \textbf{Location Selection:} The decoder identifies the set of indices $\mathcal{I}_{\text{dec}}$ corresponding to the $k$ locations with the highest occupancy probabilities:
    \begin{equation}
        \mathcal{I}_{\text{dec}} = \text{Top-K-Indices}(\{p^{\text{occ}}_i\}_{i \in \mathcal{C}_{\text{cand}}}, k).
    \end{equation}
    \item \textbf{Value Prediction:} For each selected location $c_j$ where $j \in \mathcal{I}_{\text{dec}}$, the final occupancy code $\hat{o}_j$ is predicted by selecting the most likely non-zero value:
    \begin{equation}
        \hat{o}_j = \underset{v \in \{1, \dots, 255\}}{\arg\max} \, P_j(o=v).
    \end{equation}
\end{enumerate}
The final decoded point cloud for the layer is the set of predicted pairs $\{ (c_j, \hat{o}_j) \}_{j \in \mathcal{I}_{\text{dec}}}$. Since this prediction is based on model probabilities rather than ground-truth data, it naturally introduces loss while maintaining geometric structure. This hybrid approach effectively mitigates catastrophic point loss in dense regions and enables a controllable, high-performance lossy compression mode.

\begin{table*}[]
    \centering
\begin{threeparttable}
\caption{Detailed description of the test dataset.}
\label{table:testset_des}
\begin{tabular}{cccccccc}
\toprule
\rowcolor{headergray}
\textbf{Dataset}       & \textbf{Bit-Depth} & \textbf{Frames / Seq} & \textbf{Total Frames}  & \multicolumn{1}{c}{\textbf{Cond.}}  &  \textbf{OOD} & \textbf{Avg. Points} & \textbf{Seq\tnote{*}} \\ \hline
\textbf{8iVFB}            & 10-bit             & 300                     & 600             &            Easy     & \xmark  & 760375 & Loot Red                 \\
\textbf{MVUB}          & 10-bit             & /                       & 461             &          Easy   & \xmark   & 1259792 & Phil Ricardo                     \\
\textbf{Owlii}         & 11-bit             & 600                     & 2400            &          Easy      & \xmark     & 2635971 & -                 \\
\textbf{Thuman}        & 10-bit             & /                       & 20              &       Easy       & \xmark    & 441319 & 0500-0519                    \\
\textbf{ScanNet}       & 2cm                & 100                     & 100             &       Easy      & \xmark     & 126760 & scan\_test                    \\
\textbf{KITTI}         & 1mm                & 100                     & 1100            &         Easy     & \xmark         & 119826 & 11-21               \\
\textbf{Ford}          & 1mm                & 100                     & 200             &            Easy    & \xmark    & 83061 & 02-03                  \\
\textbf{Dense\_vox12}  & 12-bit             & /                       & 2               &                  Medium    & \xmark   & 3222415 & -              \\
\textbf{Sparse\_vox12} & 12-bit             & /                       & 3               &               Medium      & \xmark    & 996846 & -             \\
\textbf{GS}            & 1mm                & /                       & 13              &               Medium     & \xmark     & 288413 & -             \\
\textbf{VGGT}          & 1mm                & /                       & 50              &            Medium       & \cmark    & 257467 & -               \\
\textbf{S3DIS}         & 16-bit             & /                       & 1              &           Medium        & \cmark   & 47062002 &  Stanford\_Area\_2       \\        
\textbf{RS}            & 10-12bit           & /                       & 16              &          Hard         & \cmark   & 1058013 & -                   \\
\textbf{NS}        & 10-12bit           & /                       & 15              &         Hard         & \cmark   & 1670503 & -                    \\ 
\textbf{CS}        & 10-12bit           & /                       & 14              &         Hard         & \cmark   & 1531537 & -              \\ 
\bottomrule
\end{tabular}
\begin{tablenotes}
    \item[*] Seq refers to the specific sequence used in the test, and - means that all sequences of the data set are used.
    \end{tablenotes}
    \end{threeparttable}
\end{table*}

\section{AnyPcc Benchmark Dataset}
\label{sec:suppl_dataset}
\subsection{Trainset}
\paragraph{Training Datasets.}
Our training data is comprehensive, incorporating a variety of sources to ensure broad coverage. Specifically, we include standard benchmarks widely used in the Point Cloud Compression (PCC) community, such as \texttt{8iVFB}, \texttt{MVUB}, \texttt{ScanNet}, \texttt{KITTI}, and \texttt{Ford}. To further enrich our training diversity, we supplement these with the \texttt{Thuman} and \texttt{GausPcc-1K} \cite{wang2025novel} datasets. Additionally, we incorporate the \texttt{Frog}, \texttt{Head}, \texttt{Egyptian}, and \texttt{ULB} datasets, which are recommended by the MPEG AI-PCC group for training purposes. All training sets are first quantized according to the precisions specified in Table \ref{table:testset_des}. Additionally, the \texttt{KITTI} and \texttt{Ford} datasets undergo a second quantization step where the resulting integer coordinates are further scaled by a factor of Q=4.

\paragraph{Training Strategy.}
The prevailing paradigm in Point Cloud Compression (PCC) involves training models that are specialized for distinct data categories. For a fair comparison with existing methods, we first follow this standard practice by training a series of \textbf{specialized models}, one for each data category. However, a primary objective of AnyPcc is to develop a single, universal compressor that performs robustly across arbitrary point clouds. To this end, we also pursue a second strategy: training a single \textbf{unified model} on a large-scale, composite dataset. This dataset is constructed by sampling from multiple sources with the following rates: \texttt{8iVFB} and \texttt{MVUB} (2.5, collectively referred to as \texttt{8iVFB} for brevity), \texttt{Thuman} (1.0), \texttt{ScanNet} (1.0), \texttt{Ford} (1.0), \texttt{KITTI} (2.5), and \texttt{GS} (2.0). This unified approach facilitates a rigorous evaluation of our model's generalization capabilities. To ensure a fair assessment, all training sequences are strictly disjoint from our test set, as detailed in Table~\ref{table:testset_des}.

\subsection{Testset}

As detailed in Table~\ref{table:testset_des}, we provide a comprehensive analysis of the datasets used in our evaluation. We distinguish these datasets as either Out-of-Distribution (OOD) or In-Distribution (ID). A dataset is considered \textbf{OOD} if our model was not exposed to semantically similar data during training. In our benchmark, this includes \texttt{GS}, \texttt{VGGT}, \texttt{S3DIS}, \texttt{RS}, and \texttt{Random}. Conversely, a dataset is considered \textbf{ID} when the test data is of the same type as the training data. For instance, for \texttt{KITTI}, we train on sequences 00-10 and test on the first 100 frames of sequences 11-21. Based on characteristics such as point distribution, bit-depth or quantization step, and scale, we categorize all datasets into three levels of compression difficulty. For the \texttt{KITTI} and \texttt{Ford} sequences, we employ a two-step quantization process. First, the original coordinates are quantized to a 1mm precision by dividing them by 0.001. Second, these integer coordinates are further scaled by a quantization parameter Q=16, a setting aligned with previous work \cite{you2025reno}, before lossless compression. For all other datasets, this second scaling step is omitted (i.e., Q=1).

\paragraph{Easy Difficulty Datasets.} This category includes several benchmarks widely adopted in the PCC community: \texttt{8iVFB}, \texttt{MVUB}, \texttt{Owlii}, \texttt{ScanNet}, \texttt{KITTI}, and \texttt{Ford}. These datasets predominantly feature either dense or sparse distributions. This allows learning-based methods to achieve strong performance by specializing in a specific data distribution. We also include \texttt{Thuman}, a mesh dataset that we converted to 10-bit point clouds through sampling and quantization. Given its distribution similarity to other human-centric datasets, we classify its compression difficulty as easy.

\paragraph{Medium Difficulty Datasets.} This group comprises datasets with greater scale and complexity. \texttt{Dense-vox12} and \texttt{Sparse-vox12} are standard test sets defined by MPEG AI-PCC group. Additionally, \texttt{S3DIS} is a large-scale indoor scene dataset with point counts reaching tens of millions. The higher point counts and bit-depth of these datasets justify their classification as medium difficulty. Furthermore, we introduce two specialized datasets into this category. The first is our custom \texttt{VGGT} dataset, generated by reconstructing point clouds from 50 multi-view images (resized to 518x518) using the VGGT \cite{wang2025vggt} algorithm. The second is the Gaussian Splatting (\texttt{GS}) dataset, which is created by extracting geometric coordinates from a pre-trained 3DGS model, following the methodology in~\cite{kerbl20233d} and \cite{wang2025novel}. Both \texttt{VGGT} and \texttt{GS} point clouds exhibit a unique characteristic of being locally dense yet globally sparse, distinguishing them from conventional datasets and thus warranting a medium difficulty rating.

\paragraph{Hard Difficulty Datasets.}
To evaluate the robustness of Point Cloud Compression (PCC) algorithms under challenging conditions, we systematically construct three distinct datasets by applying controlled perturbations to source point clouds from the \texttt{8iVFB}, \texttt{Owlii}, and \texttt{Dense-vox12} collections. These datasets, designated as Random Sampling (\texttt{RS}), Noise-added (\texttt{NS}), and Complex Sinusoidal (\texttt{CS}), are designed to disrupt the intrinsic structural regularities of point clouds. They serve as out-of-distribution (OOD) benchmarks to probe the generalization capabilities of PCC models and simulate real-world corruptions like sensor noise, packet loss, or non-rigid deformations.
\begin{table*}[]
    \centering
    \caption{Performance comparison with SparsePCGC-U and RENO-U. The table presents the compression performance of AnyPcc against 2 methods across 7 datasets, with the best and second-best results highlighted in \colorbox[HTML]{FFC7CE}{\textcolor{black}{red}} and \colorbox[HTML]{FFEB9C}{\textcolor{black}{yellow}} cells.}
\label{table:abla_unidata}
    \begin{tabular}{c|ccccccc}
        \toprule
        \rowcolor{headergray}
    \textbf{Dataset}               & \textbf{RENO}               & \textbf{RENO-U} & \textbf{SparsePCGC} & \textbf{SparsePCGC-U}        & \textbf{Ours}               & \textbf{Ours-U} & \textbf{GPCC}        \\ \toprule
    \textbf{8iVFB}          & 0.70                        & 0.76            & \colorbox[HTML]{FFEB9C}{0.57}                & 0.58                         & \colorbox[HTML]{FFC7CE}{0.54}                        & \colorbox[HTML]{FFEB9C}{0.57}            & 0.76                 \\
    \textbf{MVUB}           & 1.00                        & 1.07            & 0.69                & \colorbox[HTML]{FFEB9C}{0.73   }                      & \colorbox[HTML]{FFC7CE}{0.67}                        & 0.75            & 0.94                 \\
    \textbf{Owlii}          & 0.59                        & 0.61            & \colorbox[HTML]{FFEB9C}{0.48 }               & 0.49                         & \colorbox[HTML]{FFC7CE}{0.47}                        & \colorbox[HTML]{FFC7CE}{0.47}            & 0.59                 \\
    \textbf{Thuman}         & 1.64                        & 1.72            & 1.70                & \colorbox[HTML]{FFEB9C}{1.63   }                      & \colorbox[HTML]{FFC7CE}{1.58}                        & 1.64            & 2.00                 \\
    \textbf{ScanNet}        & {2.15} & 2.26            &\colorbox[HTML]{FFEB9C}{ 1.86 }               & 1.90                         & {\colorbox[HTML]{FFC7CE}{1.83}} & 1.88            & 2.03                 \\
    \textbf{KITTI}          & 7.06                        & 7.36            & 6.80                & {7.43}  & \colorbox[HTML]{FFC7CE}{6.18}                        & \colorbox[HTML]{FFEB9C}{6.45  }          & 8.17                 \\
    \textbf{Ford}           & 9.38                        & 9.60            & 9.77                & { 10.20} & \colorbox[HTML]{FFC7CE}{8.40}                        &\colorbox[HTML]{FFEB9C}{ 8.57   }         & 10.32                \\ \toprule
    \textbf{Gain over GPCC} & -5.17\%                      & -0.10\%          & -16.48\%             & -13.96\%                      & \colorbox[HTML]{FFC7CE}{-21.76\%}                     & \colorbox[HTML]{FFEB9C}{-18.45\% }        & 0.00\%                \\
    \textbf{Enc/Dec Times}        & \multicolumn{2}{c}{\colorbox[HTML]{FFC7CE}{0.19/0.20}}                          & \multicolumn{2}{c}{2.2/2.1}                              & \multicolumn{2}{c}{\colorbox[HTML]{FFEB9C}{0.39/0.40}}                          & \multicolumn{1}{c}{3.3/2.4} \\ \toprule
    \end{tabular}
    \end{table*}

\begin{itemize}
\item \textbf{Random Sampling (RS).} The \texttt{RS} dataset is generated via a spatially-dependent probabilistic downsampling. For each point $\mathbf{p}_{i}=(x_i, y_i, z_i)$ in the original cloud $\mathcal{P}_{\text{ori}}$, we first normalize its coordinates to a unit cube $[0, 1]^3$ based on the cloud's bounding box, yielding $\hat{\mathbf{p}}_{i}$. A retention probability $P(\hat{\mathbf{p}}_{i})$ is then computed using a composite trigonometric function:
\begin{equation}
\begin{split}
    P(\hat{\mathbf{p}}_{i}) = \frac{1}{6} \Biggl( & \sin(2\pi f_x \hat{x}_i + \phi_x) \\
    & + \cos(2\pi f_y \hat{y}_i + \phi_y) \\
    & + \sin(2\pi f_z \hat{z}_i + \phi_z) + 3 \Biggr),
\end{split}
\end{equation}
where $\{f_x, f_y, f_z\}$ are frequency parameters and $\{\phi_x, \phi_y, \phi_z\}$ are phase shifts. A point is retained if a random variable $r_i \sim \mathcal{U}(0, 1)$ satisfies $r_i \leq P(\hat{\mathbf{p}}_{i})$. This process introduces non-uniform density variations, creating a sparsely sampled point cloud $\mathcal{P}_{\text{RS}}$.

\item \textbf{Noise-added (NS).} The \texttt{NS} dataset is formed by corrupting point coordinates with Additive White Gaussian Noise (AWGN). The noise variance is adaptively scaled with the point cloud's spatial extent. The transformation for each point $\mathbf{p}_i \in \mathcal{P}_{\text{ori}}$ is:
\begin{equation}
    \mathbf{p}'_i = \mathbf{p}_i + \boldsymbol{\epsilon}_i, \quad \text{where} \quad \boldsymbol{\epsilon}_i \sim \mathcal{N}\left(\mathbf{0}, \sigma^2 \mathbf{I}_3\right).
\end{equation}
The standard deviation $\sigma$ is proportional to the diagonal length of the original cloud's axis-aligned bounding box (AABB), controlled by a ratio parameter $\gamma$:
\begin{equation}
    \sigma = \gamma \cdot \|\mathbf{p}_{\max} - \mathbf{p}_{\min}\|_2.
\end{equation}
Crucially, following the noise injection, the coordinates of the resulting points are rounded to the nearest integers, and any duplicate points are removed. This simulates sensor quantization effects. The final point cloud is thus $\mathcal{P}_{\text{NS}} = \text{Unique}(\text{Round}(\{\mathbf{p}'_i\}))$.

\item \textbf{Complex Sinusoidal (CS).} The \texttt{CS} dataset is created by applying a non-rigid deformation. This transformation, $T_{\text{deform}}$, first normalizes the input cloud $\mathcal{P}_{\text{ori}}$ to a unit cube, applies a sinusoidal deformation, and then rescales the result to the original bounding box dimensions. The core deformation is a composition of radial and per-axis sinusoidal scaling functions:
\begin{align}
    \mathcal{P}_{\text{CS}} &= T_{\text{deform}}(\mathcal{P}_{\text{ori}}, \alpha), \\
    \text{where } \mathbf{p}' &= \mathbf{S}_{\text{coord}} \circ \mathbf{S}_{\text{radial}}(\hat{\mathbf{p}}), \\
    \text{and } \mathbf{S}_{\text{radial}}(\hat{\mathbf{p}}) &= \hat{\mathbf{p}} \cdot (1 + \alpha \sin(5\pi \|\hat{\mathbf{p}}\|_2)).
\end{align}
Here, $\hat{\mathbf{p}}$ is a normalized point, $\alpha$ controls the deformation degree, and $\mathbf{S}_{\text{coord}}$ represents a per-axis sinusoidal scaling.
The resulting datasets, with their deliberately degraded and less predictable structures, present a significant challenge to existing compression algorithms, thus serving as robust benchmarks for evaluating model generalization under adverse conditions.
\end{itemize}

\paragraph{Summary.}
To our knowledge, we establish the most comprehensive benchmark for point cloud compression to date, designed to rigorously evaluate algorithms on two critical axes: generalization and stability. 
To assess generalization, our benchmark incorporates a wide data spectrum, ranging from standard objects and scenes to LiDAR scans and challenging point clouds from modern pipelines such as feed-forward reconstruction (\texttt{VGGT}) and neural representations (\texttt{GS}). For stability, we introduce demanding datasets with procedurally generated corruptions that stress-test algorithmic robustness against real-world imperfections. By systematically evaluating performance in these extensive scenarios, our benchmark aims to guide future research toward developing more versatile and resilient compression solutions.

\section{More Experimental Results}

\subsection{Lossless Compression}
\begin{table*}[]
    \centering
    \caption{Performance comparison on the dense point cloud Benchmark. The table presents the compression performance of AnyPcc against eleven methods across 2 datasets, with the best and second-best results highlighted in \colorbox[HTML]{FFC7CE}{\textcolor{black}{red}} and \colorbox[HTML]{FFEB9C}{\textcolor{black}{yellow}} cells.}
    \label{table:dense_bench}
    \setlength{\tabcolsep}{4pt} 
    \begin{tabular}{cccccccc}
        \toprule
    \rowcolor{headergray}
    \multicolumn{1}{c}{} & \multicolumn{4}{c}{\textbf{MPEG 8i}} & \multicolumn{2}{c}{\textbf{MVUB}} & \textbf{Metrics} \\  
    \rowcolor{headergray}
    \multicolumn{1}{c}{\multirow{-2}{*}{\textbf{Method}}} & {Loot ×300} & {Red\&black ×300} & {Boxer ×1} & \multicolumn{1}{c}{{Thaidancer ×1}} & {Phil ×245} & \multicolumn{1}{c}{{Ricardo ×216}} & {Ours-CR Gain} \\ 
    \midrule 
    \rowcolor{headergray}
    \multicolumn{8}{c}{\textbf{Non-Learning}} \\ 
    \midrule 
    \multicolumn{1}{c}{\textbf{GPCC v23} \cite{gpcc}} & 0.82 & 0.69 & 0.65 & \multicolumn{1}{c}{0.70} & 0.96 & \multicolumn{1}{c}{0.89} & -28.42\% \\ 
    \midrule 
    \rowcolor{headergray}
    \multicolumn{8}{c}{\textbf{Octree-Based}} \\ 
    \midrule 
    \multicolumn{1}{c}{\textbf{OctAttention} \cite{fu2022octattention}} & 0.62 & 0.73 & 0.59 & \multicolumn{1}{c}{0.65} & 0.79 & \multicolumn{1}{c}{0.72} & -18.54\% \\
    \multicolumn{1}{c}{\textbf{ECM-OPCC} \cite{jin2024ecm}} & 0.55 & 0.66 & 0.51 & \multicolumn{1}{c}{0.58} & 0.76 & \multicolumn{1}{c}{0.69} & -10.71\% \\
    \multicolumn{1}{c}{\textbf{EHEM} \cite{song2023efficient}} & 0.58 & 0.69 & 0.55 & \multicolumn{1}{c}{0.62} & / & \multicolumn{1}{c}{/} & -17.37\% \\
    \multicolumn{1}{c}{\textbf{TopNet} \cite{topnet}} & 0.53 & 0.65 & 0.51 & \multicolumn{1}{c}{0.65} & 0.73 & \multicolumn{1}{c}{\colorbox[HTML]{FFC7CE}{0.64}} & -9.67\% \\ 
    \midrule 
    \rowcolor{headergray}
    \multicolumn{8}{c}{\textbf{INR-Based}} \\ 
    \midrule 
    \multicolumn{1}{c}{\textbf{LINR-PCGC} \cite{huang2025linr}} & 0.57 & 0.69 & / & \multicolumn{1}{c}{/} & 0.84 & \multicolumn{1}{c}{0.80} & -16.74\% \\
    \multicolumn{1}{c}{\textbf{LINR-PCGC-2} \cite{huang2025linr}} & 0.52 & 0.63 & / & \multicolumn{1}{c}{/} & 0.77 & \multicolumn{1}{c}{0.71} & -8.25\% \\ 
    \midrule 
    \rowcolor{headergray}
    \multicolumn{8}{c}{\textbf{Voxel-Based}} \\ 
    \midrule 
    \multicolumn{1}{c}{\textbf{VoxelDNN} \cite{voxeldnn}} & 0.58 & 0.67 & 0.55 & \multicolumn{1}{c}{0.58} & 0.76 & \multicolumn{1}{c}{0.69} & -12.73\% \\
    \multicolumn{1}{c}{\textbf{MsVoxelDNN} \cite{msvoxel}} & 0.73 & 0.87 & 0.70 & \multicolumn{1}{c}{0.85} & 1.02 & \multicolumn{1}{c}{0.95} & -34.60\% \\ 
    \multicolumn{1}{c}{\textbf{UniPCGC} \cite{unipcgc}} & \colorbox[HTML]{FFEB9C}{0.49} & \colorbox[HTML]{FFC7CE}{0.59} & \colorbox[HTML]{FFEB9C}{0.45} & \multicolumn{1}{c}{\colorbox[HTML]{FFC7CE}{0.51}} & 0.80 & \multicolumn{1}{c}{0.73} & \colorbox[HTML]{FFEB9C}{-4.85\%} \\
    \multicolumn{1}{c}{\textbf{SparsePCGC} \cite{sparsepcgc}} & 0.53 & 0.64 & 0.49 & \multicolumn{1}{c}{0.56} & \colorbox[HTML]{FFEB9C}{0.71} & \multicolumn{1}{c}{\colorbox[HTML]{FFEB9C}{0.65}} & -6.55\% \\
    \multicolumn{1}{c}{\textbf{Unicorn} \cite{unicorn}} & 0.53 & 0.64 & 0.49 & \multicolumn{1}{c}{0.56} & \colorbox[HTML]{FFEB9C}{0.71} & \multicolumn{1}{c}{\colorbox[HTML]{FFEB9C}{0.65}} & -6.55\% \\
    \multicolumn{1}{c}{\textbf{RENO} \cite{you2025reno}} & 0.65 & 0.77 & 0.61 & \multicolumn{1}{c}{0.69} & 1.03 & \multicolumn{1}{c}{0.97} & -28.26\% \\ 
    \midrule 
    
    \multicolumn{1}{c}{\textbf{AnyPcc (Ours)}} & \cellcolor[HTML]{FFC7CE}{0.48} & \cellcolor[HTML]{FFC7CE}{0.59} & \cellcolor[HTML]{FFC7CE}{0.44} & \multicolumn{1}{c}{\cellcolor[HTML]{FFC7CE}{0.51}} & \cellcolor[HTML]{FFC7CE}{0.70} & \multicolumn{1}{c}{\cellcolor[HTML]{FFC7CE}{0.64}} & \cellcolor[HTML]{FFC7CE}{0.00\%} \\
    \multicolumn{1}{c}{\textbf{AnyPcc-U (Ours-U)}} & 0.52 & 0.62 & 0.48 & \multicolumn{1}{c}{0.54} & 0.78 & \multicolumn{1}{c}{0.72} & -7.96\% \\ 
    \bottomrule 
    \end{tabular}
\end{table*}
 
\paragraph{Comparison with SparsePCGC-U and RENO-U.}
\label{sec:unidata_train}
To facilitate a comparison with Unicorn-U \cite{unicorn}, we conduct experiments using our unified training dataset. Since Unicorn is not yet open-source, we use its predecessor SparsePCGC \cite{sparsepcgc} as our primary baseline. We retrain both RENO \cite{you2025reno} and SparsePCGC \cite{sparsepcgc} on our unified dataset, denoting these versions as RENO-U and SparsePCGC-U, respectively. The results are presented in Table \ref{table:abla_unidata}, under strictly aligned training and testing conditions for all methods. The comparison is conducted on seven datasets where AnyPcc is evaluated without its Instance-Adaptive Fine-Tuning (IAFT) module to ensure a fair and direct comparison of the base models.

Our results indicate that SparsePCGC-U exhibits more stable performance on object point clouds. We attribute this to SparsePCGC's use of a spatial auto-regressive eight-group context, which introduces a strong inductive bias well-suited for dense object data. In contrast, AnyPcc-U employs a four-group context that combines spatial and channel information, resulting in a weaker inductive bias and consequently superior performance in sparser scenarios. As discussed in Sec. \ref{sec:contex_model_abla}, AnyPcc can be configured with additional spatial and channel groups to introduce a stronger inductive bias if needed.
Furthermore, AnyPcc features a lightweight architecture, achieving over a 5x speedup in both encoding and decoding compared to SparsePCGC, while also delivering better performance.

In summary, training on a unified dataset leads to a slight degradation in performance on specific datasets, an observation consistent with the findings for Unicorn-U. However, this approach significantly enhances the model's generalization capabilities, which is a crucial step towards universal point cloud compression. Nevertheless, pre-training on large-scale data alone is insufficient for robust generalization to out-of-distribution (OOD) data, underscoring the necessity of the IAFT strategy proposed in this work.

\begin{table*}[]
\centering
\begin{threeparttable}
\caption{Lossy Compression performance comparison on the AnyPcc Benchmark. The table presents the compression performance of AnyPcc against six methods across 8 \tnote{†} diverse datasets using BD-Rate. The best and second-best results highlighted in \colorbox[HTML]{FFC7CE}{\textcolor{black}{red}} and \colorbox[HTML]{FFEB9C}{\textcolor{black}{yellow}} cells.}
\label{table:all_method_lossy}
\setlength{\tabcolsep}{5pt} 
\begin{tabular}{c|c|cccccc|cc} 
\toprule
\rowcolor{headergray}
\textbf{Dataset}  & \textbf{OOD} & \textbf{RENO} & \textbf{SparsePCGC} & \textbf{Unicorn\tnote{*}} & \textbf{OctAttention} & \textbf{TopNet} & \textbf{GPCC} & \textbf{Ours} & \textbf{Ours-U} \\ 
\midrule
\textbf{8iVFB}         & \xmark & /         & -20.29\%     & -20.29\% & / & / & -94.39\%             & \cellcolor[HTML]{FFC7CE}{0.00\%}  & \cellcolor[HTML]{FFEB9C}{-10.54\%}       \\
\textbf{MVUB}                          & \xmark & /         & -16.00\%  & -16.00\% & / &  /    & -93.61\%             & \cellcolor[HTML]{FFC7CE}{0.00\%}    & \cellcolor[HTML]{FFEB9C}{-12.13\%}     \\
\textbf{Owlii}                             & \xmark & /         & -23.70\%   & -23.70\%  & / & / & -93.76\% & \cellcolor[HTML]{FFC7CE}{0.00\%}  & \cellcolor[HTML]{FFEB9C}{-11.75\%}       \\
\textbf{Thuman}                           & \xmark & /         & -54.89\%  & -54.89\%  & / & /           & -91.41\%             & \cellcolor[HTML]{FFC7CE}{0.00\%}  & \cellcolor[HTML]{FFEB9C}{-16.54\%}       \\
\textbf{ScanNet}                          & \xmark & /         & -82.15\%  & -37.34\% & / & /              & -81.57\%             & \cellcolor[HTML]{FFC7CE}{0.00\%} & \cellcolor[HTML]{FFEB9C}{-2.96\%}        \\
\midrule
\multicolumn{2}{c|}{\textbf{Ours BD-Rate Gains}} & / & -39.41\% & -30.44\% & / & / & -90.95\% & \cellcolor[HTML]{FFC7CE}{0.00\%} & \cellcolor[HTML]{FFEB9C}{-10.78\%} \\
\midrule
\textbf{KITTI}                             & \xmark &  -12.87\%        & -4.49\% & \underline{-3.62\%} & -10.17\% & -4.48\%             & -22.45\%              & \cellcolor[HTML]{FFC7CE}{0.00\%}   & -3.85\%      \\
\textbf{Ford}                             & \xmark & -11.08\%        & -6.79\% & \cellcolor[HTML]{FFC7CE}{1.18\%} & -5.96\% & -1.18\%               & -18.44\%            & \cellcolor[HTML]{FFEB9C}{0.00\%}  & -0.99\%       \\ 
\textbf{VGGT}                            & \cmark & -8.13\%         & -2.33\%   & / & -8.33\% & -3.65\%            & \cellcolor[HTML]{FFEB9C}{0.94\%}             & 0.00\% & \cellcolor[HTML]{FFC7CE}{5.42\%}       \\
\midrule
\multicolumn{2}{c|}{\textbf{Ours BD-Rate Gains}} & -10.69\% & -4.54\% & / & -8.15\% & -3.10\% & -13.31\% & \cellcolor[HTML]{FFEB9C}{-0.00\%} & \cellcolor[HTML]{FFC7CE}{0.19\%} \\ \midrule
\multicolumn{2}{c|}{\textbf{Enc/Dec Time (s)}} & \cellcolor[HTML]{FFC7CE}{0.17/0.17} & 1.28/1.20 & / & 1.20/562 & 1.50/758 & 1.83/0.97 & \multicolumn{2}{c}{\cellcolor[HTML]{FFEB9C}{0.20/0.18}}  \\
\multicolumn{2}{c|}{\textbf{Total Params (M)}} & 9.03 & 27.03 & / & 29.61 & 23.59 & / & 104.44 & 14.92 \\
\bottomrule
\end{tabular}
\begin{tablenotes}
    \item[†] For lossy compression, our evaluation is focused on 8 datasets. The other 7 were excluded because competing methods produced unstable results, preventing a fair alignment of rate points. Since the lossy strategy for these datasets mirrors that of KITTI, their lossless performance trends offer a reliable indication of their likely lossy performance.
      \item[*] The results for Unicorn are cited directly from the original publication as its implementation is not open-source.
    \end{tablenotes}
\end{threeparttable}
\end{table*}

\paragraph{Comparison in Dense Point Cloud Benchmark.}
\label{sec:dense point cloud benchmark}
To ensure a fair comparison with leading methods such as OctAttention \cite{fu2022octattention}, VoxelDNN \cite{voxeldnn}, UniPCGC \cite{unipcgc}, and SparsePCGC \cite{sparsepcgc} on human point clouds, we evaluate our method under their specific training and testing protocols, with results presented in Table \ref{table:dense_bench}. This protocol involves training on 10-bit versions of \texttt{Soldier} and \texttt{Longdress} from the \texttt{8iVFB} \cite{dataset8i} dataset, and \texttt{Andrew}, \texttt{David}, and \texttt{Sarah} from the \texttt{MVUB} \cite{mvub} dataset. The evaluation is performed on the \texttt{Loot}, \texttt{Redandblack}, \texttt{Phil}, \texttt{Ricardo}, \texttt{Boxer}, and \texttt{Thaidancer} sequences. Our comprehensive benchmark results are detailed in Table \ref{table:dense_bench}. We compare against a wide range of baselines: the traditional method GPCC v23 \cite{gpcc}; voxel-based deep learning methods including UniPCGC \cite{unipcgc}, SparsePCGC \cite{sparsepcgc}, RENO \cite{you2025reno}, VoxelDNN \cite{voxeldnn}, and MsVoxelDNN \cite{msvoxel}; octree-based methods like OctAttention \cite{fu2022octattention}, ECM-OPCC \cite{jin2024ecm}, EHEM \cite{song2023efficient}, and TopNet \cite{topnet}; and implicit compression methods such as LINR-PCGC \cite{huang2025linr} and LINR-PCGC-2 \cite{huang2025linr}. For human point cloud compression, Unicorn \cite{unicorn} employs the same methodology as SparsePCGC. Since Unicorn is not open-source, we report the performance of this method using the official SparsePCGC implementation. 
Note that LINR-PCGC-2 is a variant of LINR-PCGC with a longer fine-tuning duration, resulting in a higher encoding time.
\begin{itemize}
    \item \textbf{Comparison to SparsePCGC.} Compared to SparsePCGC / Unicorn, AnyPcc is more lightweight as it introduces only four predictive groups, yet it delivers superior performance. Furthermore, AnyPcc can be extended with additional channel and spatial groups to potentially achieve even higher compression efficiency. The impact of our context modeling choices is further analyzed in the Sec. \ref{sec:contex_model_abla}.
    
    \item \textbf{Comparison to UniPCGC.} It is noteworthy that UniPCGC builds upon SparsePCGC by introducing a non-uniform eight-group strategy. This approach is particularly effective for point clouds with a uniform spatial distribution. In the \texttt{8iVFB} dataset, UniPCGC achieves performance second only to AnyPcc. However, this strategy introduces a strong inductive bias by assuming that the initial groups can always provide sufficient prior information. This assumption does not hold for many real-world point clouds with non-uniform distributions, which explains its suboptimal performance on the \texttt{MVUB} dataset.
     \item \textbf{Comparison to LINR-PCGC.} Furthermore, the practical application of implicit methods like LINR-PCGC and LINR-PCGC-2 is limited by their substantial encoding times (ranging from 2 to 18 seconds) and their restriction to multi-frame point cloud sequences. Moreover, their compression efficiency is not state-of-the-art.
      \item \textbf{Summary.} We have aligned our evaluation with the established community standards for human point cloud compression and conducted a comprehensive benchmark. The results confirm that AnyPcc consistently achieves state-of-the-art performance across all tested conditions.
\end{itemize}

\subsection{Lossy Compression}
\label{sec:lossy compression res}
\begin{figure*}[t]
    \centering
    \includegraphics[width=1\linewidth]{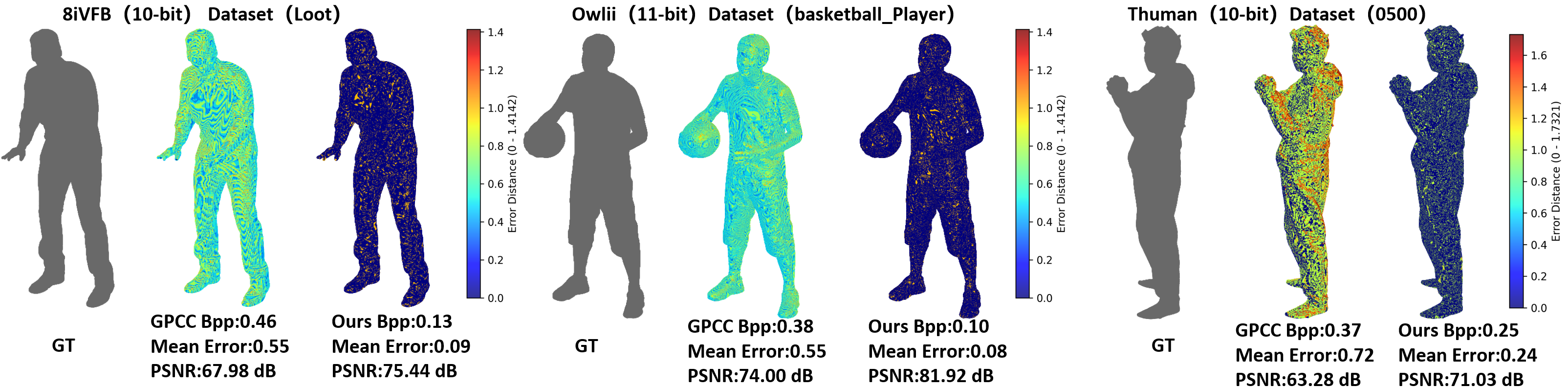}
    \caption{Point-to-point error heatmaps for the dense 8iVFB, Owlii and Thuman datasets.}
    \label{fig:vis_dense}
\end{figure*}

We present further experimental results for lossy compression in Table~\ref{table:all_method_lossy}. Our method demonstrates superior performance, outperforming various learning-based approaches and the conventional G-PCC v23 codec. The table quantifies this advantage by reporting the Bjontegaard Delta Rate (BD-Rate) gains of our model (``Ours'') over competitors. A lower BD-Rate value signifies a greater improvement. For instance, -90\% indicates that our method achieves a 90\% bitrate saving for the same PSNR quality on average. Furthermore, our encoding and decoding times are comparable to those of RENO, one of the fastest methods, underscoring AnyPcc's role as a lightweight yet unified solution for point cloud compression. Notably, our ``Ours-U'' variant utilizes a single model to perform both lossy and lossless compression across diverse point cloud types, which substantially enhances the practicality of deep learning-based PCC. Moreover, We provide point-to-point error heatmaps for the dense 8iVFB, Owlii and Thuman dataset in Figure~\ref{fig:vis_dense}. The error distribution visually confirms that AnyPcc preserves fine geometric details effectively.

\subsection{Ablation Studies}

\paragraph{Context Model.}
\begin{table*}[]
    \centering
    \caption{Ablation study of the context model. The table is split into two main experiments: varying the number of groups for Spatial2 / Channel2 and Spatial1 / Channel1 models respectively.}
    \label{table:abla_contex}
    \setlength{\tabcolsep}{9pt} 
    \begin{tabular}{c|c|cccc|ccc}
        \toprule
    \multicolumn{1}{c|}{\cellcolor{headergray}} & \multicolumn{1}{c|}{\cellcolor{headergray}} & \multicolumn{4}{c|}{\cellcolor{headergray}\textbf{Spatial2 (Channel: 1 to 4)}} & \multicolumn{3}{c}{\cellcolor{headergray}\textbf{Channel2 (Spatial: 1 to 4)}} \\
    \multicolumn{1}{c|}{\cellcolor{headergray}\multirow{-2}{*}{\textbf{Dataset}}} & \multicolumn{1}{c|}{\cellcolor{headergray}\multirow{-2}{*}{\textbf{OOD}}} & \cellcolor{headergray}\textbf{1-group} & \cellcolor{headergray}\textbf{2-group} & \cellcolor{headergray}\textbf{4-group} & \cellcolor{headergray}\textbf{ue-4-group} & \cellcolor{headergray}\textbf{1-group} & \cellcolor{headergray}\textbf{2-group} & \cellcolor{headergray}\textbf{4-group} \\
        \midrule
    \textbf{8iVFB}       & \xmark       & 0.604            & 0.570            & \textbf{0.563}            & \textbf{0.563}               & 0.635            & \textbf{0.570}            & 0.579                   \\
    \textbf{MVUB}        & \xmark       & 0.832            & 0.752            & \textbf{0.744}            & 0.752               & 0.854            & \textbf{0.752}            & 0.772                   \\
    \textbf{Owlii}       & \xmark       & 0.498            & 0.465            & \textbf{0.456}            & \textbf{0.456}               & 0.512            & \textbf{0.465}            & 0.469                   \\
    \textbf{Thuman}      & \xmark       & 1.667            & 1.636            & \textbf{1.633}            & 1.634               & 1.677            & \textbf{1.636}            & 1.649                   \\
    \textbf{ScanNet}     & \xmark       & 1.922            & \textbf{1.826}            & 1.857            & 1.849               & 2.062            & \textbf{1.826 }           & 1.877                   \\
    \textbf{KITTI}       & \xmark       & 6.511            & 6.453            & 6.371            & \textbf{6.353}               & 6.877            & \textbf{6.453}            & 6.489                   \\
    \textbf{Ford}        & \xmark       & 8.632            & 8.565            & 8.459            & \textbf{8.452}               & 9.106            & \textbf{8.565}            & 8.599                   \\
    \textbf{Dense}       & \xmark       & 6.201            & \textbf{6.179}            & 6.194            & 6.195               & 6.275            & \textbf{6.179}            & 6.217                   \\
    \textbf{Sparse}      & \xmark       & 9.710            & \textbf{9.659}            & 9.690            & 9.731               & 9.728            & \textbf{9.659}            & 9.680                   \\
    \textbf{GS}          & \xmark       & 13.393           & \textbf{13.304}           & 13.333           & 13.330              & 13.647           & 13.304           & \textbf{13.248}                  \\
    \rowcolor{rowgray}
    \textbf{VGGT}        & \cmark       & 7.310            & 7.222            & \textbf{7.218}            & 7.250               & 7.320            & \textbf{7.222}            & 7.230                   \\
    \rowcolor{rowgray}
    \textbf{S3DIS}       & \cmark       & 11.454           & 11.317           & 11.166           & \textbf{11.163}              & 11.772           & 11.317           & \textbf{11.296}                  \\
    \rowcolor{rowgray}
    \textbf{RS}          & \cmark       & 3.674            & 3.643            & \textbf{3.637}            & 3.642               & 3.692            & 3.643            & \textbf{3.625}                   \\
    \rowcolor{rowgray}
    \textbf{NS}          & \cmark       & 4.887            & 4.863            & \textbf{4.840}            & 4.865               & 4.883            & 4.863            & \textbf{4.850 }                  \\
    \rowcolor{rowgray}
    \textbf{CS}          & \cmark       & 3.405            & 3.347            & \textbf{3.308}            & 3.335               & 3.548            & 3.347            & \textbf{3.322}                   \\ 
    \midrule
    \textbf{Avg bpp}     & --           & 5.380            & 5.320            & \textbf{5.298}           & 5.305               & 5.506            & \textbf{5.320}            & 5.327                   \\
    \textbf{Enc Time}     & --           &   0.492               &      \textbf{0.436}            &     0.571             &   0.568                  &     \textbf{0.382}            &          0.436        &      0.539                   \\
    \textbf{Dec Time}     & --           &   0.507               &      \textbf{0.445}            &     0.584             &   0.581                  &     \textbf{0.391}           &          0.445        &      0.551                   \\  
    \midrule[\heavyrulewidth]
    \multicolumn{1}{c|}{\cellcolor{headergray}} & \multicolumn{1}{c|}{\cellcolor{headergray}} & \multicolumn{4}{c|}{\cellcolor{headergray}\textbf{Spatial1 (Channel: 1 to 8)}} & \multicolumn{3}{c}{\cellcolor{headergray}\textbf{Channel1 (Spatial: 1 to 4)}} \\
    \multicolumn{1}{c|}{\cellcolor{headergray}\multirow{-2}{*}{\textbf{Dataset}}} & \multicolumn{1}{c|}{\cellcolor{headergray}\multirow{-2}{*}{\textbf{OOD}}} & \cellcolor{headergray}\textbf{1-group} & \cellcolor{headergray}\textbf{2-group} & \cellcolor{headergray}\textbf{4-group} & \cellcolor{headergray}\textbf{8-group}    & \cellcolor{headergray}\textbf{1-group} & \cellcolor{headergray}\textbf{2-group} & \cellcolor{headergray}\textbf{4-group} \\
    \midrule
    \textbf{8iVFB}       & \xmark       & 0.731            & 0.635            & 0.595            & \textbf{0.565}               & 0.731            & 0.604            & \textbf{0.581}                   \\
    \textbf{MVUB}        & \xmark       & 1.045            & 0.854            & 0.796            & \textbf{0.733}               & 1.045            & 0.832            & \textbf{0.782}                   \\
    \textbf{Owlii}       & \xmark       & 0.593            & 0.512            & 0.482            & \textbf{0.460}               & 0.593            & 0.498            & \textbf{0.473}                   \\
    \textbf{Thuman}      & \xmark       & 1.767            & 1.677            & 1.652            & \textbf{1.625}               & 1.767            & 1.667            & \textbf{1.646}                   \\
    \textbf{ScanNet}     & \xmark       & 2.196            & 2.062            & 1.986            & \textbf{1.909}               & 2.196            & 1.922            & \textbf{1.878}                   \\
    \textbf{KITTI}       & \xmark       & 7.048            & 6.877            & 6.571            & \textbf{6.390}               & 7.048            & 6.511            & \textbf{6.484}                   \\
    \textbf{Ford}        & \xmark       & 9.268            & 9.106            & 8.704            & \textbf{8.483}               & 9.268            & 8.632            & \textbf{8.599}                   \\
    \textbf{Dense}       & \xmark       & 6.436            & 6.275            & 6.314            & \textbf{6.217}               & 6.436            & 6.201            & \textbf{6.197}                   \\
    \textbf{Sparse}      & \xmark       & 9.810            & 9.728            & 9.804            & \textbf{9.794}               & 9.810            & 9.710            & \textbf{9.675}                   \\
    \textbf{GS}          & \xmark       & 13.839           & 13.647           & 13.650           & \textbf{13.570}              & 13.839           & 13.393           & \textbf{13.297}                  \\
    \rowcolor{rowgray}
    \textbf{VGGT}        & \cmark       & 7.529            & 7.320            & 7.320            & \textbf{7.265}               & 7.529            & 7.310            & \textbf{7.241}                   \\
    \rowcolor{rowgray}
    \textbf{S3DIS}       & \cmark       & 12.094           & 11.772           & 11.441           & \textbf{11.220}              & 12.094           & 11.454           & \textbf{11.312}                  \\
    \rowcolor{rowgray}
    \textbf{RS}          & \cmark       & 3.799            & 3.692            & 3.681            & \textbf{3.659}               & 3.799            & 3.674            & \textbf{3.629}                   \\
    \rowcolor{rowgray}
    \textbf{NS}          & \cmark       & 4.952            & \textbf{4.883}            & 4.892            & 4.932               & 4.952            & 4.887            &\textbf{ 4.835 }                  \\
    \rowcolor{rowgray}
    \textbf{CS}          & \cmark       & 3.740            & 3.548            & 3.481            & \textbf{3.411}              & 3.740            & 3.405            & \textbf{3.326}                   \\ 
    \midrule
    \textbf{Avg bpp}     & --           & 5.656            & 5.506            & 5.425            & \textbf{5.349}              & 5.656            & 5.380            & \textbf{5.330}                   \\
    \textbf{Enc Time}     & --           &  0.436                &       \textbf{0.382}           &    0.496              &     0.724                &    \textbf{0.436}              &        0.492          &       0.554                  \\
    \textbf{Dec Time}     & --           &  0.451                &       \textbf{0.391}           &    0.507              &     0.742                &    \textbf{0.451}              &        0.507          &       0.568                  \\  
    \bottomrule
    \end{tabular}
\end{table*}
\label{sec:contex_model_abla}
\begin{figure*}[t]
    \centering
    \begin{subfigure}{\textwidth}
        \includegraphics[width=0.24\textwidth]{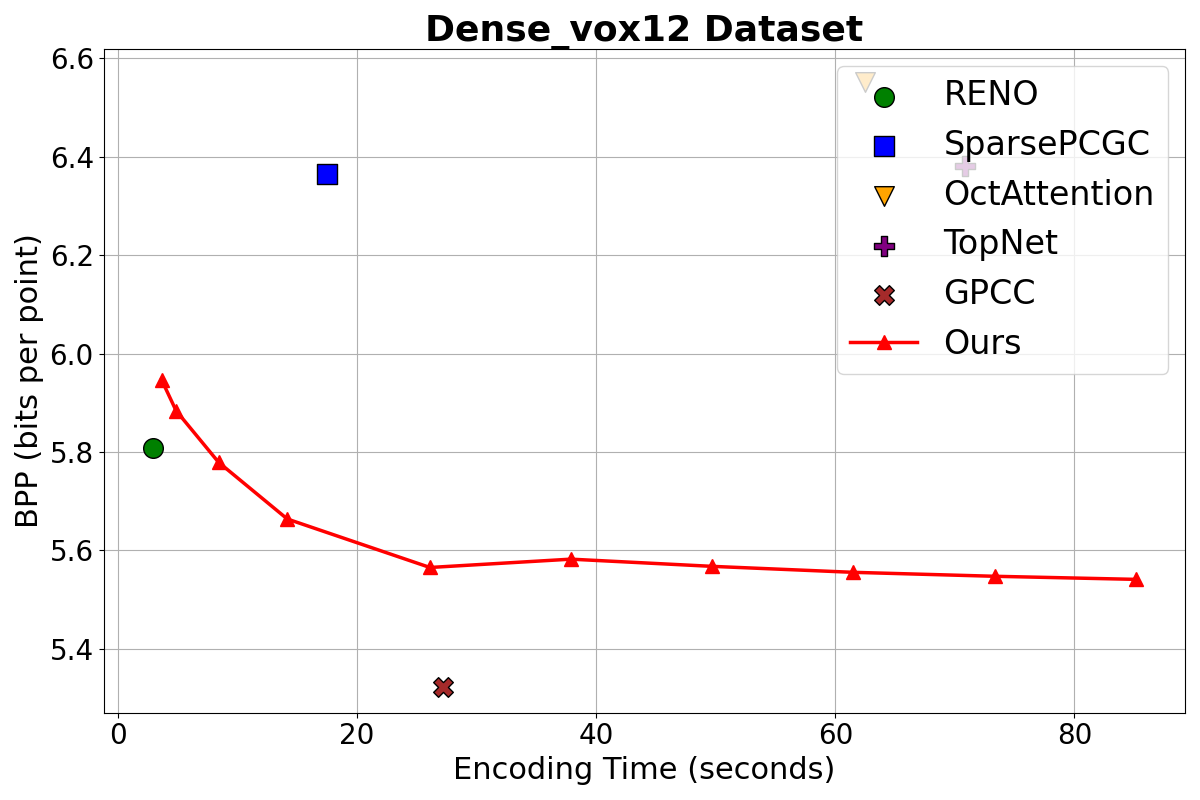}\hfill
        \includegraphics[width=0.24\textwidth]{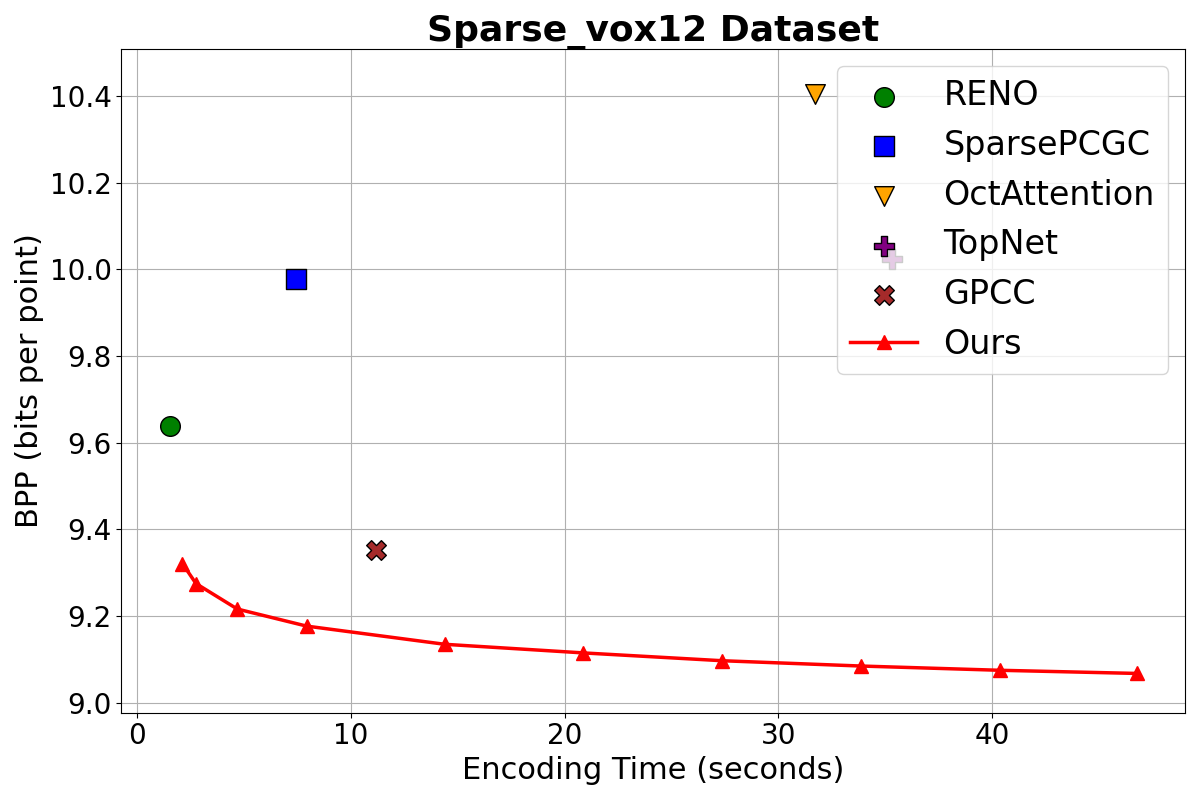}\hfill
        \includegraphics[width=0.24\textwidth]{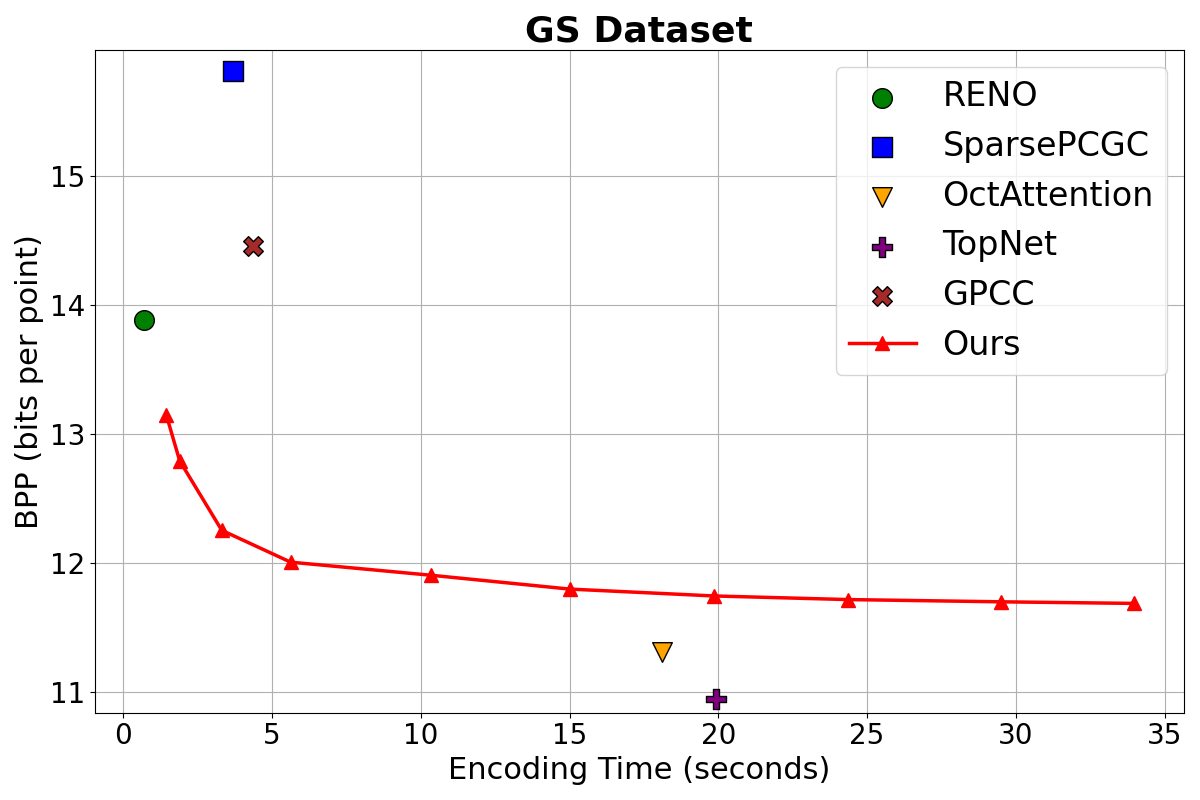}\hfill
        \includegraphics[width=0.24\textwidth]{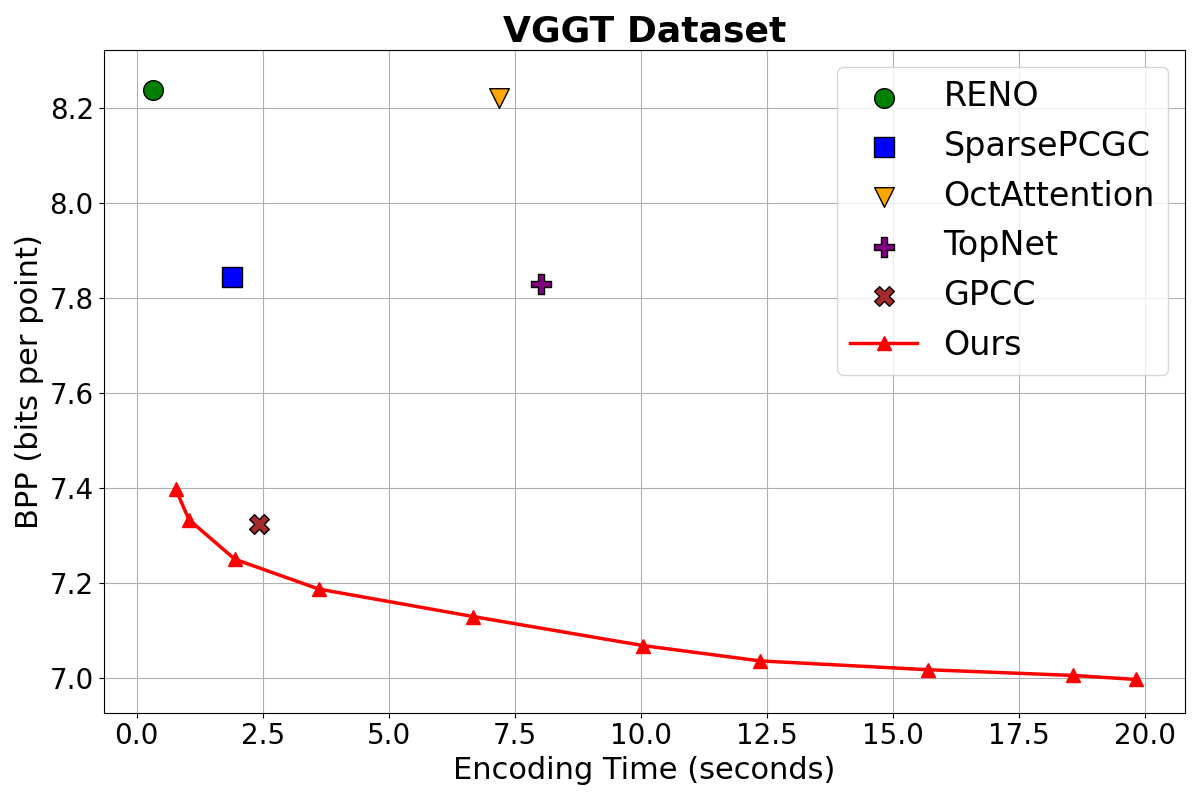}
        \includegraphics[width=0.24\textwidth]{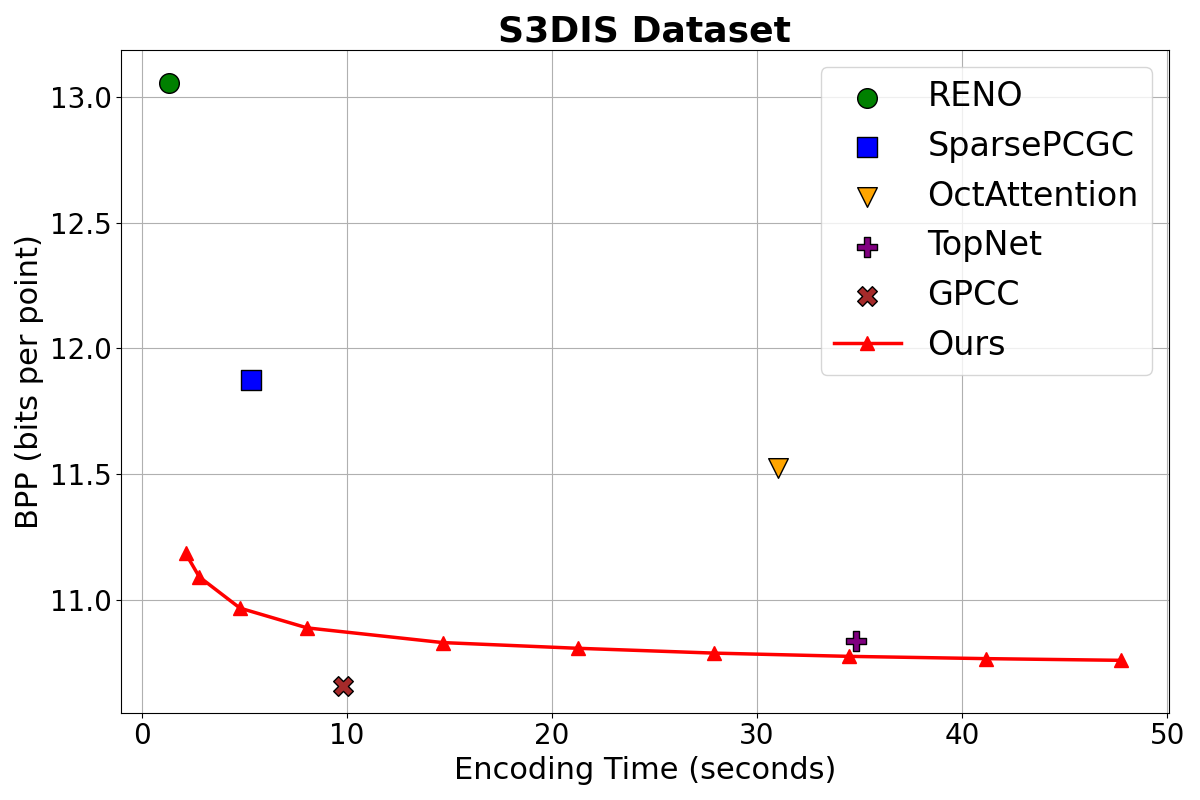}\hfill
        \includegraphics[width=0.24\textwidth]{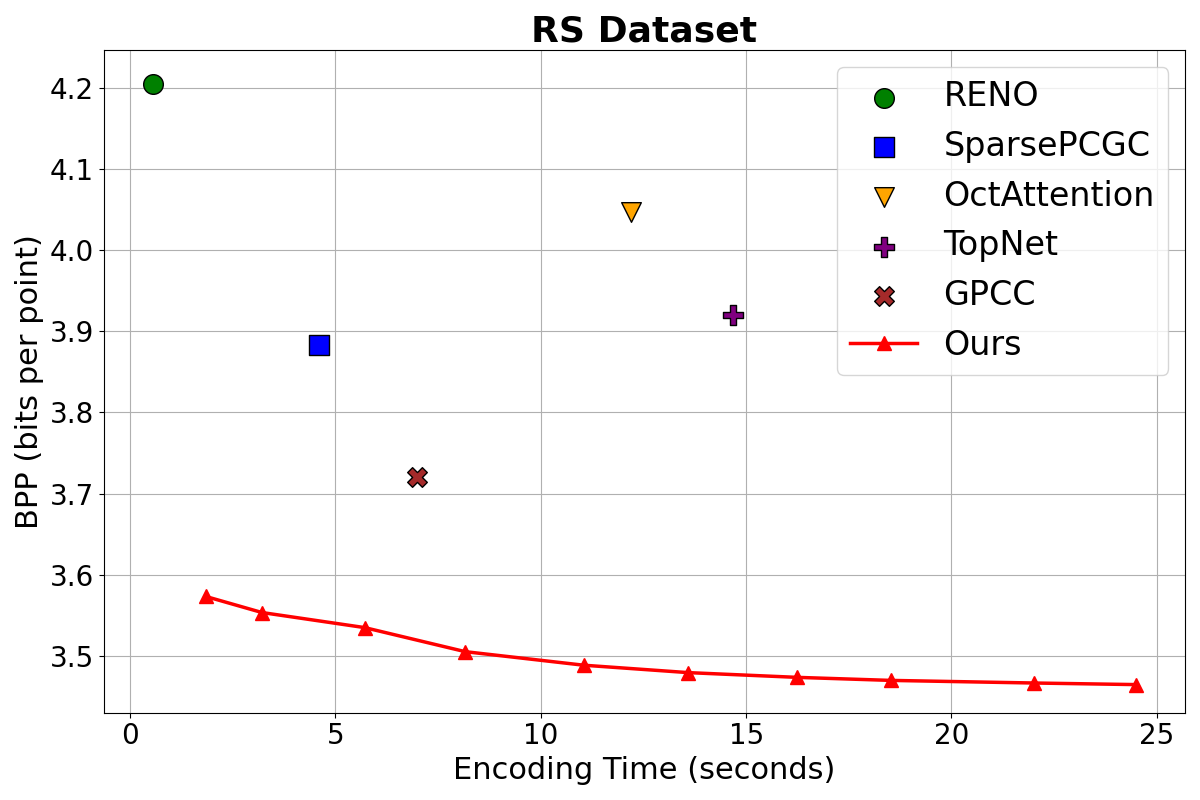}\hfill
        \includegraphics[width=0.24\textwidth]{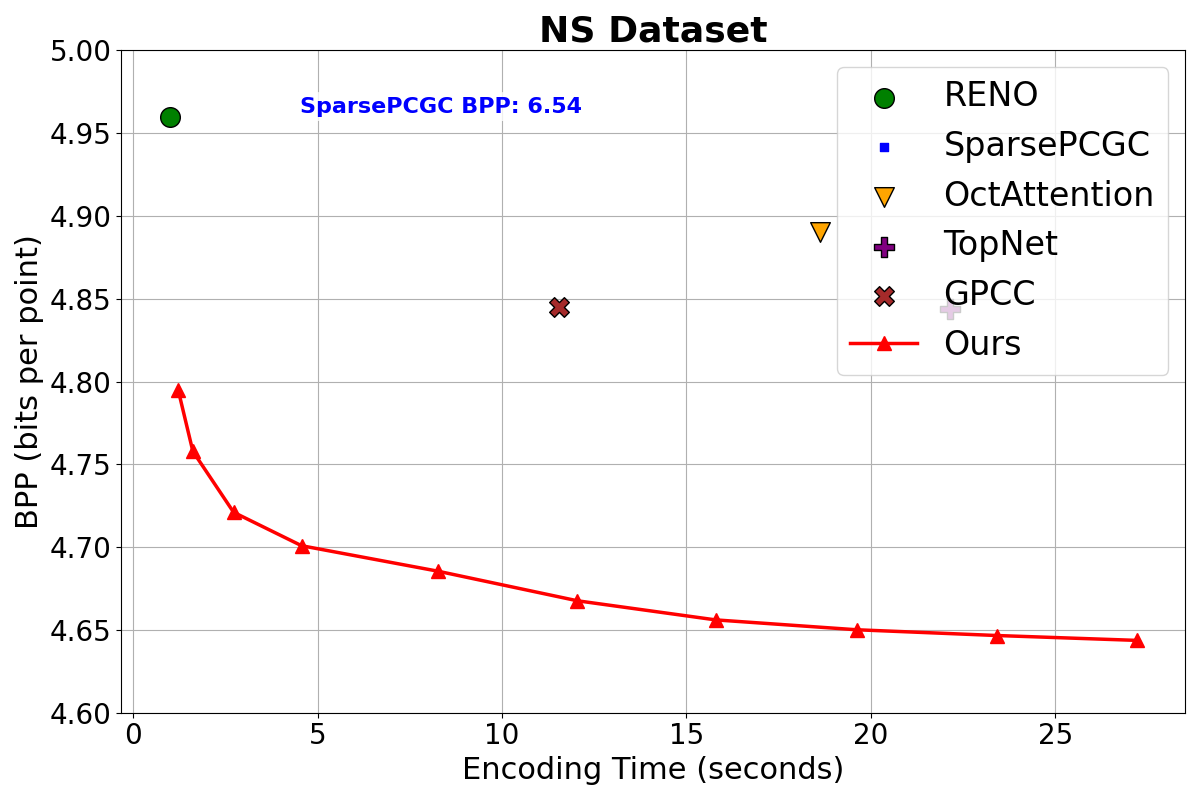}\hfill
        \includegraphics[width=0.24\textwidth]{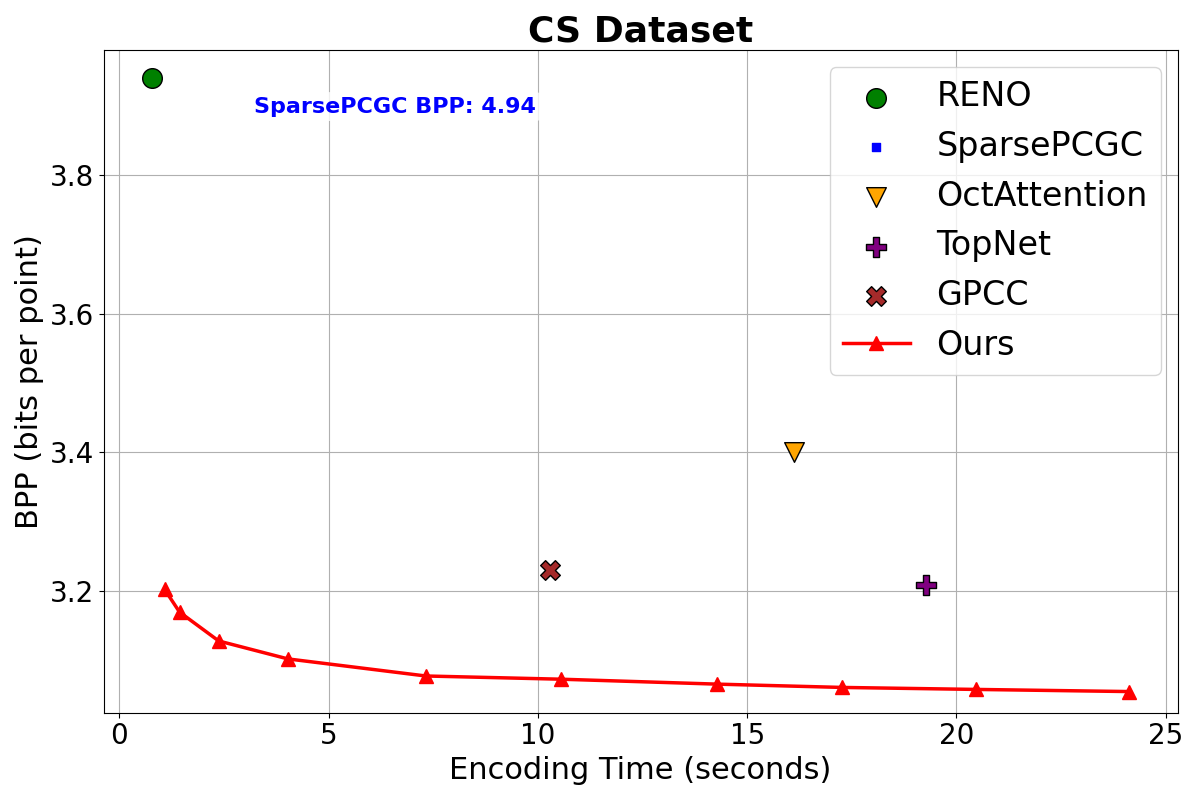}
    
    \end{subfigure}
    \caption{Trade-off between encoding complexity and performance. Our method allows for a flexible balance between these two factors. On the \texttt{NS} and \texttt{CS} datasets, the BPP value for SparsePCGC is annotated directly due to its significantly poor performance, ensuring visual clarity.}
    \label{fig:bpp_time}
\end{figure*}
\begin{figure*}[t]
    \centering
    \includegraphics[width=1\linewidth]{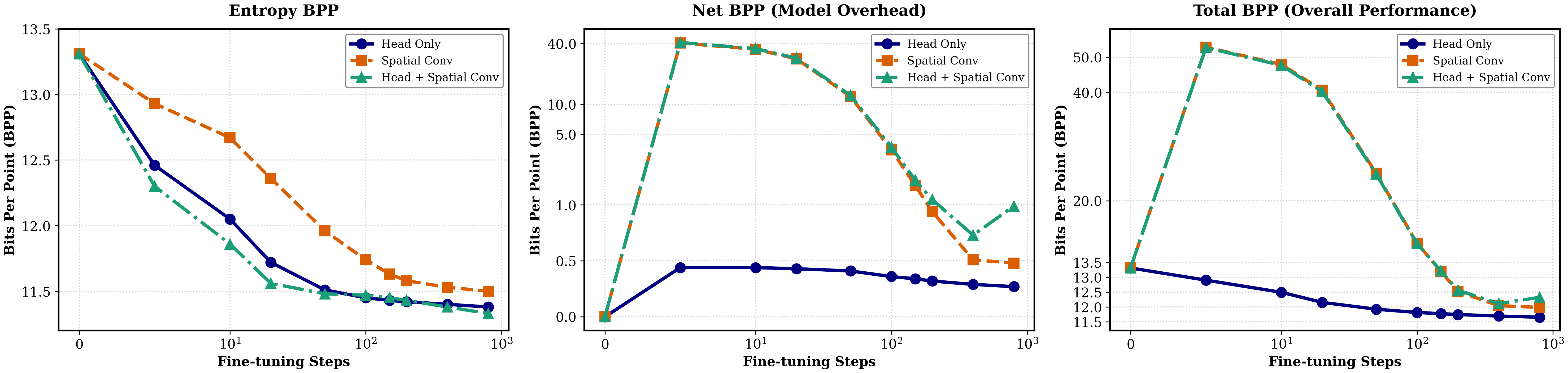}
    \caption{Impact of finetune steps and module sensitivity in GS dataset.}
    \label{fig:rebuttal_ablation}
\end{figure*}
We provide a comprehensive ablation study of our context model in Table~\ref{table:abla_contex}, investigating the impact of increasing the number of channel and spatial groups. Within our framework, we reimplement the core ideas of SparsePCGC~\cite{sparsepcgc} (uniform spatial grouping) and UniPCGC~\cite{unipcgc} (non-uniform spatial grouping). These correspond to our 8-group model in the Spatial1 experiment and our ue-4-group model in the Spatial2 experiment, respectively. It is important to note that these are conceptual replications and may differ from the original implementations. Our findings show that the non-uniform grouping of ue-4-group yields worse overall performance than uniform grouping, suggesting its applicability is limited to specific datasets. Furthermore, the uniform eight-group strategy of SparsePCGC is outperformed by our proposed hybrid approach (e.g., 2 spatial groups + multiple channel groups), demonstrating the superior efficiency of our context design.

Generally, when the number of groups in one dimension (either spatial or channel) is fixed, increasing the number of groups in the other dimension tends to improve performance. This is because a greater number of groups introduces a stronger inductive bias, which can simplify the learning process for the model. An interesting exception occurs when the number of channel groups is fixed at 2: increasing the spatial groups from 2 to 4 leads to a performance degradation on in-distribution datasets. We hypothesize that at the occupancy code scale, an excessive number of spatial groups creates too large a gap between contexts, diminishing the relevance of the reference information and thus hindering performance. Conversely, when starting from a single group in either dimension, increasing the group count consistently improves performance.

It is also crucial to consider the trade-off between performance and complexity. For spatial grouping, a higher number of groups necessitates more sequential auto-regressive decoding steps, which increases both encoding and decoding time. A similar trade-off exists for channel grouping. However, a special case arises with a single channel group: this requires the arithmetic coder to handle a 256-way categorical distribution at once, which is computationally expensive and slows down the coding process. This observation is consistent with findings in RENO. As the number of channel groups increases from 2 to 8, performance generally improves at the cost of increased computational complexity.
Considering these trade-offs, a configuration with 2 spatial groups and multiple channel groups emerges as a favorable choice. For our final model, we select 2 spatial and 2 channel groups. We do not opt for 4 or 8 channel groups because they demand significantly more GPU memory for training and increase decoding complexity. In contrast, the 2-group configuration provides a more lightweight and efficient solution.

The analysis above considers performance averaged across all datasets. However, the detailed breakdown in Table \ref{table:abla_contex} offers a novel insight: different context designs excel on different types of point clouds. This comprehensive ablation provides a valuable reference for future research, highlighting that the optimal context modeling strategy is likely data-dependent.

\paragraph{Fine-tuning Iteration and Module Sensitivity.}
\label{sec:fine-tune time}
In this subsection, we demonstrate the effectiveness of our Instance-Adaptive Fine-Tuning (IAFT) strategy, particularly for out-of-distribution (OOD) data.
As illustrated in Figure~\ref{fig:bpp_time}, the points along the red curve represent the model's performance at various fine-tuning iterations, ranging from 10 to 1200.
The results clearly show that the model converges most rapidly during the initial stages, yielding the most significant bitrate reduction.
After approximately 200 iterations, the performance gains begin to saturate, with further improvements becoming marginal.
We therefore select this as our default fine-tuning stopping point.
This highlights a key advantage of our method: it offers a dynamic trade-off between encoding time and compression performance.
In resource-constrained scenarios, one can opt for fewer iterations to achieve faster encoding, sacrificing a small amount of compression efficiency.
Conversely, when computational resources are readily available, a longer fine-tuning duration can be used to maximize compression performance.
Our framework thus enables a controllable complexity-performance trade-off.
Importantly, even with zero fine-tuning, our base model already outperforms several state-of-the-art methods on many datasets, underscoring the strong foundational performance of our approach.

\begin{figure*}[t]
    \centering
    \includegraphics[width=1\linewidth]{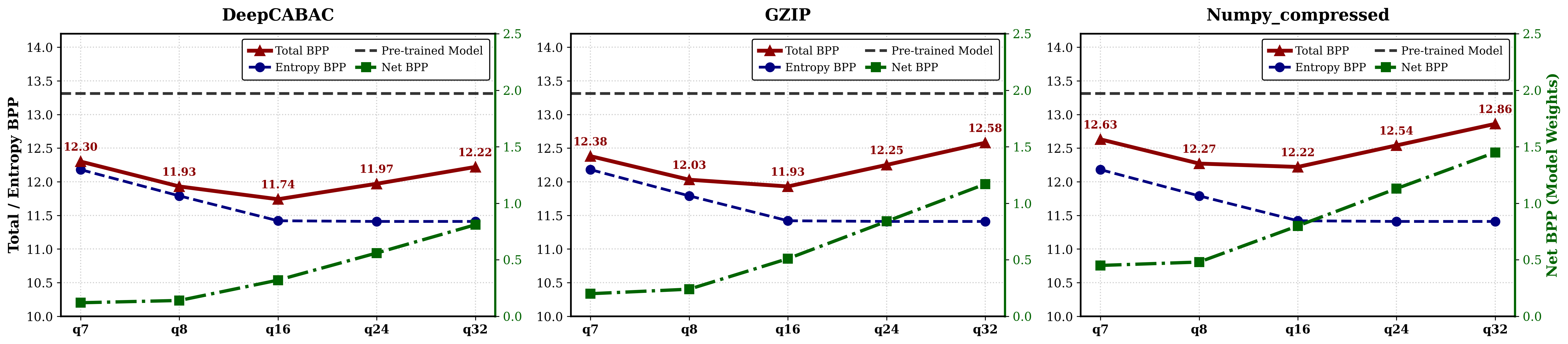}
    \caption{Impact of weight quantization and coder selection in GS dataset.}
    \label{fig:rebuttal_quan}
\end{figure*}

Moreover, We performed ablation studies on IAFT, as shown in Figure~\ref{fig:rebuttal_ablation}. Just \textbf{10 steps} of IAFT yield significant gains on OOD data, confirming robustness under low latency constraints. Extending fine-tuning to Spatial Convolutions incurs a \textbf{15$\times$ time increase} and higher weight overhead without proportional coding gains. Thus, fine-tuning only the lightweight head offers the optimal trade-off.

\paragraph{Quantization \& Coder Selection.} 
To efficiently compress the network weights transmitted to the decoder, we employ uniform scalar quantization with a step size $\Delta = 2^{-0.5 \times q}$. As illustrated in Figure~\ref{fig:rebuttal_quan}, varying the quantization parameter $q$ reveals a clear trade-off: larger $q$ values preserve finer weight details, which effectively lowers the point cloud's entropy coding bitrate but simultaneously increases the weight payload. Consequently, the Total BPP exhibits a stable, gentle U-shaped trend, reaching an optimal balance at $q=16$ across all tested coders. 
Crucially, all quantization settings consistently outperform the pre-trained model baseline. Furthermore, we observe that our method is highly robust to the choice of weight coders. At the optimal $q16$ setting, a standard \texttt{GZIP} coder achieves a Total BPP of 11.93, which closely matches the 11.74 BPP achieved by the computationally heavier \texttt{DeepCABAC}. This demonstrates that the significant coding gains primarily derive from our proposed strategy itself, rather than relying on complex parameter entropy coders.

\paragraph{Training Data Ratio.}
\label{sec:training data ratio}
\begin{figure*}[t]
    \centering
    \includegraphics[width=1\linewidth]{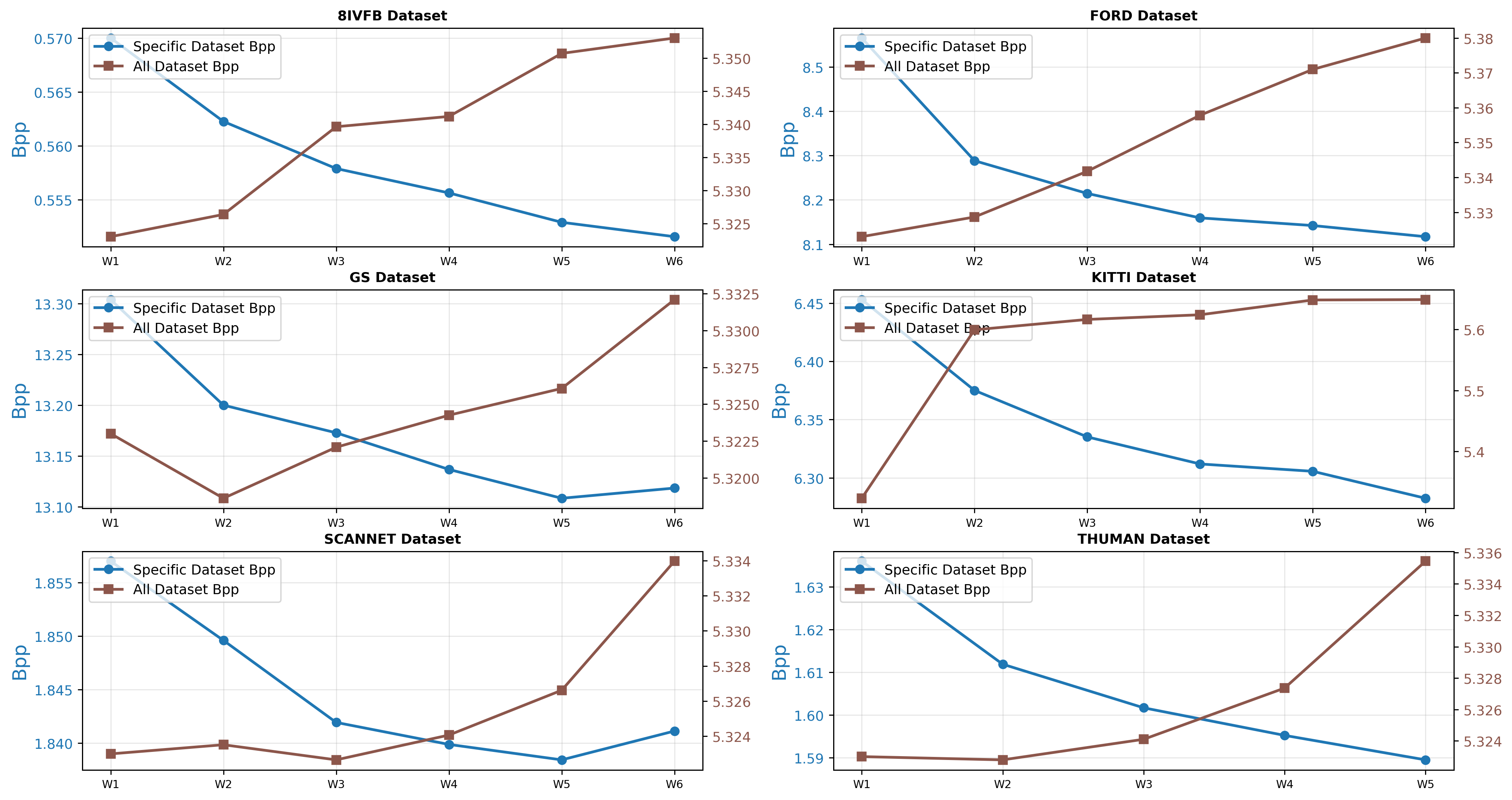}
    \caption{A comparative analysis of compression performance on the Specific Dataset and the All Dataset as a function of training set composition. The $w_1$  model is trained on a balanced dataset. For models $w_2$  and beyond, the proportion of the specific subset in the training data is systematically increased. The blue and orange-brown lines illustrate the model's performance on the Specific Dataset and the All Dataset, respectively.}
    \label{fig:weight}
\end{figure*}
We train the Ours-U model on a mixed dataset comprising \texttt{8iVFB}, \texttt{Thuman}, \texttt{ScanNet}, \texttt{KITTI}, \texttt{Ford}, and a general \texttt{GS} dataset.
In this section, we investigate the impact of the training data mixing ratio on the final model performance.
We acknowledge that determining an analytically optimal mixing ratio is intractable due to the infinite solution space.
Therefore, our objective is to identify an empirically effective and well-balanced configuration.
To this end, we systematically adjust the sampling weight of key subsets within the total training data, generating multiple training configurations.
The performance of each configuration is then quantitatively analyzed by evaluating the resulting model's bitrate on both the specific target dataset and our comprehensive test suite (denoted ``All Datasets'').

We begin with a baseline weight configuration, denoted as $w_1$ : \{\texttt{8iVFB}: 2.5, \texttt{Thuman}: 1.0, \texttt{ScanNet}: 1.0, \texttt{KITTI: 2.5}, \texttt{Ford}: 1.0, \texttt{GS}: 2.0\}.
We note that during training, these weights are normalized to form a probability distribution from which datasets are randomly sampled for each training batch.
As Figure \ref{fig:weight} depicts, $w_1$ represents this initial configuration.
The subsequent configurations, $w_2$ to $w_6$, are generated by incrementally increasing the weight of the target dataset by 2.0 for each step.
The results illustrate a clear trend: as the sampling weight of a specific dataset increases, the model's performance on that particular dataset improves.
However, this is often accompanied by a decrease in performance on the comprehensive benchmark.
This outcome validates our initial choice of $w_1$ as a reasonable and well-balanced configuration for general-purpose compression.

\subsection{Practical Deployment and Latency Analysis.} 
\label{sec:prac_deploy}
It is important to note that the encoding latency of 2.84s reported in Table~\ref{table:all_method} represents the extreme upper bound of our framework, corresponding to the maximum fine-tuning setting (IAFT-200) designed strictly for peak rate-distortion performance. In real-world deployment, our approach offers high flexibility to balance efficiency and coding gain:
\noindent (1) \textit{Standard Benchmarks:} For standard autonomous driving scenarios like the KITTI dataset, our pre-trained model exhibits strong generalization capabilities. It achieves a superior 6.18 BPP compared to RENO's 7.06 BPP without any instance-specific fine-tuning. Under this configuration, the encoding latency is merely 0.28s, which fully satisfies the stringent low-latency requirements of practical applications.
\noindent (2) \textit{Favorable Efficiency Trade-off:} The proposed IAFT module converges rapidly. By limiting the fine-tuning process to just 10 iterations, the latency on the dense GS dataset drastically drops to 1.35s, effectively lowering the overall average encoding time across all datasets to a highly competitive \textbf{0.56s}. Even with this accelerated configuration, our method consistently maintains a clear performance margin over existing baselines.
\noindent (3) \textit{Standardization Compatibility:} From a system-level perspective, the updated IAFT weights are highly compact and seamlessly transmitted as standard side information. This architectural design aligns perfectly with existing point cloud compression standards (e.g., MPEG and AVS pipelines), ensuring straightforward integration into established industrial systems.

\section{Point Cloud Visualization }
\begin{figure*}[t]
    \centering
    
    \begin{subfigure}{0.3\textwidth}
        \centering
        \includegraphics[width=\textwidth]{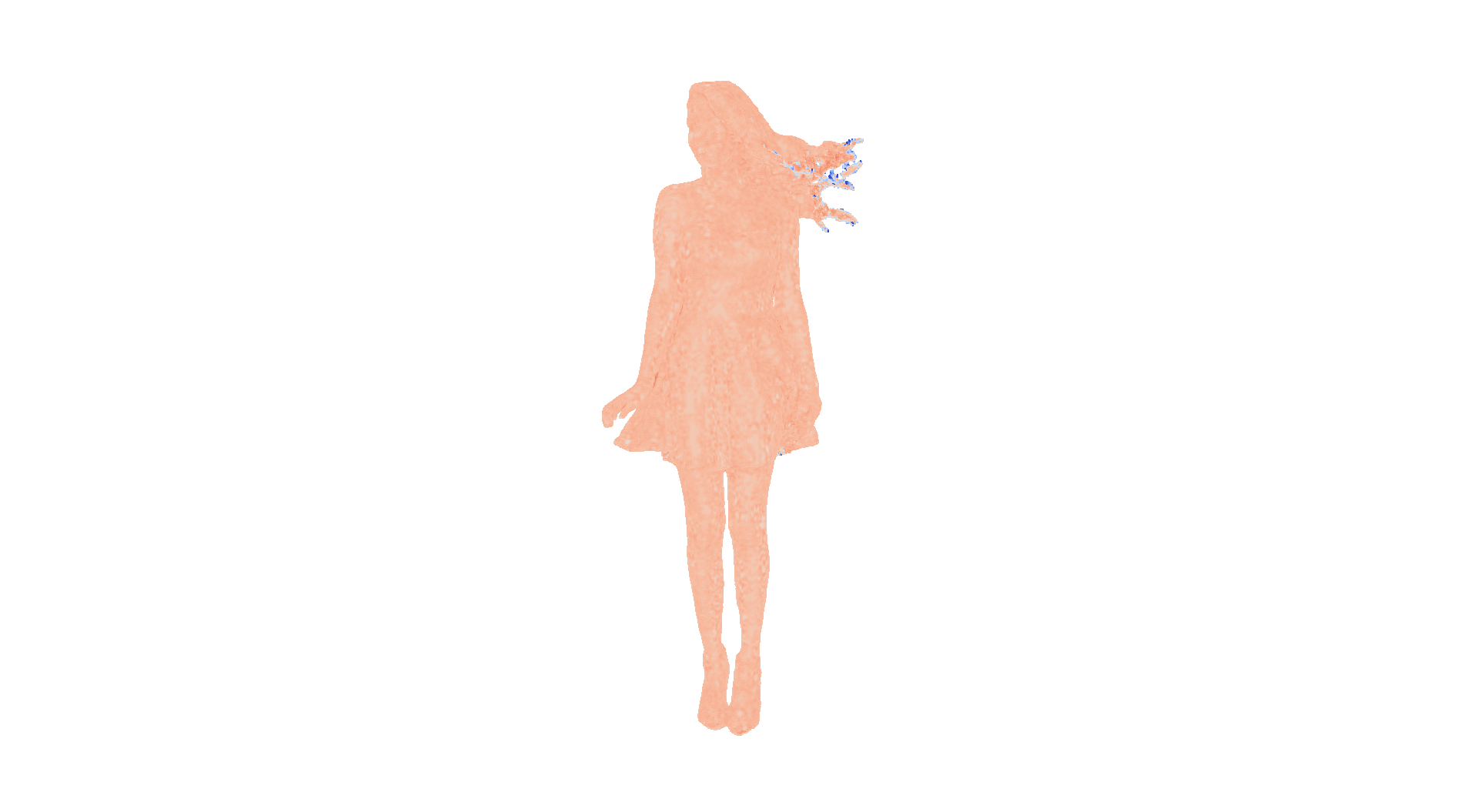}
        \caption{8iVFB}
    \end{subfigure}
    \hspace{0.02\textwidth}
    \begin{subfigure}{0.3\textwidth}
        \centering
        \includegraphics[width=\textwidth]{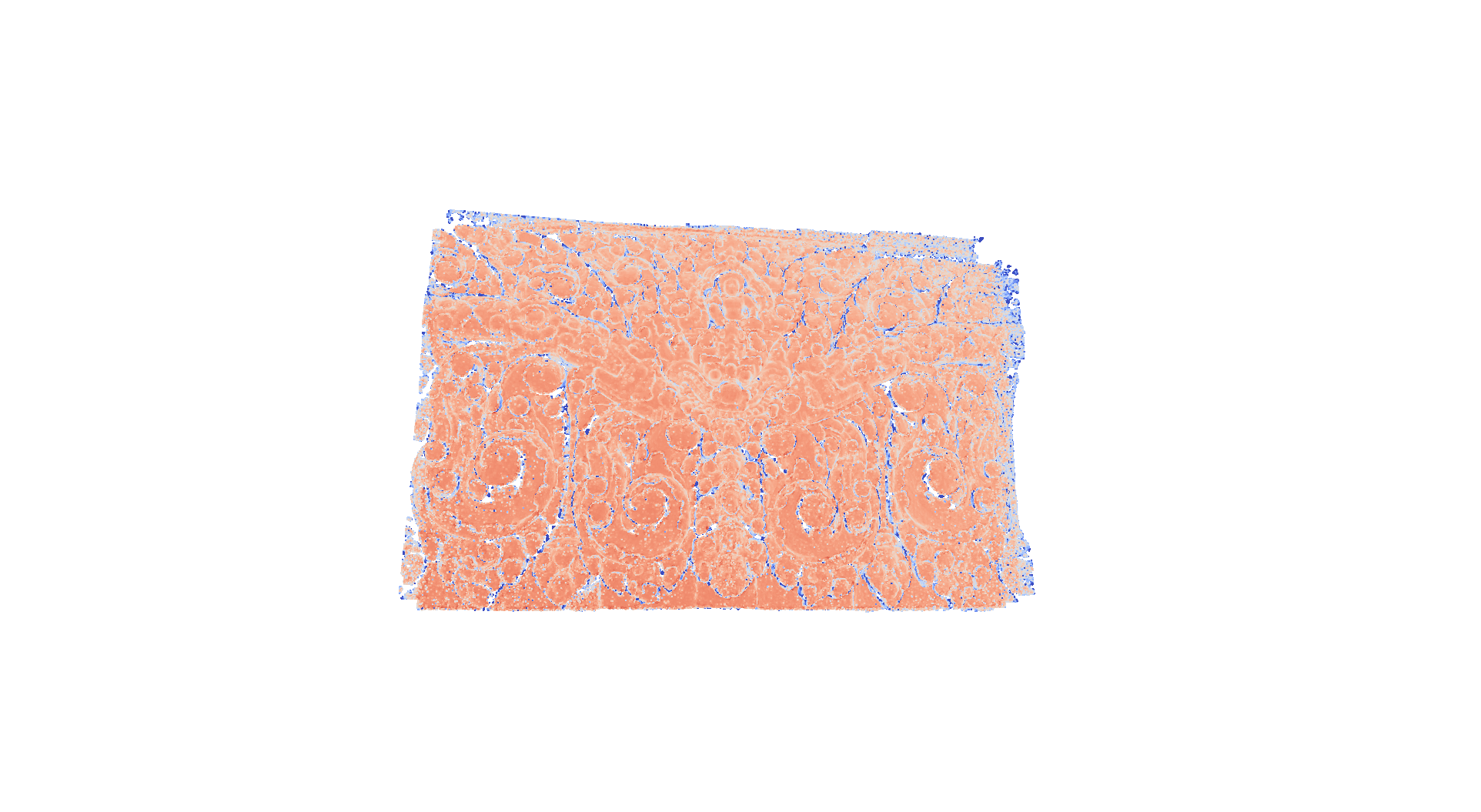}
        \caption{Dense-vox12}
    \end{subfigure}
    \hspace{0.02\textwidth}
    \begin{subfigure}{0.3\textwidth}
        \centering
        \includegraphics[width=\textwidth]{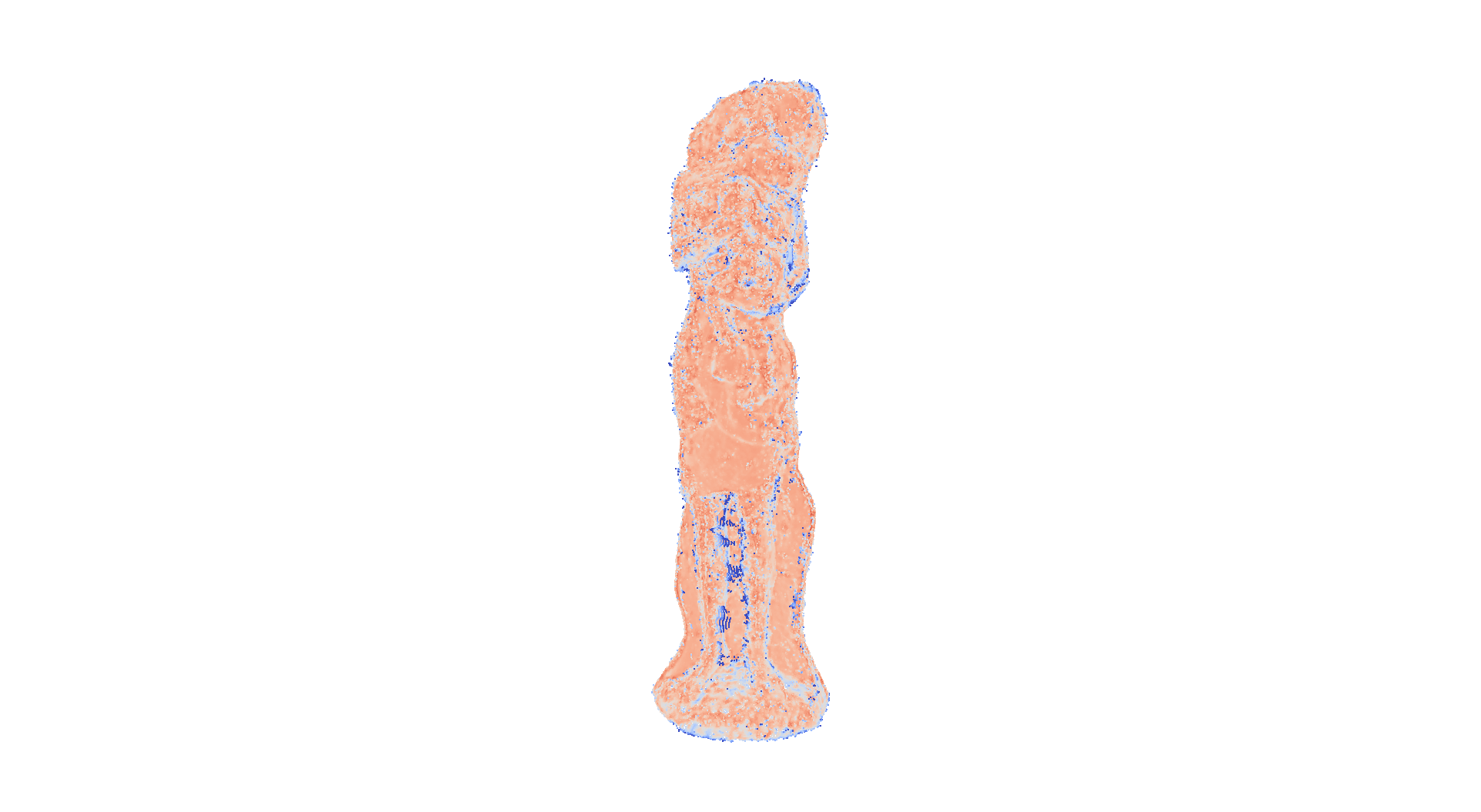}
        \caption{Sparse-vox12}
    \end{subfigure}
    
    \vspace{0.3em}
    
    \begin{subfigure}{0.3\textwidth}
        \centering
        \includegraphics[width=\textwidth]{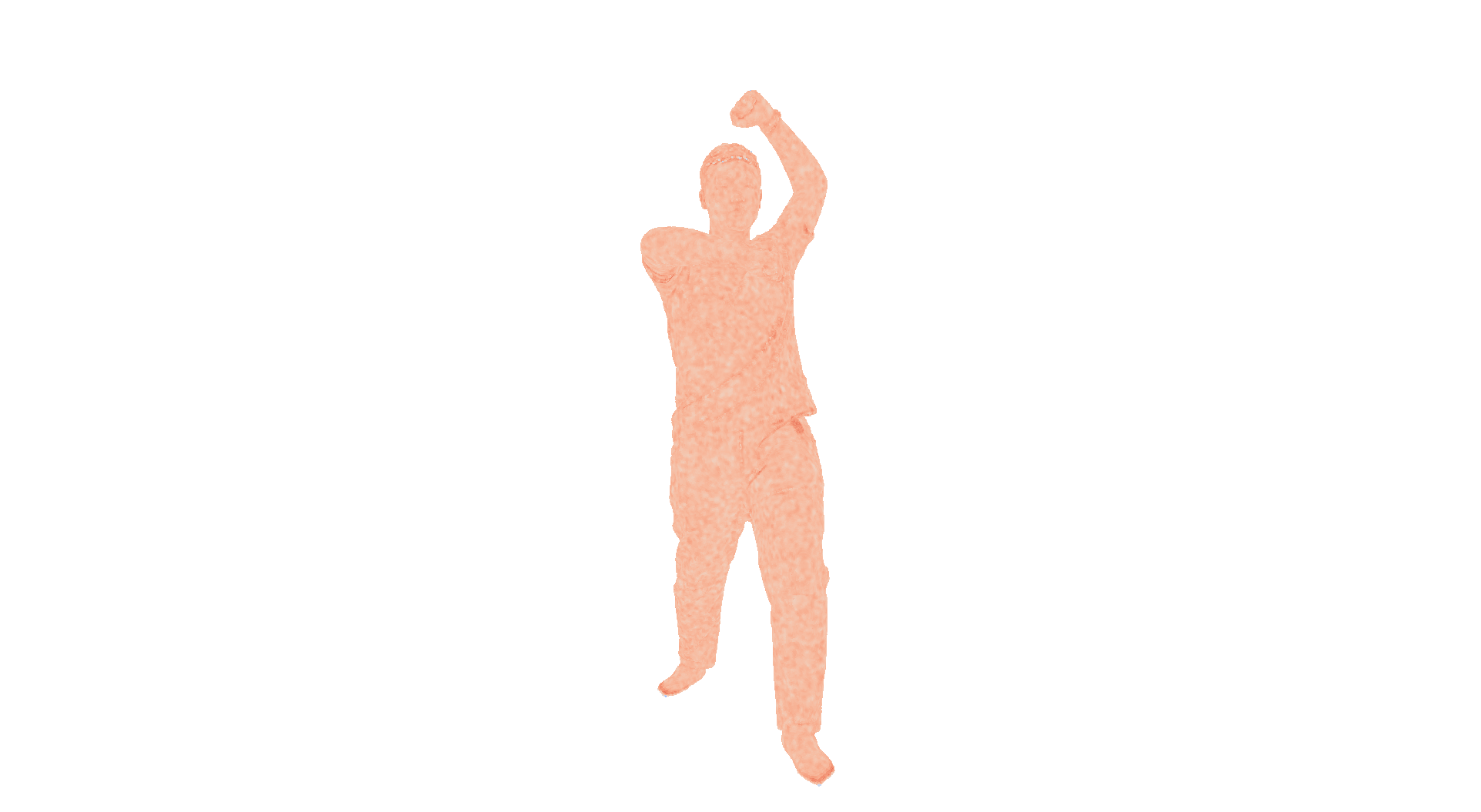}
        \caption{Thuman}
    \end{subfigure}
    \hspace{0.02\textwidth}
    \begin{subfigure}{0.3\textwidth}
        \centering
        \includegraphics[width=\textwidth]{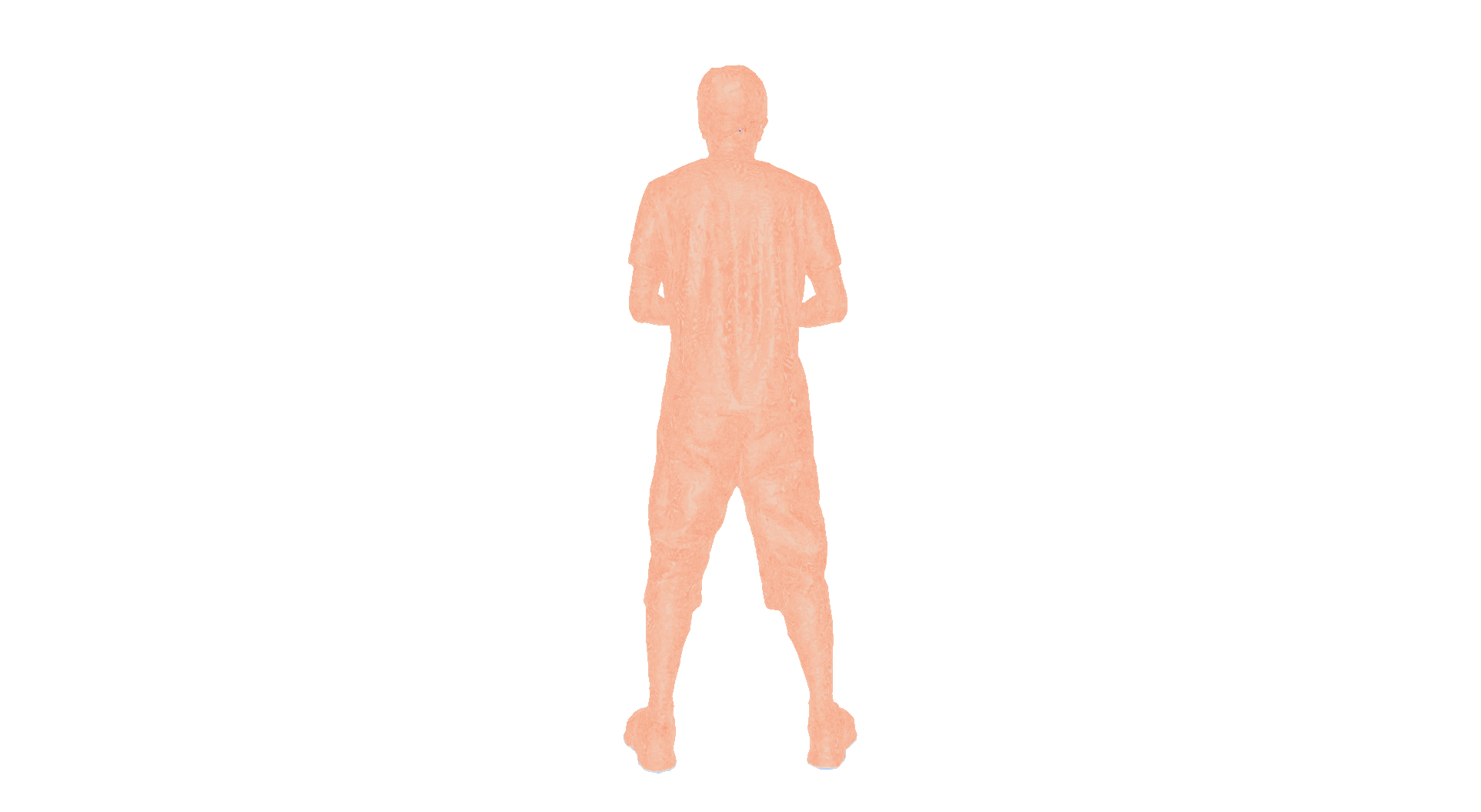}
        \caption{Owlii}
    \end{subfigure}
    \hspace{0.02\textwidth}
    \begin{subfigure}{0.3\textwidth}
        \centering
        \includegraphics[width=\textwidth]{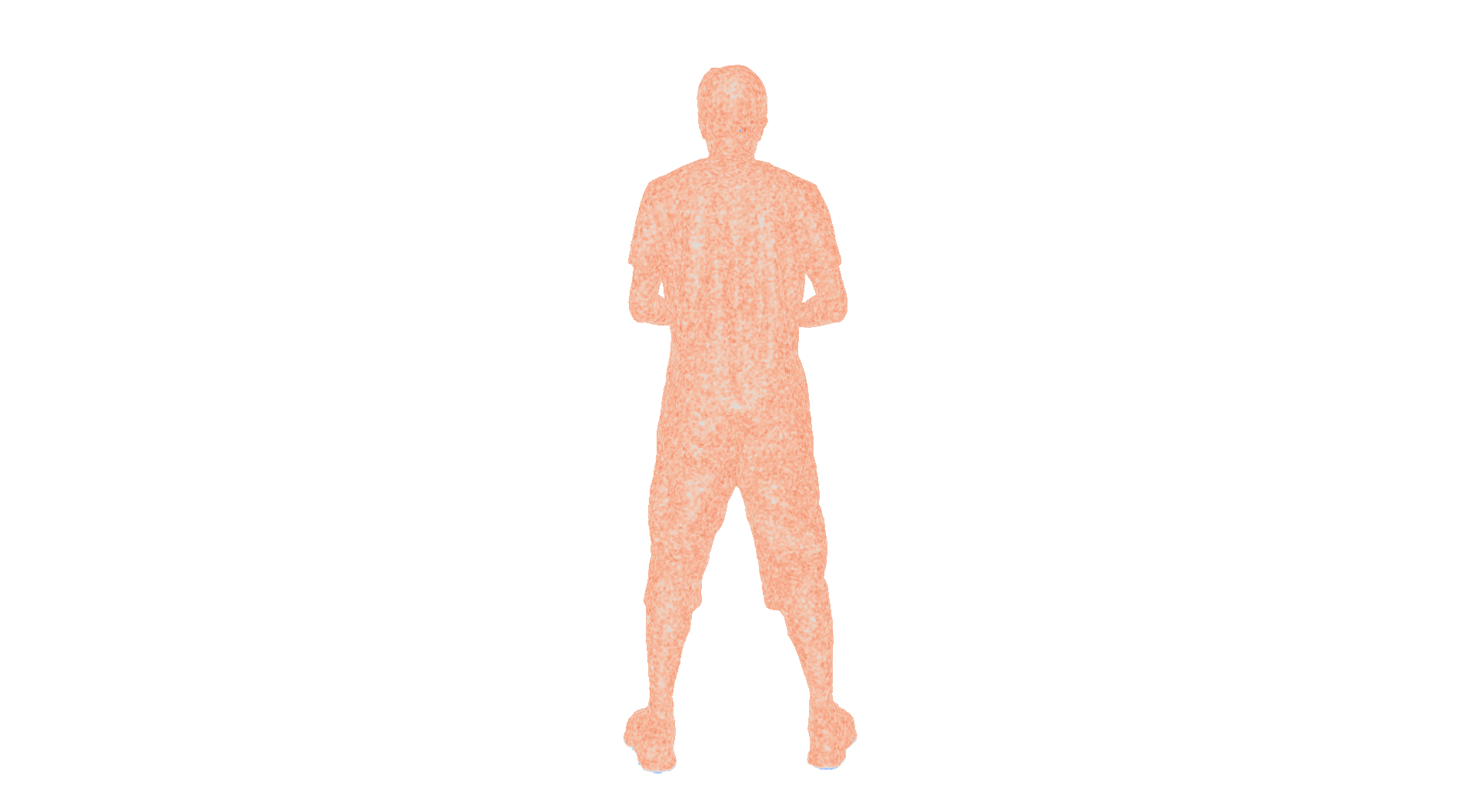}
        \caption{RS}
    \end{subfigure}
    
    \vspace{0.3em}
    
    \begin{subfigure}{0.3\textwidth}
        \centering
        \includegraphics[width=\textwidth]{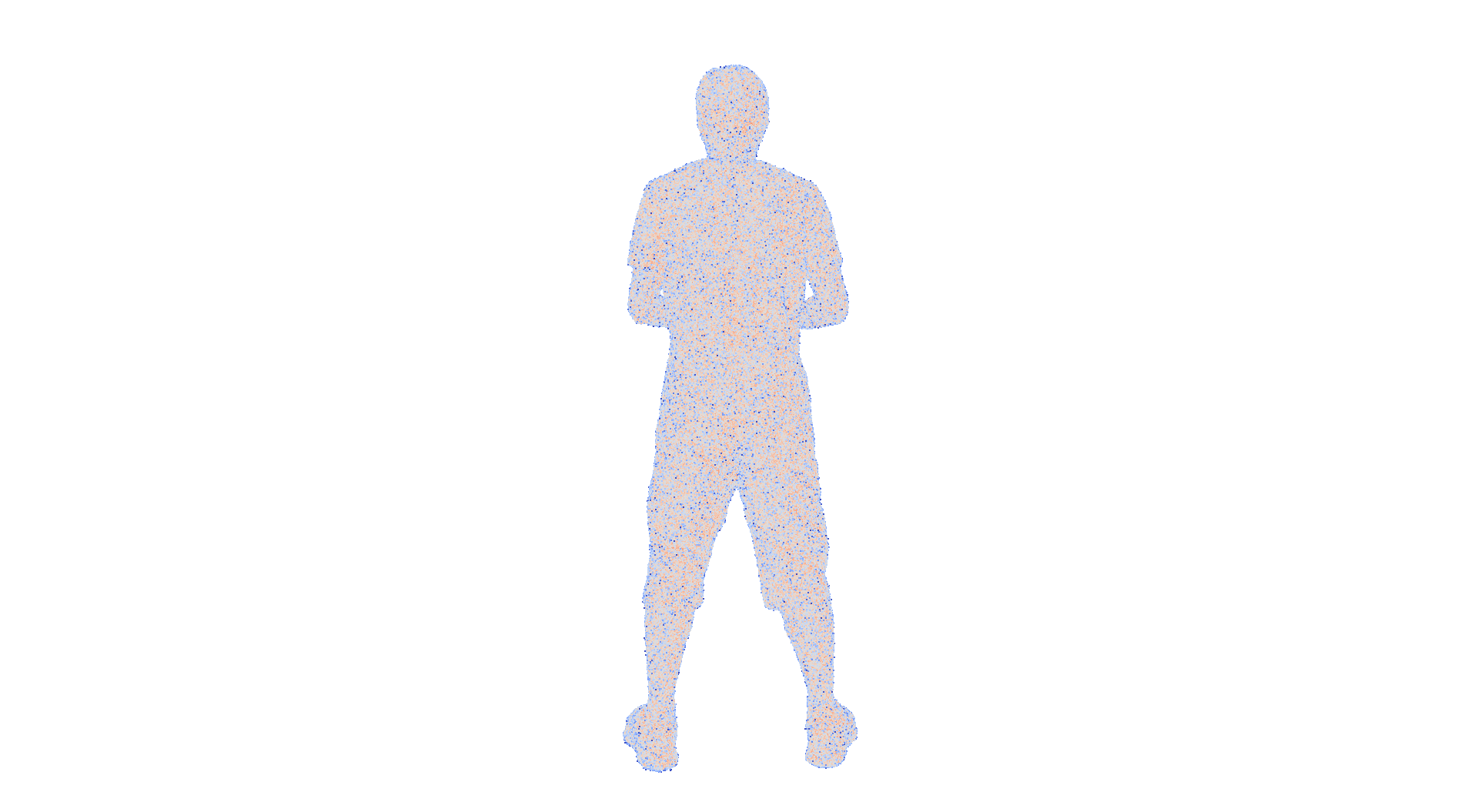}
        \caption{NS}
    \end{subfigure}
    \hspace{0.02\textwidth}
    \begin{subfigure}{0.3\textwidth}
        \centering
        \includegraphics[width=\textwidth]{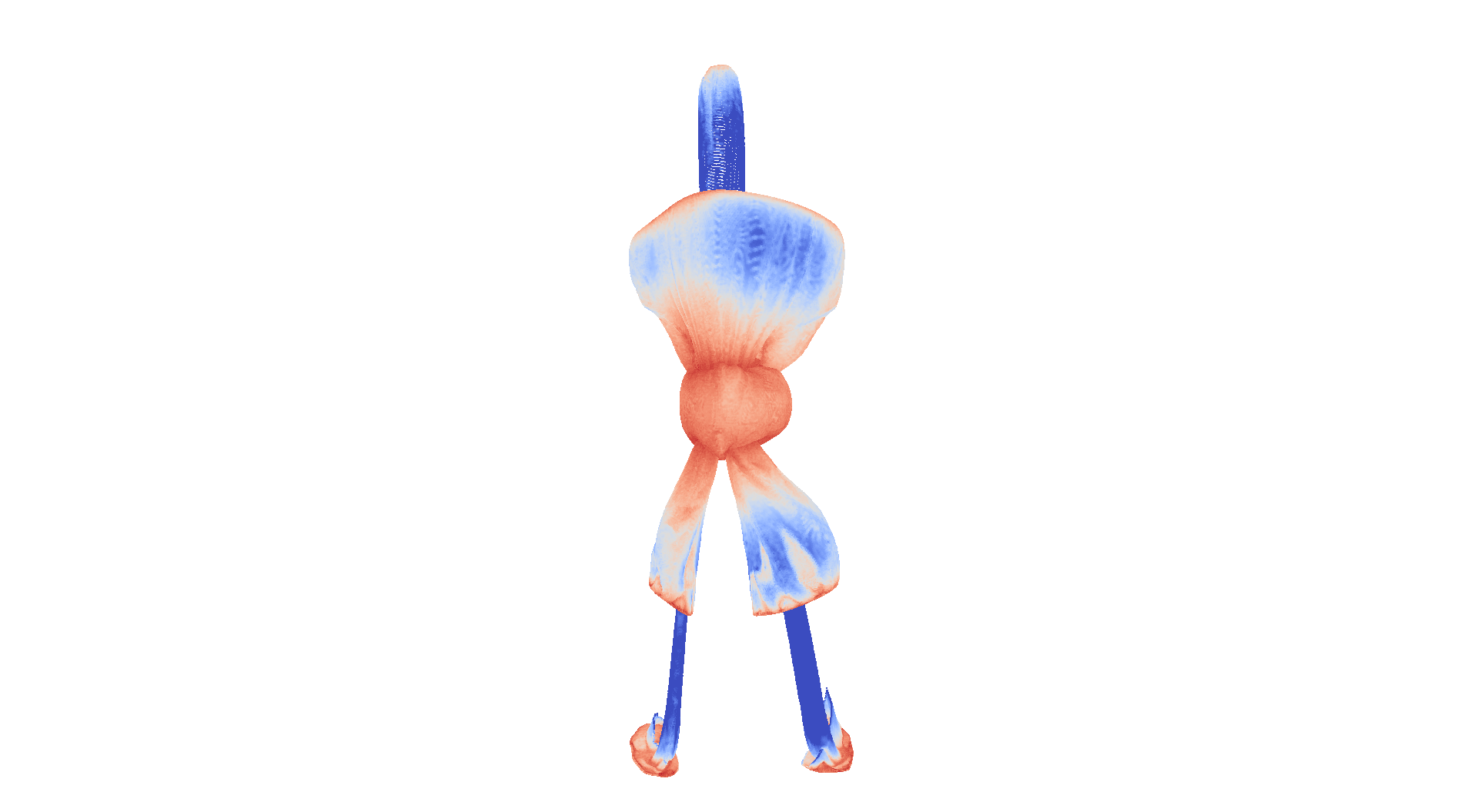}
        \caption{CS}
    \end{subfigure}
    \hspace{0.02\textwidth}
    \begin{subfigure}{0.3\textwidth}
        \centering
        \includegraphics[width=\textwidth]{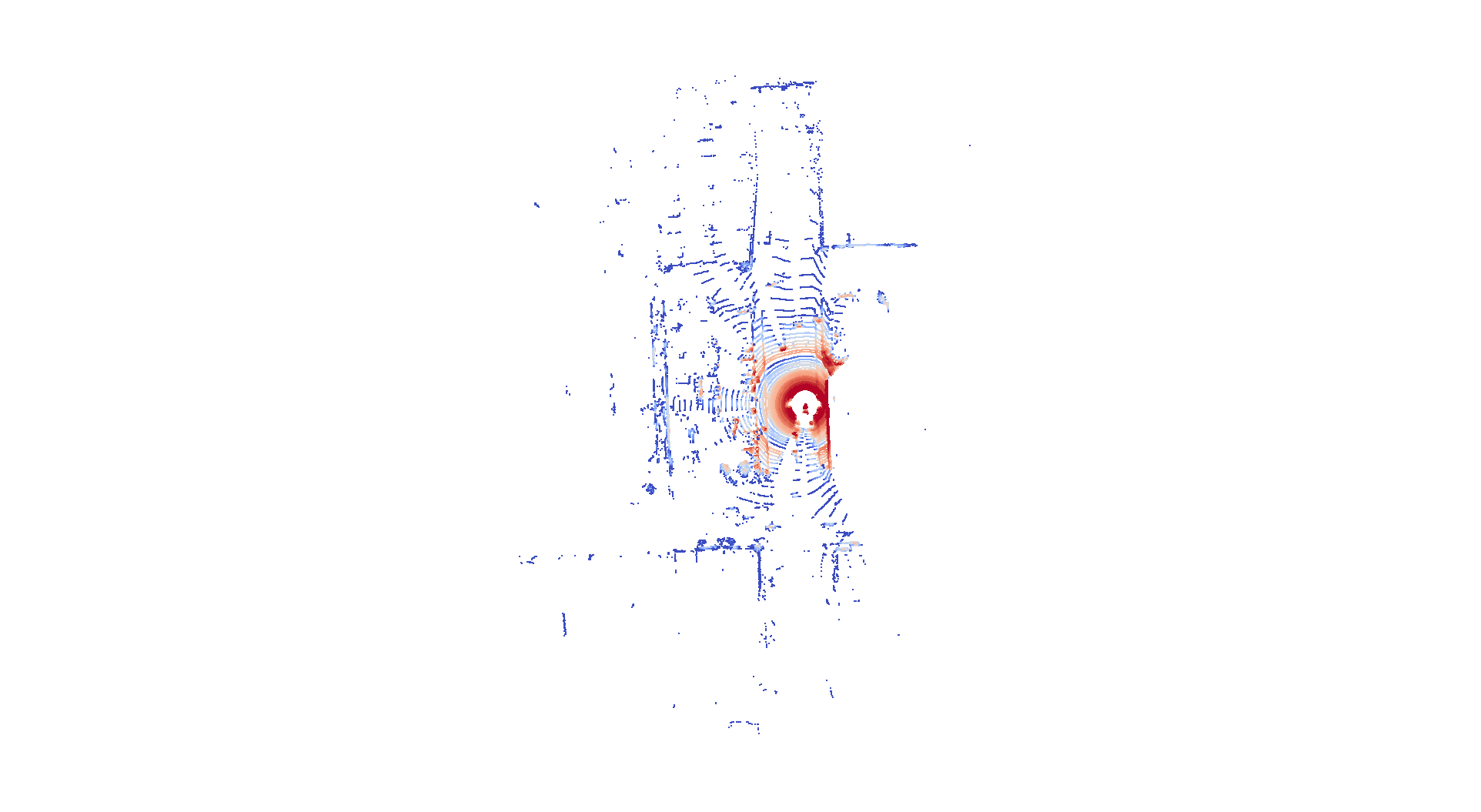}
        \caption{Ford}
    \end{subfigure}
    
    \vspace{0.3em}
    
    \begin{subfigure}{0.3\textwidth}
        \centering
        \includegraphics[width=\textwidth]{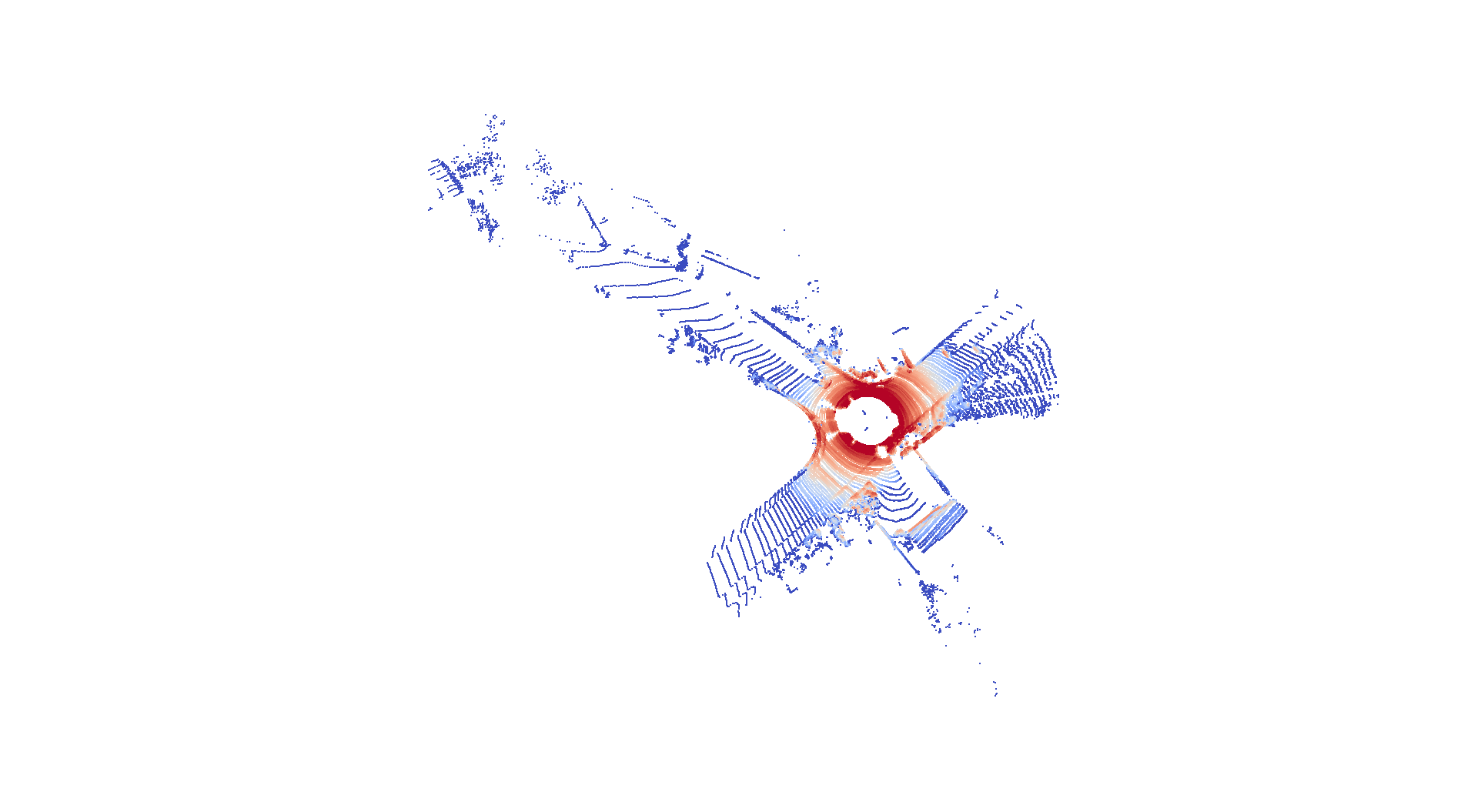}
        \caption{KITTI}
    \end{subfigure}
    \hspace{0.02\textwidth}
    \begin{subfigure}{0.3\textwidth}
        \centering
        \includegraphics[width=\textwidth]{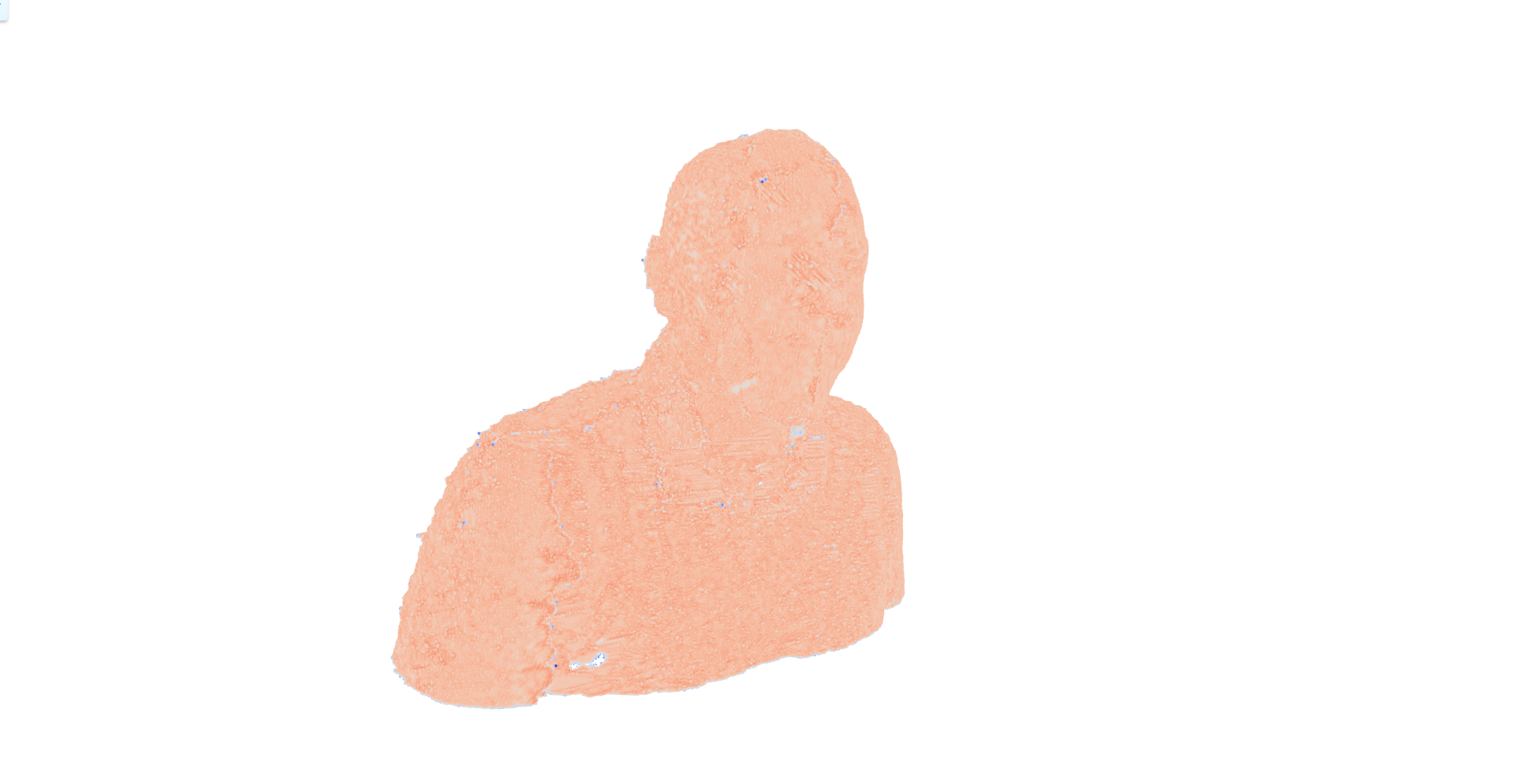}
        \caption{MVUB}
    \end{subfigure}
    \hspace{0.02\textwidth}
    \begin{subfigure}{0.3\textwidth}
        \centering
        \includegraphics[width=\textwidth]{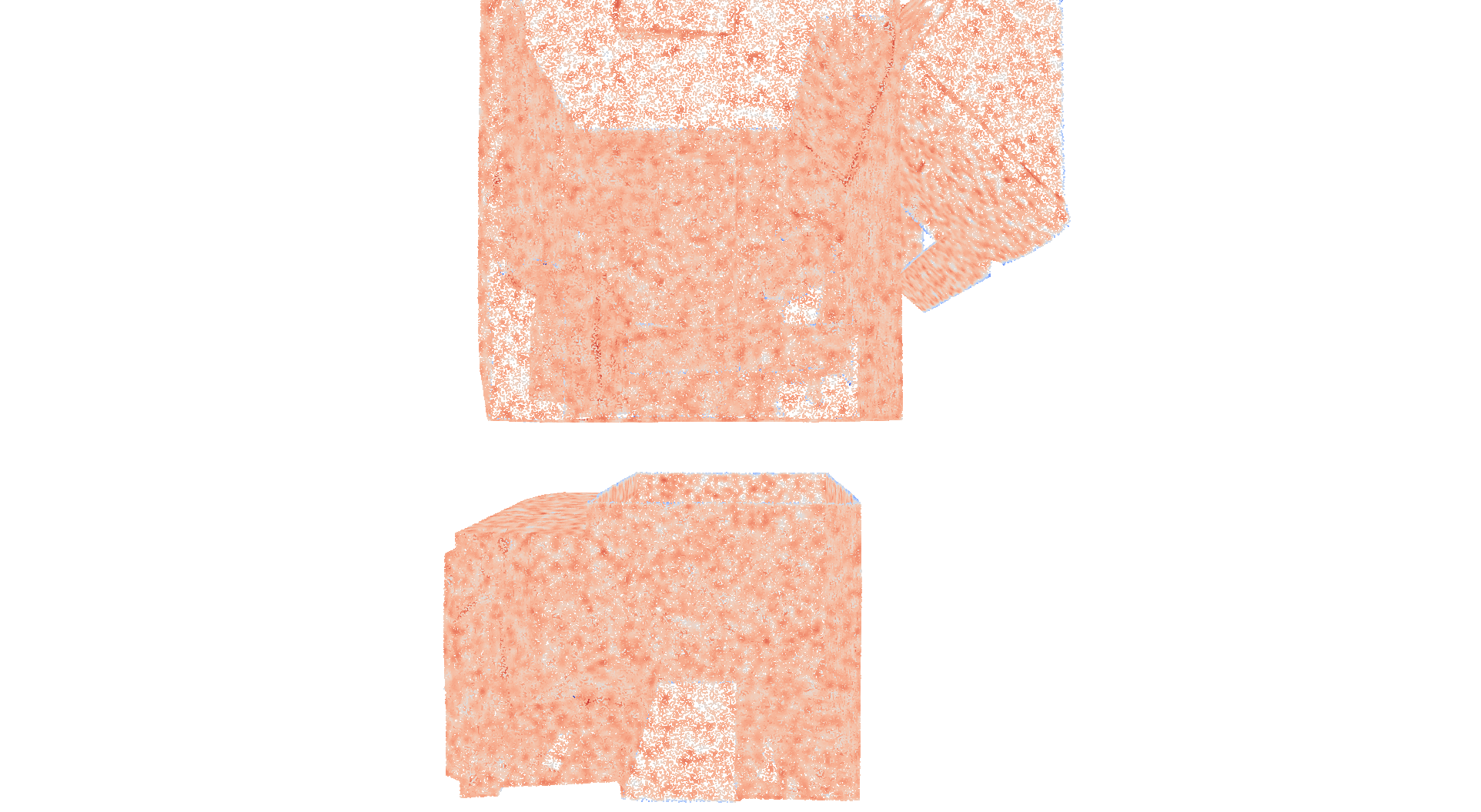}
        \caption{S3DIS}
    \end{subfigure}
    
    \vspace{0.3em}
    
    \begin{subfigure}{0.3\textwidth}
        \centering
        \includegraphics[width=\textwidth]{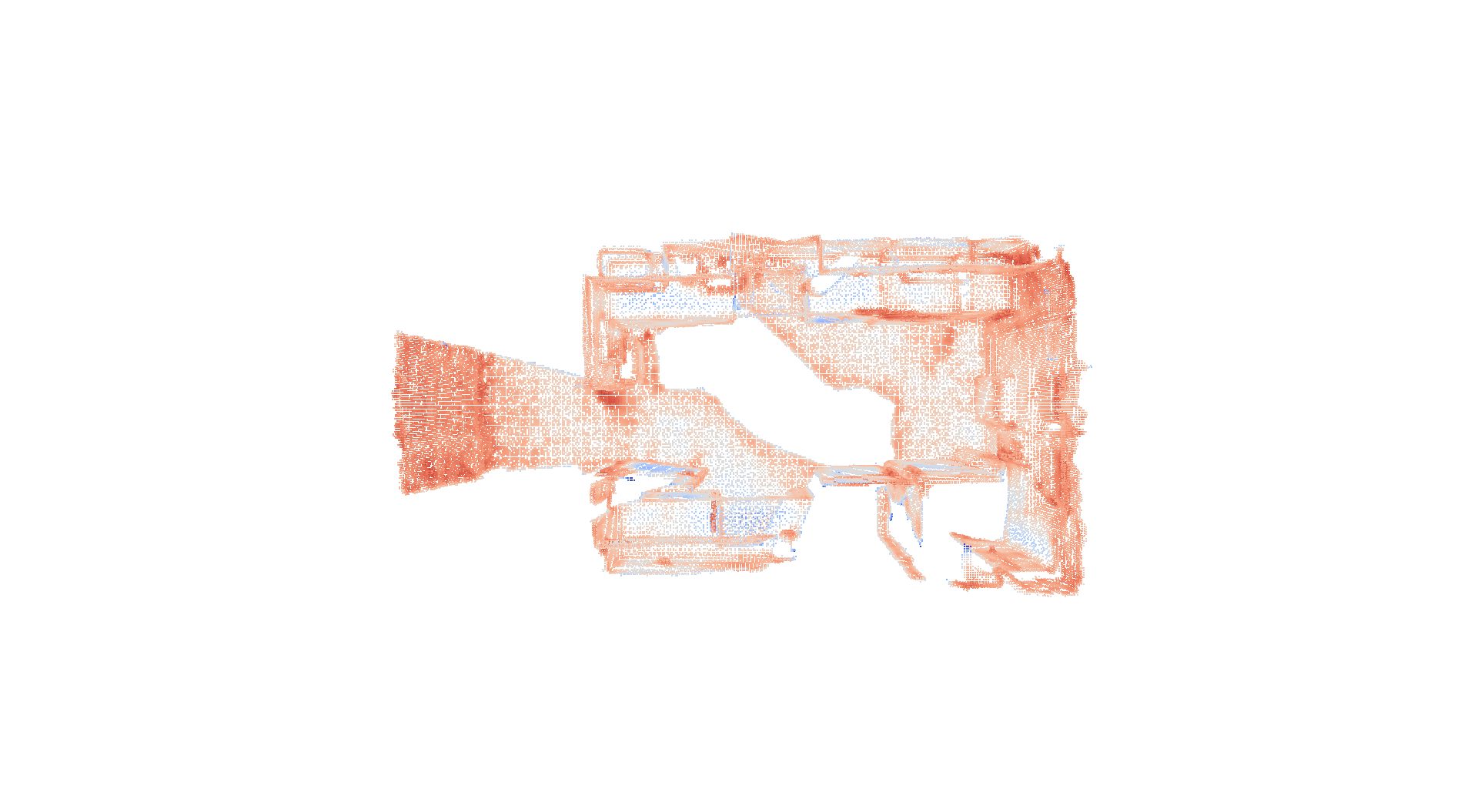}
        \caption{ScanNet}
    \end{subfigure}
    \hspace{0.02\textwidth}
    \begin{subfigure}{0.3\textwidth}
        \centering
        \includegraphics[width=\textwidth]{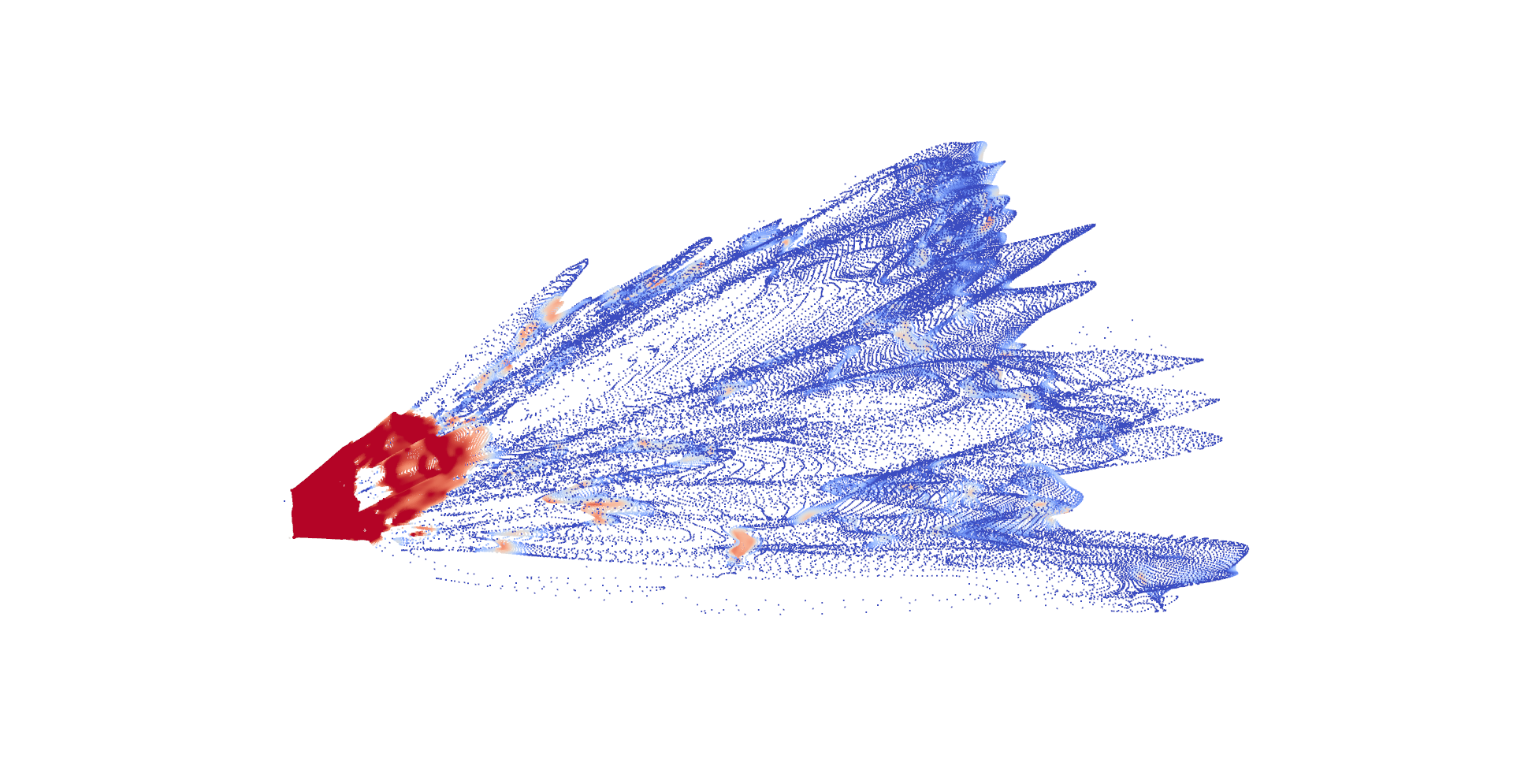}
        \caption{VGGT}
    \end{subfigure}
    \hspace{0.02\textwidth}
    \begin{subfigure}{0.3\textwidth}
        \centering
        \includegraphics[width=\textwidth]{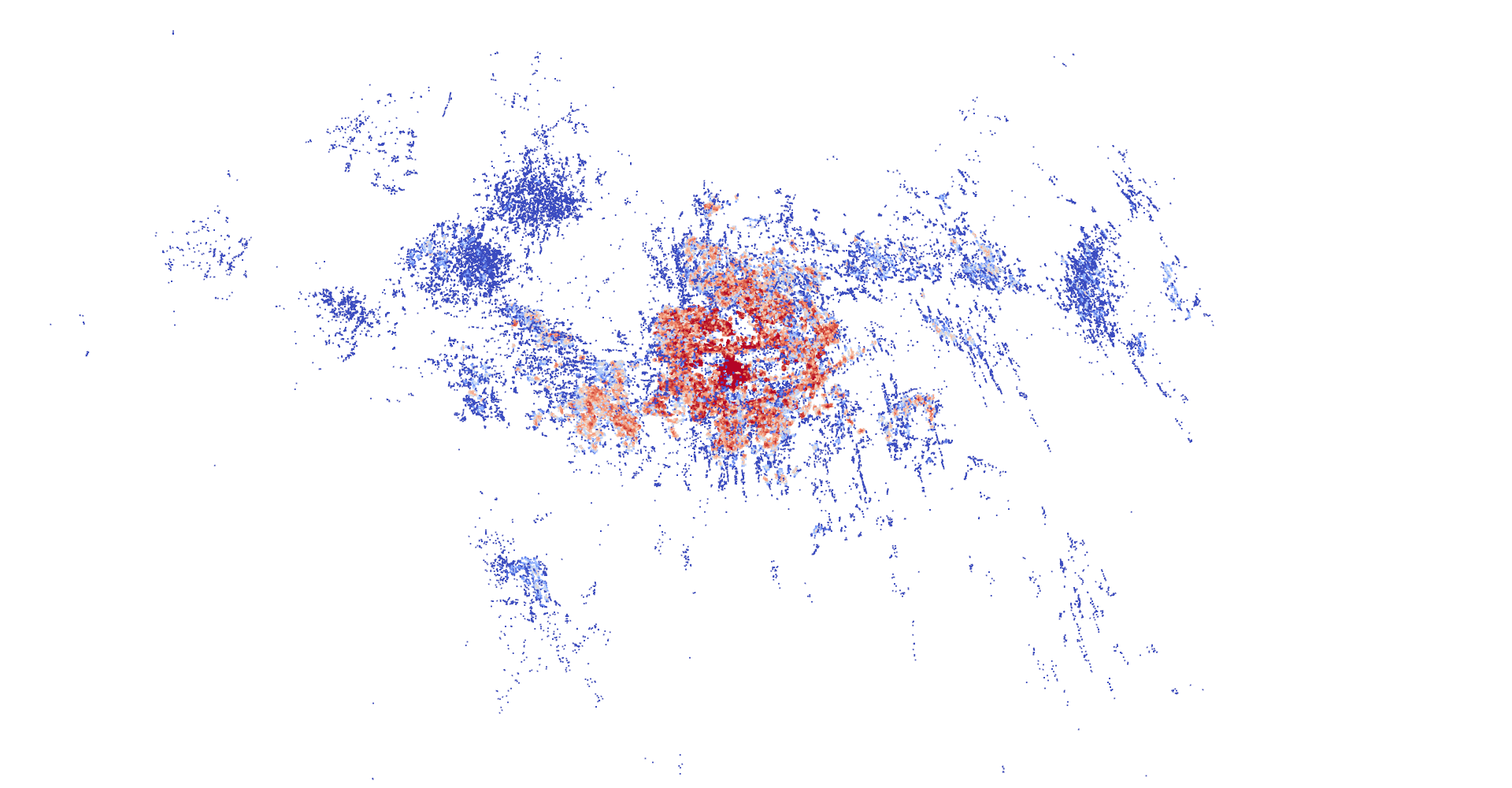}
        \caption{GS}
    \end{subfigure}

    \vspace{0.3em}
    
    \begin{subfigure}{0.75\textwidth}
        \centering
        \includegraphics[width=\textwidth]{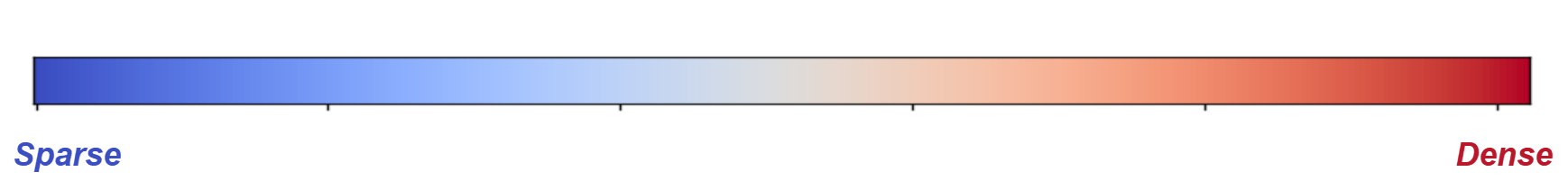}
    \end{subfigure}
    
    \caption{Visualization of 15 different point cloud datasets used in our experiments.}
    \label{fig:pcc_data_visual}
\end{figure*}

In this section, we visualize the 15 datasets used in our experiments, as depicted in Figure~\ref{fig:pcc_data_visual}.
Each point cloud is color-coded by its local point density, transitioning from blue (sparse) to red (dense).
The visualization clearly shows the diverse geometric structures in the benchmark: each category has a unique density profile, and challenging datasets show highly non-uniform distributions. Our unified model (Ours-U) successfully compresses all these structurally heterogeneous point clouds, which is a strong testament to its stability and generalization.

\section{Future Work}

\paragraph{Hybrid Explicit-Implicit Compression.}
AnyPcc demonstrates a powerful principle: fine-tuning a small subset of network parameters for each instance can yield significant compression gains. This approach becomes advantageous whenever the bitrate cost of transmitting the updated weights is less than the resulting bitrate savings in the geometry. This hybrid paradigm, which synergizes the strengths of explicit (entropy coding) and implicit (network-based representation) compression, can be readily extended to other domains such as point cloud attribute compression, as well as image and video compression. It offers the unique benefit of adjustable encoding complexity through control of the fine-tuning process. Furthermore, by integrating recent advances in implicit neural representations and leveraging parameter-efficient fine-tuning techniques from the large model domain, such as LoRA~\cite{hu2022lora}, we can develop even more efficient and versatile compression schemes.

\paragraph{Scaling Laws of Compression Models.}
Our ablation study, which explores channel counts from 32 to 128, reveals a clear trend of improved compression performance with increased model size, suggesting the potential for a scaling law. Drawing inspiration from the success of Large Language Models (LLMs), a promising future direction is to systematically scale up the AnyPcc architecture (e.g., to one billion parameters or beyond) to investigate the ultimate limits of neural point cloud compression.

\paragraph{Self-Supervised Learning and Data Curation.}
At its core, point cloud compression is a self-supervised learning task where the model learns to predict the data from itself. As we scale these models, the availability of vast and diverse datasets becomes paramount. Future work should therefore focus on two key areas: curating larger and more comprehensive collections of real-world point cloud data, and leveraging powerful generative models to synthesize high-quality training data. These efforts will be critical for successfully scaling self-supervised compression models and pushing the boundaries of generalization.

\noindent \textbf{Acknowledgment.} This work was supported by The Major Key Project of PCL (PCL2024A02), Natural Science Foundation of China (62271013), Guangdong Provincial Key Laboratory of Ultra High Definition Immersive Media Technology (2024B1212010006), Guangdong Province Pearl River Talent Program (2021QN020708), Guangdong Basic and Applied Basic Research Foundation (2024A1515010155), Shenzhen Science and Technology Program (JCYJ20240813160202004, JCYJ20230807120808017, SYSPG20241211173440004). (\textit{Corresponding author: Wei Gao})

{
    \small
    \bibliographystyle{ieeenat_fullname}
    \bibliography{main}
}

\end{document}